\documentclass[journal]{IEEEtran}

\usepackage{cite}
\usepackage[pagebackref=true,breaklinks=true,colorlinks,bookmarks=false]{hyperref}  
\usepackage{url}
\usepackage{float}
\usepackage{graphicx}
\usepackage{subfig}
\usepackage{amsmath}
\usepackage{multirow}

\begin{document}
%
\title{Learned Point Cloud Geometry Compression}
%
%
\author{Jianqiang Wang, Hao Zhu, Zhan Ma, Tong Chen, Haojie Liu, and Qiu Shen \\
Nanjing University}

\maketitle

\begin{abstract}
This paper presents a novel end-to-end {\it Learned Point Cloud Geometry Compression} (a.k.a., Learned-PCGC) framework, to efficiently compress the point cloud geometry (PCG) using deep neural networks (DNN) based variational autoencoders (VAE).  
In our approach, PCG is first voxelized, scaled and partitioned into non-overlapped 3D cubes, which is then fed into stacked 3D convolutions for compact latent feature and hyperprior generation. Hyperpriors are used to improve the conditional probability modeling of latent features. A weighted binary cross-entropy (WBCE) loss is applied in training while an adaptive thresholding is used in inference to remove unnecessary voxels and reduce the distortion.  Objectively, our method exceeds the geometry-based point cloud compression (G-PCC) algorithm standardized by well-known Moving Picture Experts Group (MPEG) with a significant performance margin, e.g., at least 60\% BD-Rate (Bj\"ontegaard Delta Rate) gains, using common test datasets.  
Subjectively, our method has presented better visual quality with smoother surface reconstruction and appealing details, in comparison to all existing MPEG standard compliant PCC methods.
Our method requires about 2.5MB parameters in total, which is a fairly small size for practical implementation, even on embedded platform. Additional ablation studies analyze a variety of aspects (e.g., cube size, kernels, etc) to explore the application potentials of our learned-PCGC.
\end{abstract}

\begin{IEEEkeywords}
Point cloud compression, geometry, 3D convolution, classification, end-to-end learning.
\end{IEEEkeywords}

\ifCLASSOPTIONpeerreview
\begin{center} \bfseries EDICS Category: 3-BBND \end{center}
\fi

\IEEEpeerreviewmaketitle

\section{Introduction}
\label{sec:intro}
\IEEEPARstart{P}{oint} cloud is a collection of discrete points with 3D geometry positions and other attributes (e.g., color, opacity, etc), which can be used to represent the volumetric visual data such as 3D scenes and objects efficiently~\cite{schwarz2019emerging}.  Recently, with the explosive growth of point cloud enabled applications such as 3D free viewpoint video and holoportation, high-efficiency Point Cloud Compression (PCC) technologies are highly desired.

Existing representative standard compliant PCC methodologies were developed under the efforts from the MPEG-I 3 Dimensional Graphics coding group (3DG)~\cite{3dg2017call, schwarz2019emerging}, of which geometry-based PCC (G-PCC)
for static point clouds and video-based PCC (V-PCC) for dynamic point clouds were two typical examples.  Both G-PCC and V-PCC relied on conventional models, such as octree decomposition~\cite{meagher1982geometric}, triangulated surface model, region-adaptive hierarchical transform~\cite{RAHT,TransformCoding_PC}, and 3D-to-2D projection. 
Other explorations related to the PCC are based on graphs~\cite{graph_dynamicPCC}, binary tree embedded with quardtree~\cite{BTQT}, or recently volumetric model~\cite{volumetric_PCC}. 

In another avenue, a great amount of deep learning-based image/video compression  methods~\cite{liu2019gated, balle2018variational, minnen2018joint} have emerged recently. 
Most of them have offered promising compression performance improvements over the traditional JPEG~\cite{JPEG}, JPEG 2000~\cite{JPEG2K}, and even High-Efficiency Video Coding (HEVC) intra profile-based image compression~\cite{liu2019gated,minnen2018joint,liu2019non}. These learned compression schemes have leveraged stacked DNNs to generate more compact latent feature representation for better compression~\cite{balle2018variational}, mainly for 2D images or video frames.

Motivated by facts that redundancy in 2D images can be well exploited by stacked 2D convolutions (and relevant nonlinear activation), we have attempted to explore the possibility to use 3D convolutions to exploit voxel correlation efficiently in a 3D space. In other word, we aim to use proper 3D convolutions to represent the 3D point cloud compactly, mimicking the way that 2D image blocks can be well synthesized by stacked 2D convolutions~\cite{minnen2018joint,liu2019non}. 
This paper focuses on static geometry compression, leaving other aspects (such as the compression of color attributes, etc) for our future study.

\begin{figure*}[t]
	\centering
	\subfloat[]{\includegraphics[width=7in]{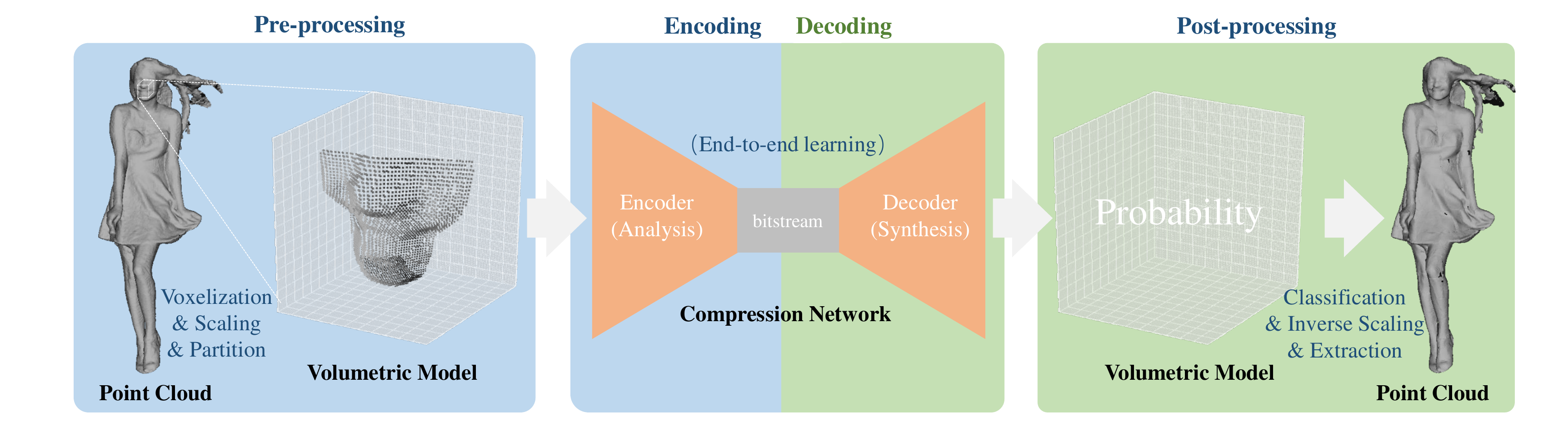}\label{fig:overview}}\\
	\subfloat[]{\includegraphics[width=7in]{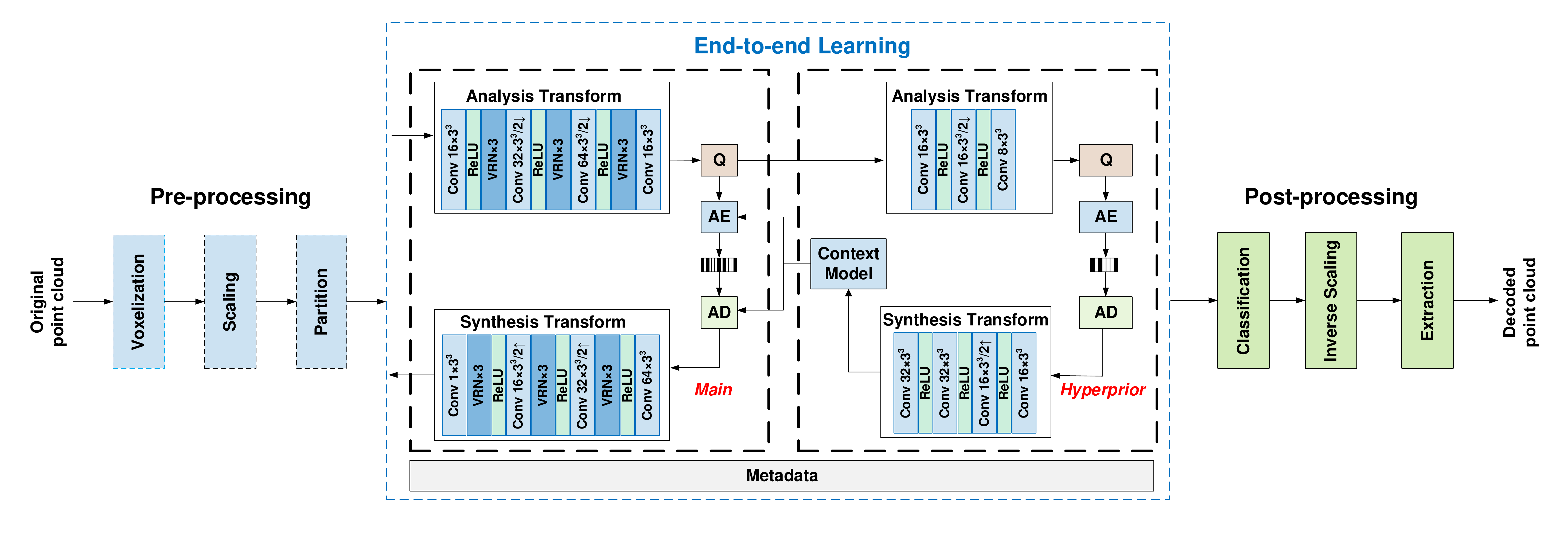}\label{fig:framework_diagram}}
	\caption{{\bf Learned-PCGC.} An illustrative overview in (a) and detailed diagram in (b) for point cloud geometry compression consisting of a pre-processing for PCG voxelization, scaling \& partition, a compression network for compact PCG representation and metadata signaling, and a post-processing for PCG reconstruction and rendering. ``Q'' stands for ``Quantization'', ``AE'' and ``AD'' are Arithmetic Encoder and Decoder respectively. ``Conv'' denotes convolution layer with the number of the output channels and kernel size, ``$\times3$'' means cascading three VRNs, ``$s\uparrow$'' and ``$s\downarrow$'' represent upscaling and downscaling at a factor of $s$. ``ReLU'' stands for the Rectified Linear Unit.}
	\label{fig:intro}
\end{figure*}

A high-level overview of our Learned-PCGC is given in {Fig.~\ref{fig:overview}, consisting of 1) a pre-processing for point cloud voxelization, scaling, and partition; 2) a variational autoencoder (VAE) based compression network; and 3) a post-processing for proper voxel classification, inverse scaling, and extraction (for display and storage).} Note that voxelization and extraction may be optional in the case that input point clouds are already in 3D volumetric presentation and not required to be stored in another non-volumetric format.

Generally, PCG data is typically voxelized for a 3D volumetric presentation. Each voxel uses a binary bit (1 or 0) to represent whether the current position at ($i$, $j$, $k$) is occupied as a positive and valid point (and its associated attributes). An analogous example of a voxel in a 3D space is a pixel in a 2D image. A (down)-scaling operation can be implemented to downsize input 3D volumetric model for better compression under a bit rate budget, especially at low bitrate scenarios. Corresponding (up)-scaling is required at another end for subsequent rendering.
 
Inspired by the successful block-based image processing, we propose to partition/divide the entire 3D volumetric model into non-overlapped {\it cubes}\footnote{Each 3D cube is measured by its height, width and depth, similar as the 2D block represented by its height and width. }, each of which contains $ W\times W\times W$ voxels.
The compression process runs cube-by-cube. In this work, operations are contained within the current cube without exploiting the inter-cube correlation. This ensures the complexity efficiency for practical application, by offering the parallel processing and affordable memory consumption, on a cubic basis.

For each individual cube, we use Voxception-ResNet~\cite{brock2016generative} (VRN) to exemplify the 3D convolutional neural network (CNN) for compact latent feature extraction. Similar as~\cite{liu2019gated, balle2018variational, liu2019non}, a VAE architecture is applied to leverage {\it hyperpriors} for better conditional context (probability) modeling when encoding the latent features. For an end-to-end training, the weighted binary cross-entropy (WBCE) loss is introduced to measure the compression distortion for rate-distortion optimization, while an adaptive thresholding scheme is embedded for appropriate voxel classification in inference.

To ensure the model generalization, our learned-PCGC is trained using various shape models from ShapeNet~\cite{chang2015shapenet}, and is evaluated using the common test datasets suggested by MPEG PCC group and JPEG Pleno group.  Extensive simulations have revealed that our Learned-PCGC exceeds existing MPEG standardized G-PCC by a significant margin, e.g., $\approx$ 67\%, and 76\% BD-Rate (Bjontegaard delta bitrate) gains using D1 (point2point) distance, and $\approx$ 62\%, and 69\% gains using D2 (point2plane) distance, against G-PCC using octree and trisoup models respectively.
Our method also achieves comparable performance in comparison to another standardized V-PCC. In addition to the aforementioned objective measures,
we have also reported that our method could offer much better visual quality (e.g., smoother surface and appealing details) when rendered using the surface model. This is mainly due to the inherently 3D structural representation using learned 3D transforms. 
{A fairly small net model (e.g., 2.5MB) is used, and parallel cube processing is also doable. This offers low complexity requirement, which is friendly for hardware or embedded implementation. }

%
The contributions of this paper are highlighted as follows:
\begin{itemize} 
	\item We have explored a novel direction to apply the learning-based framework, consisting of a pre-processing module for point cloud voxelization, scaling and partition, compression network for rate-distortion optimized representation, and a post-processing module for point cloud reconstruction and rendering, to represent point clouds geometry using compact features with the state-of-the-art compression efficiency;
	\item We have exemplified objectively and subjectively the efficiency by applying stacked 3D convolutions (e.g., VRN) in a VAE structure to represent the sparse voxels in a 3D space;
	\item Instead of directly using D1 or D2 distortion for optimization,  a WBCE loss in training and adaptive thresholding in inference are applied to determine whether the current voxel is occupied, as a classical classification problem. This approach works well for exploiting the voxel sparsity.

    \item Additional ablation studies have been offered to analyze a variety of aspects (e.g., partition size, kernel size, thresholding, etc) for our method to understand its capability for practical applications.
\end{itemize} 

The rest of this paper is structured as follows: Section~\ref{sec:related_work} reviews relevant studies on the compression of point clouds, learning-based image/video coding algorithms and recently emerged studies of autoencoders for point cloud processing; Our Learned-PCGC is given in Section~\ref{sec:learned_PCC} with systematic sketch and detailed discussions, followed by the experimental explorations and ablation studies to demonstrate the efficiency of our method; and concluding remarks are drawn in Section~\ref{sec:conclusion}.


\section{Related Work} \label{sec:related_work}

Relevant researches of this work can be classified as point cloud geometry compression, learned image compression, and recent emerging autoencoder-based point cloud processing.

\subsection{Point Cloud Geometry Compression}   
Prior PCGC approaches mainly relied on conventional models, including octree, trisoup, and 3D-to-2D projection based methodologies.

{\bf Octree Model.} A very straightforward way for point cloud geometry illustration is using the octree-based representation~\cite{Jackins1980Oct} recursively. Binary labels (1, or 0) can be given to each node to indicate whether a corresponding voxel or 3D cube is positively occupied. Such binary string can be then compressed using statistical methods, with or without prediction~\cite{Schnabel2006Octree, Huang2008A}. The octree-based approach has been adopted into the popular Point Cloud Library (PCL)~\cite{PCL2011}, and referred to as benchmark solution extensively~\cite{Kammerl2012Real}. MPEG standard compliant G-PCC~\cite{khaled2019gpcc} has also applied the octree coding mechanism, which is known as the {\it octree geometry codec}.  Most octree-based algorithms have shown decent efficiency for sparse point cloud compression, but limited performance for dense point cloud compression. 

{\bf Mesh/Surface Model.} Mesh/surface model could be regarded as the combination of point cloud and fixed vertex-face topology.
Thus, an alternative approach is to use a surface model for point cloud compression, as investigated in~\cite{Anis2016Compression, Pavez2018Dynamic}. In these studies, 3D object surfaces are represented as a series of triangle meshes, where mesh vertices are encoded for delivery and storage.  Point cloud after decoding is provided by sampling the reconstructed meshes. 
MPEG G-PCC has also included such triangulation-based mesh model, a.k.a., triangle soups representation of geometry~\cite{khaled2019gpcc} into the test model. This is known as the {\it trisoup geometry codec}. Such trisoup model is preferred for a dense point cloud.

{\bf Projection-Based Approach.} Other attempts have tried to project the 3D object to multiple 2D planes from a variety of viewpoints. This approach can leverage existing and successful image and video codecs. The key issue to this solution are how to efficiently perform the 3D-to-2D projections. As exemplified in MPEG  V-PCC~\cite{Vladyslav2018vpcc}, a point cloud is decomposed into a set of patches that are packed into a regular 2D image grid with minimum unused space. Padding is often executed to fill empty space for a piece-wise smooth image. With such projection, point cloud geometry can be converted into 2D depth images that can be compressed using the HEVC~\cite{Sullivan2013Overview}.
By far, V-PCC has exhibited the state-of-the-art coding efficiency compared with the G-PCC and PCL, etc, for geometry compression.

\subsection{Learned Image Compression} 
Recent explosive studies~\cite{balle2016end, minnen2018joint,  balle2018variational, liu2019gated, chen2017deepcoder} have shown that learned image compression offers better rate-distortion performance over the traditional JPEG~\cite{JPEG}, JPEG2000~\cite{JPEG2K}, and even HEVC-based Better Portable Graphics (BPG)\footnote{\url{https://bellard.org/bpg/}},  etc~\cite{minnen2018joint,liu2019non}. These algorithms are mainly based on the VAE structure with stacked 2D CNNs for compact latent feature extraction. Hyperpriors are used to improve the conditional probability estimation of latent features. 
While end-to-end learning schemes have been deeply studied for 2D image compression or even extended to the video~\cite{chen2017deepcoder}, there lack systematic efforts to study effective and efficient neural operations for 3D point cloud compression. One reason is that pixels in the 2D grid are more well structured and can be predicted via (masked) convolutions, but voxels in 3D cube present more sparsity, and unstructured local and global correlation, which is usually difficult for compression.
\subsection{Point Cloud Autoencoders} 
Existing point cloud representation and generation models using autoencoders serve as good references for point cloud compression. For example, Achlioptas~\textit{et al.}\cite{achlioptas2017learning} proposed an end-to-end deep autoencoder that directly accepts point clouds for classification and shape completions. Brock~\textit{et al.}\cite{brock2016generative} introduced a voxel-based VAE architecture using stacked 3D CNNs for 3D object classification. Dai~\textit{et al.}\cite{Dai2017Shape} applied a 3D-Encoder-Predictor CNNs for shape completions, and Tatarchenko~\textit{et al.}\cite{Tatarchenko2017Octree} reported a deep CNN autoencoder for an efficient octree representation.  
These works are mainly developed for machine vision tasks but not for compression, but their autoencoder architectures provide references for us to represent 3D point clouds efficiently. Inspired by these studies, we try to design appropriate transforms using autoencoders for compact representation.

{Quach~\textit{et al.}~\cite{quach2019learning} proposed a convolutional transforms based PCG compression method recently, which is the most relevant literature to our work.  
When both compared with PCL, our work offers larger gains. This is mainly due to the fairly redundant features using shallow network structure with large convolutions, {inaccurate} context modeling of latent features, etc. We will show more details in subsequent ablation studies.}

\section{Learned-PCGC: An Exploration via Learning} \label{sec:learned_PCC}
This section details each component designed in our Learned-PCGC, shown in Fig.~\ref{fig:framework_diagram}, consisting of a pre-processing, an end-to-end learning based compression network, and a post-processing.

\subsection{Pre-processing}
\label{subsec: processing}
{\bf Voxelization.} Point clouds may or may not be stored in its 3D volumetric representation.
Thus, an optional step is converting its raw format to a 3D presentation, typically using a ($i$, $j$, $k$)-based Cartesian coordinate system. This is referred to as the {\it voxelization}. Given that our current focus is the geometry of point cloud in this work, a voxel at ($i$, $j$, $k$) is set to 1, e.g., $V(i,j,k) =1$, if it has positive attributes, and $V(i,j,k)=0$ otherwise. Point cloud precision sets the maximum achievable value in each dimension. For instance, 10-bit precision allows $0\leq i,j,k\leq 2^{10} -1$. PCG is referred to its volumetric representation throughout this paper unless pointed out specifically. With such a volumetric model for a PCG after voxelization, it captures inter-voxel correlations in a 3D space, which is better for us to apply the subsequent 3D convolutions to exploit the efficient and compact representation.

{\bf Scaling.}  Image downscaling was used in image/video compression~\cite{downscaling_videocoding} to preserve image/video quality under a constrained bit rate, especially at a low bit rate. Thus, this can be directly extended to point clouds for better rate-distortion efficiency at the low bit rate range. On the other hand, scaling can be also used to reduce the sparsity for better compression by zooming out the point cloud, where the distance between sparse points gets smaller, and point density within a fixed size cube increases. As will be revealed in later experiments, applying a scaling factor in pre-processing leads to noticeable compression efficiency gains for sparse point cloud geometry, such as Class C, and yields well-preserved performance at low bit rates for fairly dense Class A and B, shown in Fig.~\ref{fig:testdata} and Table~\ref{table:testdataset}.

In this work, we propose a simple yet effective operation via direct {\it downscaling} and {\it rounding} in advance. Let ${\bf X}_n = \left(i_n, j_n, k_n\right)$, ${n=1 \cdots N} $ be the set of points of the input point cloud. 
    We scale this point cloud by multiplying ${\bf X}_n$ with a scaling factor $ s $, $s < 1$, and round it to  the closest integer coordinate, i.e., 
    \begin{align}
         \hat{\bf X}_{n} &= {\sf ROUND}({\bf X}_{n} \times s) \nonumber\\
         &= {\sf ROUND}(i_n\times s, j_n\times s, k_n \times s).
    \end{align} 
Duplicate points at the same coordinate after rounding are simply removed for this study. An interesting topic is to exploring the adaptive scaling within the learning network. However, it requires substaintial efforts and is deferred as our study. On the other hand, applying the simple scaling operations in pre-processing is already demonstrated as an effective scheme as will be unfolded in later experimental studies. 

    \begin{figure}[t]
    	\begin{center}
    	\includegraphics[height=1.8in]{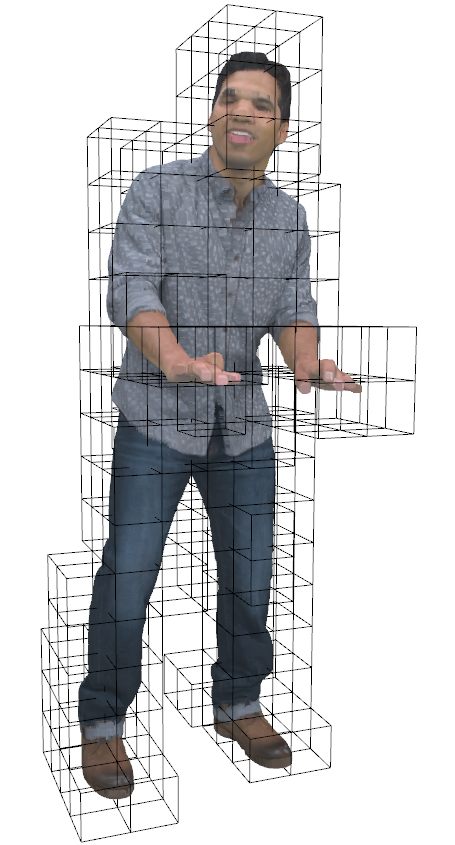}
    	\end{center}
    	\caption{Point cloud partitioned into non-overlapped cubes. Those cubes with occupied valid voxels are highlighted using bounding boxes.}
    	\label{fig:partition}
    \end{figure}

{\bf Partition.} Typically a point cloud geometry presents a large volume of data, especially for it with large precision. It is difficult and costly to process an entire point cloud at a time. Thus, motivated by the successful block based processing pipeline adopted in popular image/video standards, we have attempted to partition the entire point cloud into non-overlapped cubes, as shown in Fig.~\ref{fig:partition}. Each cube is at a size of $W\times W \times W$. 

The geometry position of each cube can be signaled implicitly following the raster scanning order from the very first one to the last one, regardless of whether a cube is completely null or not. Alternatively, we can specify the position of each cube explicitly using the existing octree decomposition method, leveraging the sparse characteristics of the point cloud. Each valid cube (e.g., with at least one occupied voxel) can be seen as a super-voxel at a size of $W\times W \times W$. Thus, the number of super-voxel is limited, in comparison to the number of voxels in the same volumetric point cloud. As revealed in later ablation studies, signaling cube position explicitly using the existing octree compression method~\cite{PCL2011} only requires a very small percentage (e.g., \textless1\%) overhead. In the meantime, the number of occupied voxels in each cube is also transmitted for later classification-based point reconstruction. In summary, we treat the geometry position and the number of occupied voxels of each individual (and valid) cube as the metadata that is encapsulated in the compressed binary strings explicitly.

In the current study, each cube is processed independently without exploring their intercorrelations. Massive parallelism can be achieved by enforcing the parallel cube processing. Assuming the geometry position of a specific cube is $ (i_c, j_c, k_c)$, global coordinates of a voxel can be easily converted to its local cubic coordinates,
\begin{equation}
       \hat{\bf X}^{\tt loc}_{n} = \hat{\bf X}_{n} - (i_c\times W, j_c\times W, k_c\times W),    
\end{equation} for the following learning-based compression.

\subsection{Cube-based Learned-PCGC}
\label{subsec: cnn}

We aim to find a more compact representation of any input cube with sparsely distributed voxels. It mainly involves the pursuit of appropriate transforms via stacked 3D CNNs, and accurate rate estimation, and a novel classification-based distortion loss measure for end-to-end optimization. 

{\bf 3D Convolution-based Transforms.} Transforms are used for decades to represent the 2D image and video data in a more compact format, from the discrete cosine transform, to recently emerged learned convolutions based approaches. Especially, those learned 2D transforms have demonstrated promising coding performance in image compression~\cite{liu2019non, balle2018variational} via stacked CNNs based autoencoders, by exploring the local and global spatial correlations efficiently.  

Thus, an extension is to design proper transforms based on stacked 3D CNNs to represent the 3D point cloud. In the encoding process, forward transform is analyzing and exploiting the spatial correlation. Thus it can be referred to as the ``Analysis Transform''. Ideally, 
for any $W\times W\times W$ cube, we aim to derive compact latent features $ y $, which are represented using a 4-D tensor with the size of $(channel, length, width, height)$. The {\it analysis transform} can be formulated as:
\begin{equation}
    y=f_{e}(x; \theta_{e}),
\end{equation}
with $\theta_{e}$ for convolutional weights.

Correspondingly, a mirroring {\it synthesis transform} is devised to decode quantized latent features $ \hat{y} $ into a reconstructed voxel cube $\tilde{x}$, which can be formulated as:
\begin{equation}
    \tilde{x}=f_{d}(\hat{y}; \phi_{d})
\end{equation} with  $\phi_{d}$ as its parameters.

Analysis and synthesis transformations are utilized in both main and hyper encoder-decoder pairs, shown in Fig.~\ref{fig:framework_diagram}. 

\begin{figure}[t]
	\begin{center}
	\includegraphics[width=2.8in]{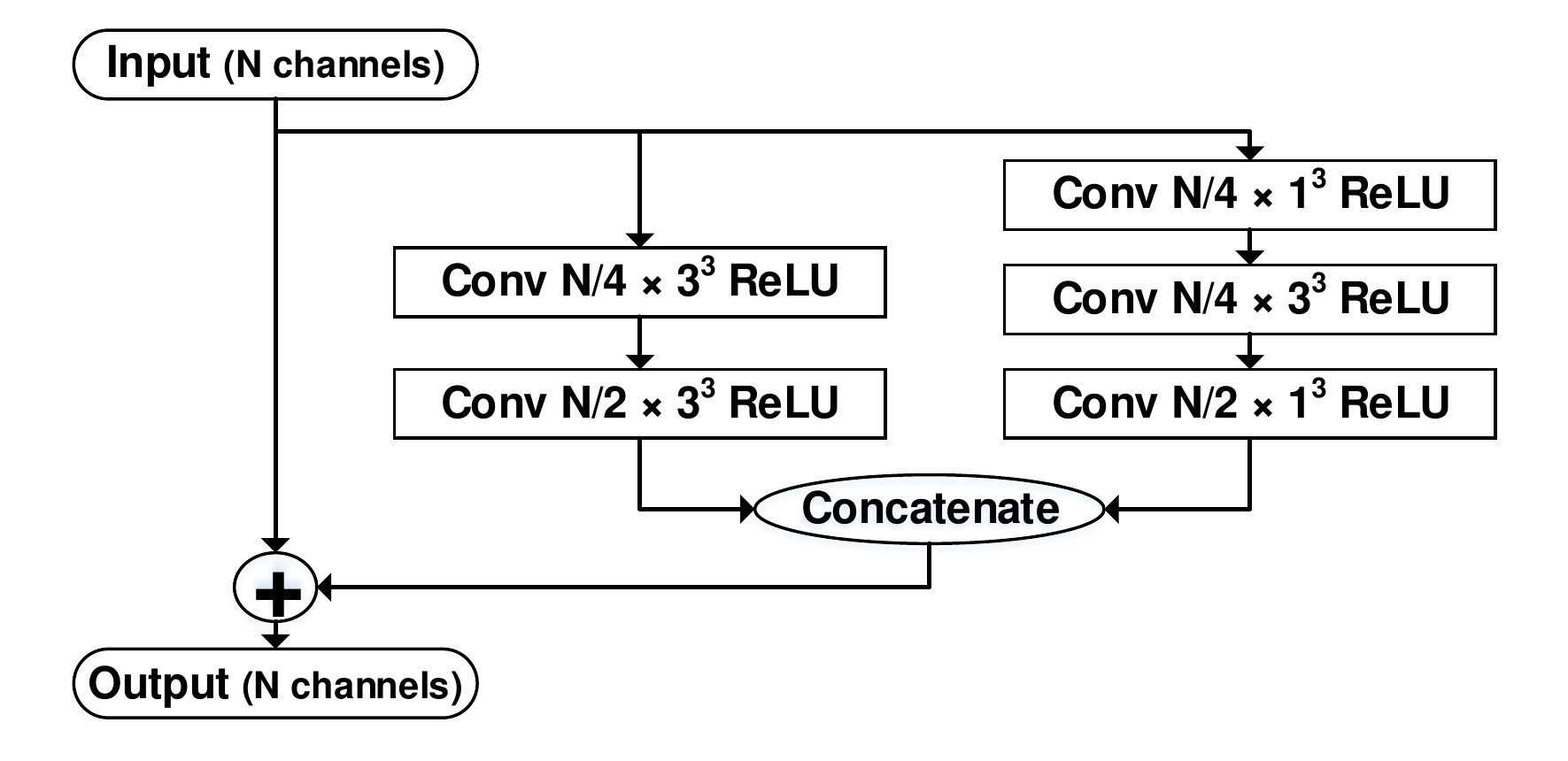}
	\end{center}
	\caption{Voxception-ResNet Blocks (VRN).}
	\label{fig:vrn}
\end{figure}

In this work, we use Voxception-ResNet (VRN) structure proposed in~\cite{brock2016generative}
as the basic 3D convolutional unit in the main codec, for its superior efficiency inherited from both residual network~\cite{he2016deep} and inception  network~\cite{szegedy2017inception}.  The architecture of VRN is illustrated in Fig.~\ref{fig:vrn}. In the main codec, nine stacked VRNs are used for both analysis and synthesis transform. 

Given that hyperpriors are mainly used for latent feature entropy modeling, we apply three consecutive lightweight 3D convolutions (with further downsampling mechanism embedded) instead of in hyper codec. Decoded {\it hyperpriors} are then used to improve the conditional probability of latent features from the main codec. Details regarding the entropy rate modeling are given in Section~\ref{subsec:entropy}. 

In this work, we have applied relative small kernels for convolutions, e.g., $1\times1\times1$ or $3\times3\times3$, which is then integrated with VRN model efficiently to capture the essential information for a compact representation. In the meantime, smaller convolution kernels are also implementation friendly with lower complexity.

\subsection{Quantization}
A simple yet effective {\it rounding} operation is used for feature quantization in inference, i.e.,
\begin{equation}
    \hat{y}={\sf ROUND}(y),
\end{equation}
where $y$ and $\hat{y}$ represent original and quantized  representations respectively. 

However, direct {\it rounding}  is not differentiable for backpropagation in  the end-to-end training scheme. Instead, we approximate the rounding process by adding uniform noise to ensure the differentiability,
\begin{equation}
   \hat{y}=y+\mu,
\end{equation}
where $\mu$ is random uniform noise ranging from $-\frac{1}{2}$ and $\frac{1}{2}$,  $\hat{y}$ represents ``noisy'' latent representations with actual rounding error. $\hat{y}$ follows a uniform distribution $\mathcal{U}$ centered on $y$:
$\hat{y}\sim \mathcal{U}(y-\frac{1}{2},y+\frac{1}{2})$. Such approximation using added noise is also used in~\cite{balle2018variational}.

\subsection{Entropy Rate Modeling}
\label{subsec:entropy}
Entropy coding is critical for source compression to exploit statistical redundancy. Among existing approaches, arithmetic code is widely used and adopted in standards and products because of its superior performance. Thus we choose the arithmetic coding to compress each element of quantized latent feature.
Theoretically, the entropy bound of the source symbol (e.g., feature element) is closely related to its probability distribution, and more importantly, accurate rate estimation plays a key role in lossy compression for rate-distortion optimization~\cite{sullivan1998rate}. 

We can approximate the actual bit rate of the quantized latent feature via
\begin{equation}
    R_{\hat{y}} = E_{\hat{y}}[-\log_{2}p_{\hat{y}}(\hat{y})], \label{eq:self_pdf}
\end{equation} with  $p_{\hat{y}}(\hat{y})$ as the self probability density function (p.d.f.) of $\hat{y}$. Rate modeling can be further improved from \eqref{eq:self_pdf} if we can have more priors. Thus, in existing learned image compression algorithms~\cite{balle2018variational,liu2019non}, a VAE structure is enforced to have both main and hyper codecs. In hyper codec, dimensions of latent features are further downscaled to provide hyperpriors $z$ without the noticeable overhead. These hyperpriors $\hat{z}$ are decoded as the prior knowledge for better probability approximation of latent feature $\hat{y}$ when conditioned on the distribution of $\hat{z}$.

Note that the same quantization process will be applied to both latent features and hyperpriors. Following the aforementioned discussion, we can model the decoded hyperpriors (e.g., with assumed uniform rounding noise) using a fully factorized model, i.e.,
\begin{equation}
    \label{pz}
        p_{\hat{z}|\psi}(\hat{z}|\psi) =\prod_{i} \left(p_{\hat{z}_{i}|\psi^{(i)}}(\psi^{(i)}) * \mathcal{U}( -\frac{1}{2},\frac{1}{2})\right)(\hat{z}_{i}),
    \end{equation}
    where $ \psi^{(i)} $ represents the parameters of each univariate distribution $ p_{\hat{z}_{i}|\psi^{(i)}} $.
 Therefore, a Laplacian distribution  $\mathcal{L}$ is used to approximate the p.d.f. of  ${\hat{y}}$ when conditioned on the hyperpriors, i.e.,
    \begin{equation}
    \label{py}
        p_{\hat{y}|\hat{z}}(\hat{y}|\hat{z}) 
        =\prod_{i}\left(\mathcal{L}(\mu_{i}, \sigma_{i})*\mathcal{U}(-\frac{1}{2},\frac{1}{2})\right)(\hat{y}_{i}).
\end{equation}
    The mean and variance parameters $ (\mu_{i},\sigma_{i}) $ of each element $\hat{y}_{i}$ are estimated from the decoded hyperpriors.
 
\subsection{Rate-distortion Optimization}
\label{subsec:loss}
Rate-distortion optimization is adopted in popular image and video compression algorithms to trade-off the distortion ($D$) and bit rate ($R$).
In our end-to-end learning framework, we follow the convention and define the Lagrangian loss for training, so as to maximize the overall rate-distortion performance, i.e.,
\begin{equation}
\label{loss}
  J_{\tt loss} = R + \lambda D,
\end{equation}
where $\lambda$ controls the trade-off for each individual bit rate.

\textbf{Rate Estimation}:
In our VAE structure based compression framework, 
a total rate consumption comes from the $\hat{y}$ and $\hat{z}$. Referring to \eqref{pz} and \eqref{py}, rate approximation can be written as
\begin{align}
    R_{\hat{y}} &= \sum_{i}-\log_{2}(p_{\hat{y}_{i}|\hat{z}_{i}}(\hat{y}_{i}|\hat{z}_{i})), \\
    R_{\hat{z}} & = \sum_{i}-\log_{2}(p_{\hat{z}_{i}|\psi^{(i)}}(\hat{z}_{i}|\psi^{(i)})).
\end{align}
The total rate can be easily derive via the summation, e.g., $ R = R_{\hat{y}} + R_{\hat{z}} $. 
Here, rate spent by hyperpriors $\hat{z}$ could be regarded as the side information or overhead, occupying merely less bits
than the the latent representations $\hat{y}$ in our design. Note that we only use hyperpriors for rate estimation, without including any autoregressive spatial neighbors~\cite{liu2019non, liu2019gated}. This is driven by the fact that voxels are distributed sparsely, thus, neighbors may not bring many gains in context modeling, but may break the voxel parallelism with large complexity.

\textbf{Distortion Measurement}:
Existing image/video compression approaches use MSE or SSIM as the distortion measures. In this work, 
we have proposed a novel classification-based mechanism to measure the distortion instead.
Such classification method fits the natural principle to extract  valid point cloud data after decoding. More specifically,
decoded voxels in each cube usually present in a predefined range, e.g., from 0 to 1 in this work, from 0 to 255 if 8-bit integer processing enforced.
Recalling that each valid voxel in a point cloud geometry tells that this position is concretely occupied. Simple binary flag, ``1'' or ``TRUE'' often refers to the occupied voxel, while ``0'' or ``FALSE'' for the null or empty voxel. Therefore, decoded voxel needs to be classified into either 1 or 0 accordingly.

Towards this goal, we use a weighted binary cross-entropy (WBCE) measurement as the distortion in training, i.e.,
\begin{equation}
\label{wbce}
	{D}_{\tt WBCE} =
	\dfrac{1}{N_{o}}\sum^{N_{o}}-\log p_{\tilde{x}_o}+\alpha\dfrac{1}{N_{n}}\sum^{N_{n}}-\log ( 1-p_{\tilde{x}_n}),
	\end{equation}  
where $ p_{\tilde{x}} = {\sf sigmoid}(\tilde{x}) $ is used in this work to enforce $ p_{\tilde{x}} \in (0, 1) $ as the probability of being occupied, $ \tilde{x}_o $ represents occupied voxels, $ {\tilde{x}_n} $ represents null voxels, and $N_o$, $N_n$ represent the numbers of occupied and null voxels, respectively. Note that we do not classify voxel $\tilde{x}$ into a fixed 1 or 0, but let $\tilde{x} \in (0,1)$ to guarantee the differentiability in backpropagation used in training.
Different from the standard BCE loss that weights positives and negatives equally, we calculate the mean loss of positive and negative samples separately with a hyperparameter $ \alpha $ to reflect their relative importance and balance the loss penalty.  We set $ \alpha $ to 3 according to our experiments.

\subsection{Post-Processing}

{\bf Classification.} 
In the inference stage, decoded voxels $\tilde{x}$ in each point cloud cube is presented as a floating number in (0,1), or an 8-bit integer in (0, 255), according to the specific implementation. Thus, we first need to classify it into binary 1 or 0.
A fixed threshold can be easily applied, for example, a median value $t_h = 0.5$, however, performance often suffers as shown in Fig.~\ref{pointsnumbersselection}. Instead, we propose an {\it adaptive thresholding} scheme for voxel classification, according to the number of occupied points in the original point cloud cube. This information is embedded for each cube as the metadata. Since $ p_{\tilde{x}} $ can be also referred to as the probability of being occupied, we sort $ p_{\tilde{x}} $ to extract the top $k$ voxels, which are most likely to be occupied.  Top-$k$ selection fits the distortion criteria used in \eqref{wbce} for end-to-end training, e.g., minimizing the WBCE by enforcing processed  voxel distribution (i.e., occupied or null) close to the original input distribution as much as possible.

Detailed discussion is given in subsequent ablation studies.
 
{\bf Inverse Scaling.} A mirroring inverse scaling with a factor of $1/s$  is implemented in post-processing, in contrast to the scaling in pre-processing, when completing the inference of all cubes for rendering and display. This work  applies a very simple linear scaling strategy. A complex scaling scheme could be used to retain reconstructed quality better, such as content adaptive scaling.  This is an interesting topic to explore as our future study.

\textbf{Extraction}
Extraction is an optional step in post-processing, as the voxelization part in pre-processing. This part is used to convert 3D point cloud into another file format for storage or exchange, such as the ASCII or polygon file format (ply) used by the MPEG PCC group. For the scenarios that original point clouds are already in 3D volumetric presentation, or decoded point clouds are used for direct display, extraction is not necessarily required.

\subsection{Bitstream Specification}
Following the above discussion, our Learned-PCGC runs iteratively for each cube in this work. It will encapsulate 1) cube position {\tt cube\_pos}, 2) the number of original occupied voxel  {\tt num\_occupied\_voxel}, 3) entropy coded features and hyperpriors, for each cube, into the binary bitstream for delivery and exchange. Here, we refer to part 1) and 2) as the {\it metadata} or (payload overhead), and 3) to as the main {\it payload}.

{\bf Metadata.} For {\tt cube\_pos}, we simply use the octree model in \cite{tmc13} to indicate the location of the current cube in a volumetric point cloud. Since {\tt num\_occupied\_voxel} is used for classification, we embed it directly here. For the worst case, we need $3\log_2^W$ bits for the cube at a size of $W\times W\times W$. In practice, {\tt num\_occupied\_voxel} might be much less than $2^{{3\log_2^W}}$ because of its sparse nature. Alternatively, we can signal another syntax element, such as {\tt max\_num\_occupied\_voxel} for the entire point cloud to bound the number of voxels in each cube then.

{\bf Payload} As seen, both features and hyperpriors are encoded for the Learned-PCGC. Syntax elements for hyperpriors and latent features (in corresponding fMaps) are encoded consecutively using arithmetic coder. Context probability of hyperpriors is based on a fully factorized distribution, while context probability of latent features is conditioned on the hyperpriors.

\section{Experimental Studies}
\label{sec:exp}

\subsection{Training}
\begin{figure}[t]
\centering
\subfloat{\includegraphics[width=0.8in]{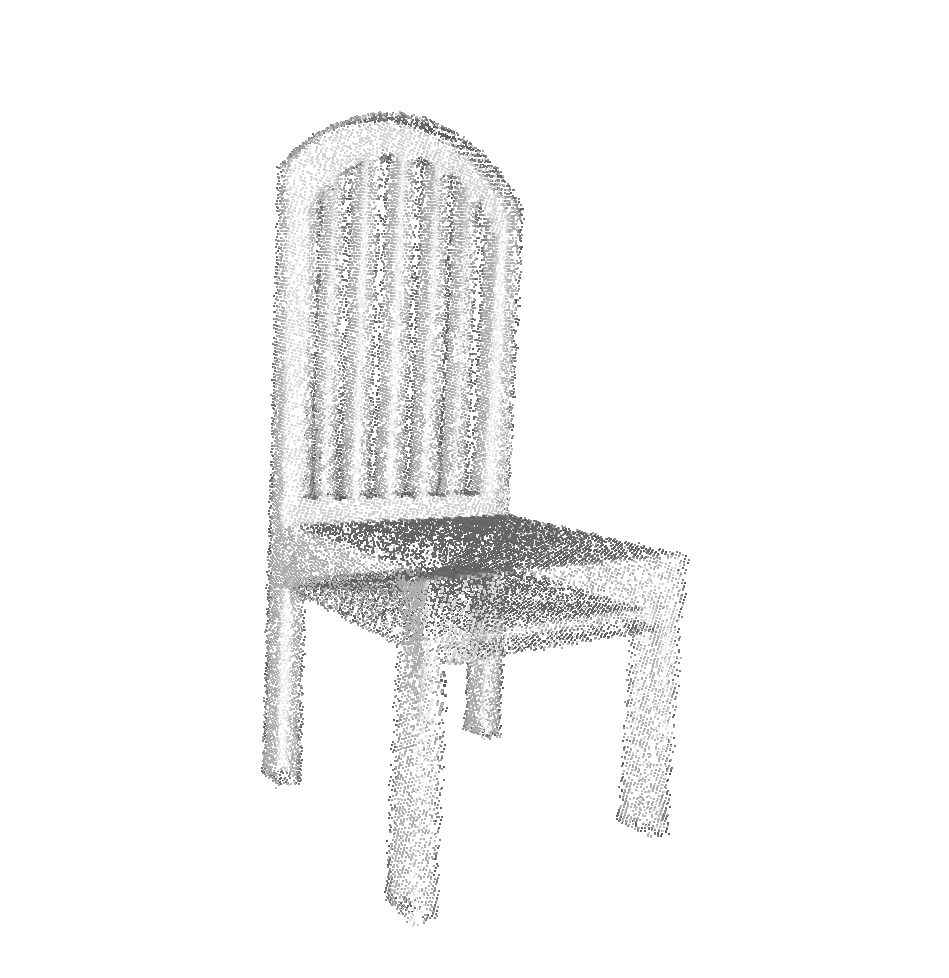}}
\subfloat{\includegraphics[width=0.8in]{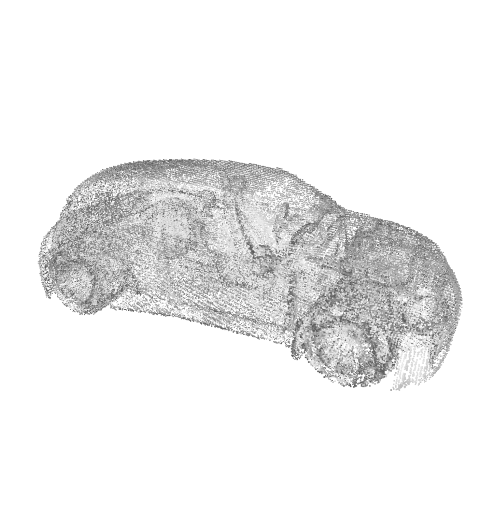}}%
\subfloat{\includegraphics[width=0.8in]{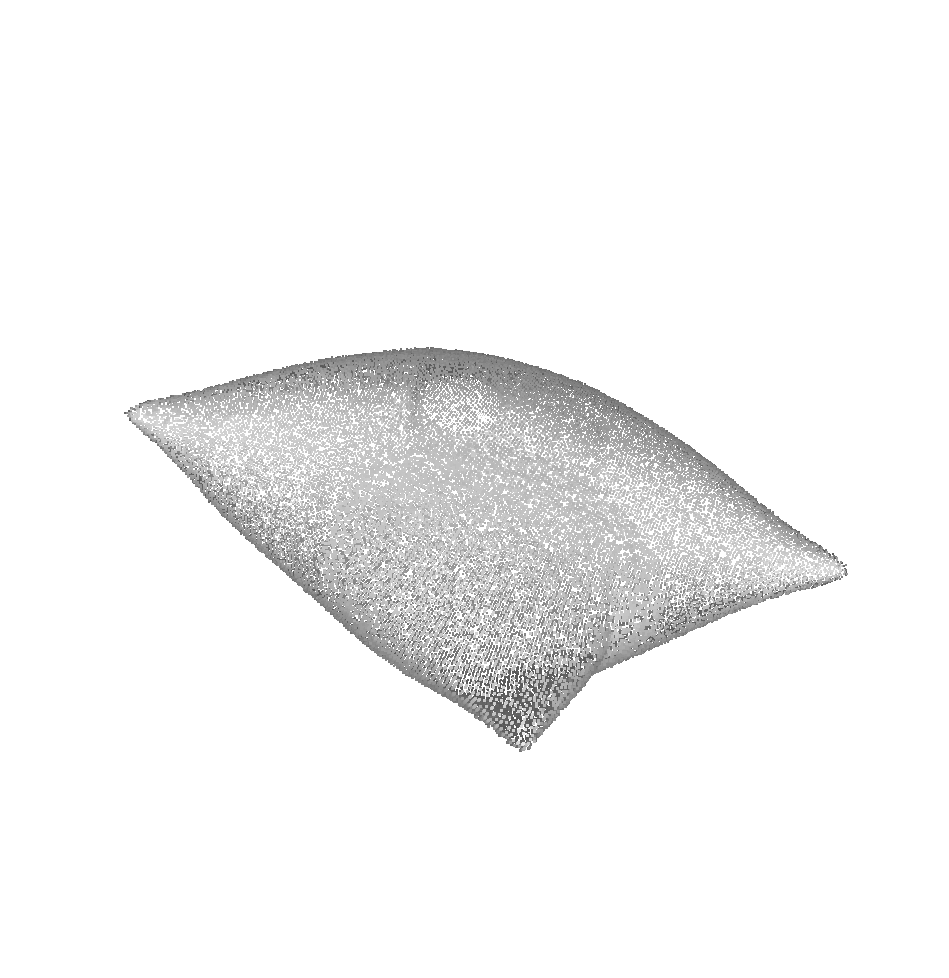}}%
\subfloat{\includegraphics[width=0.8in]{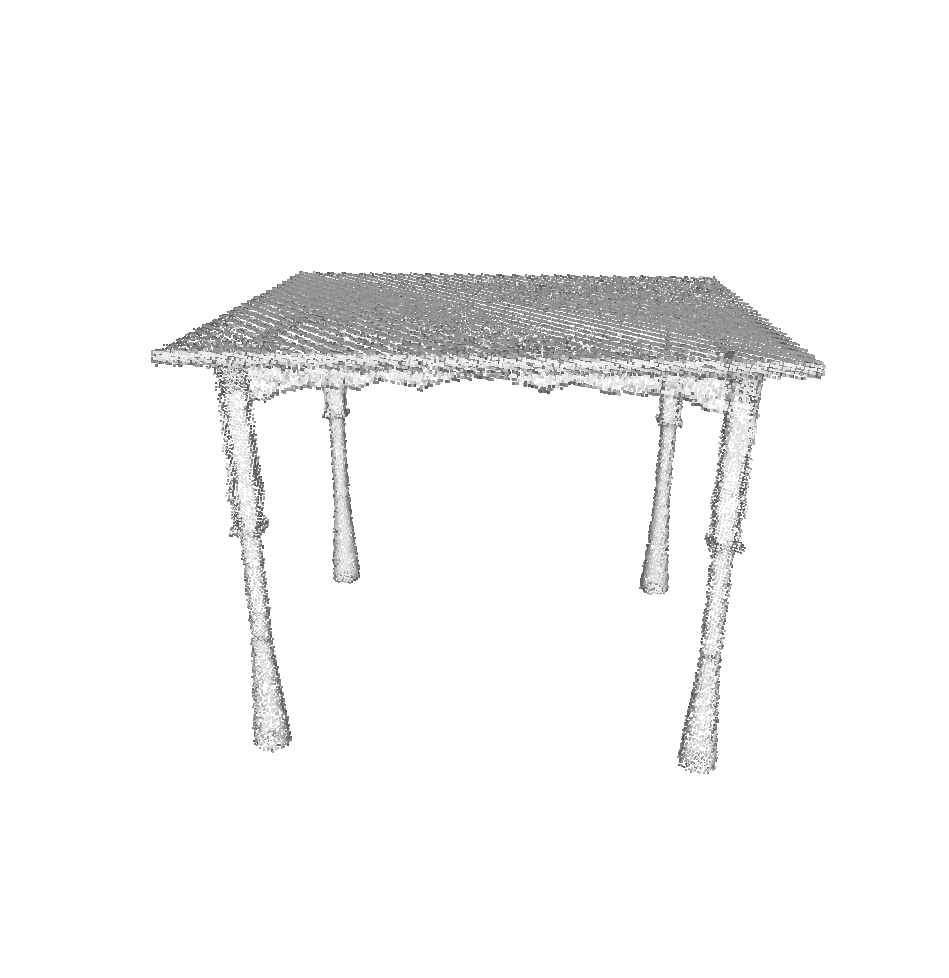}}\\
\subfloat{\includegraphics[width=0.8in]{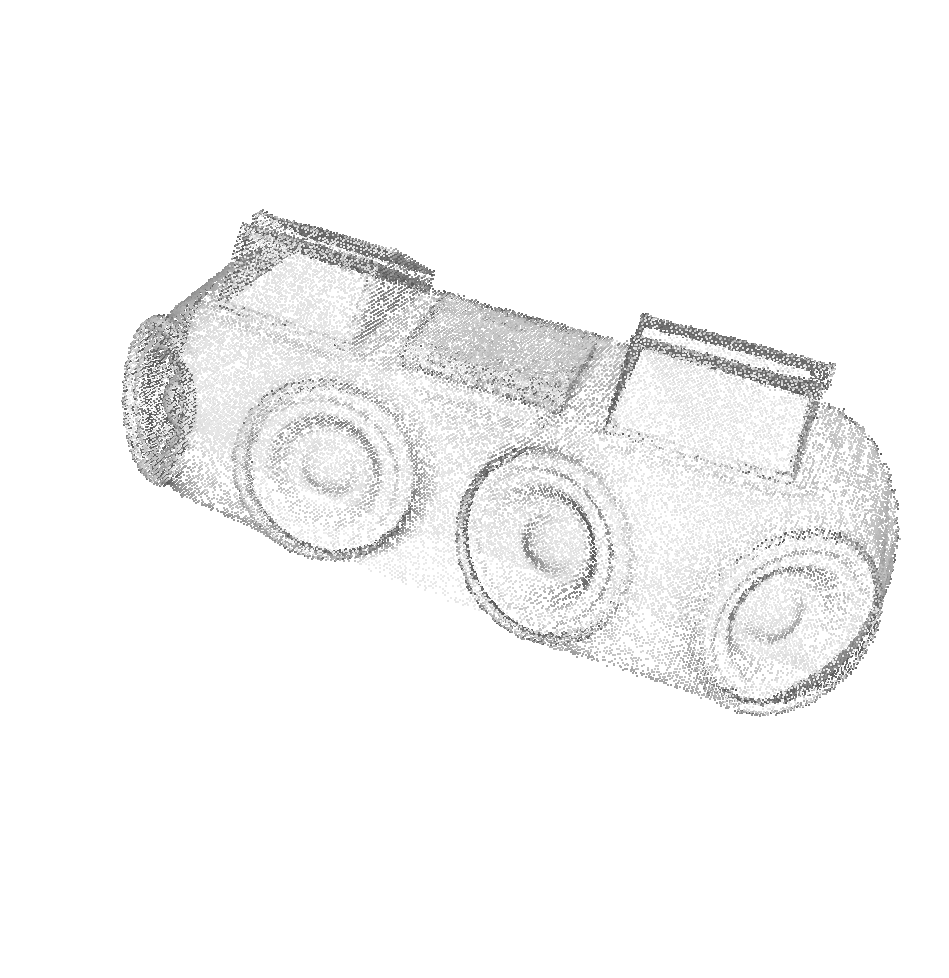}}%
\subfloat{\includegraphics[width=0.8in]{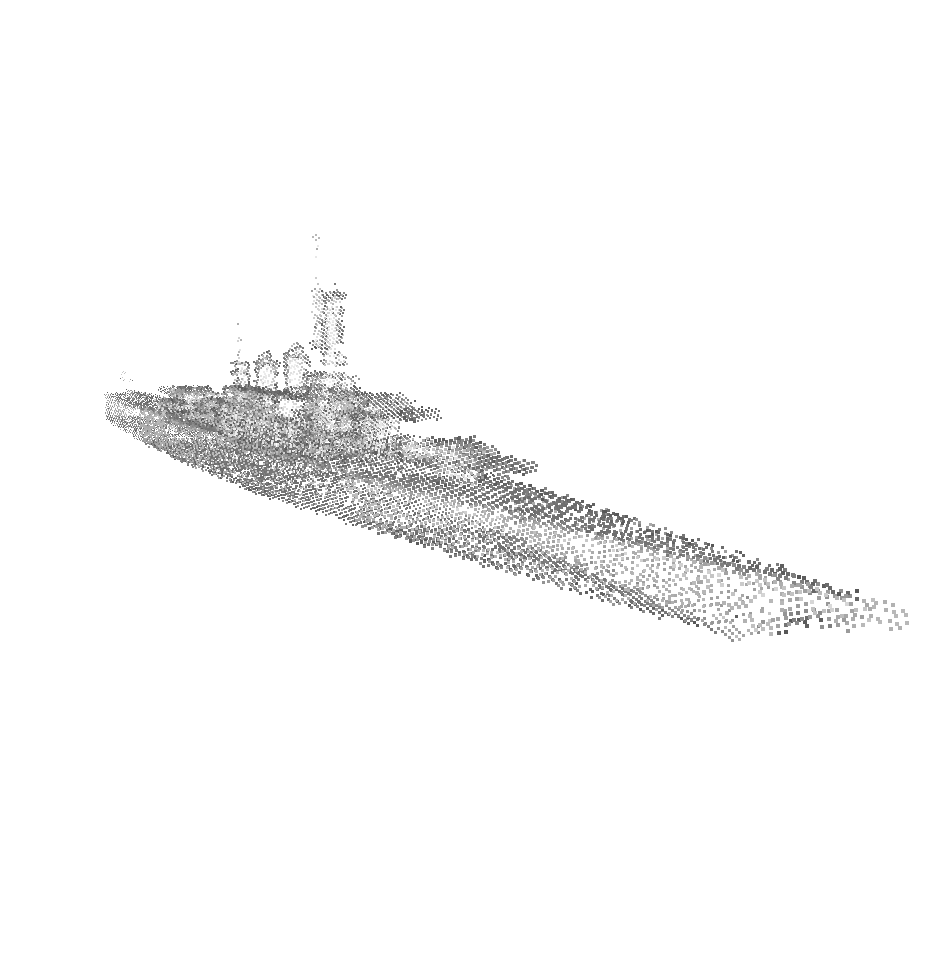}}%
\subfloat{\includegraphics[width=0.8in]{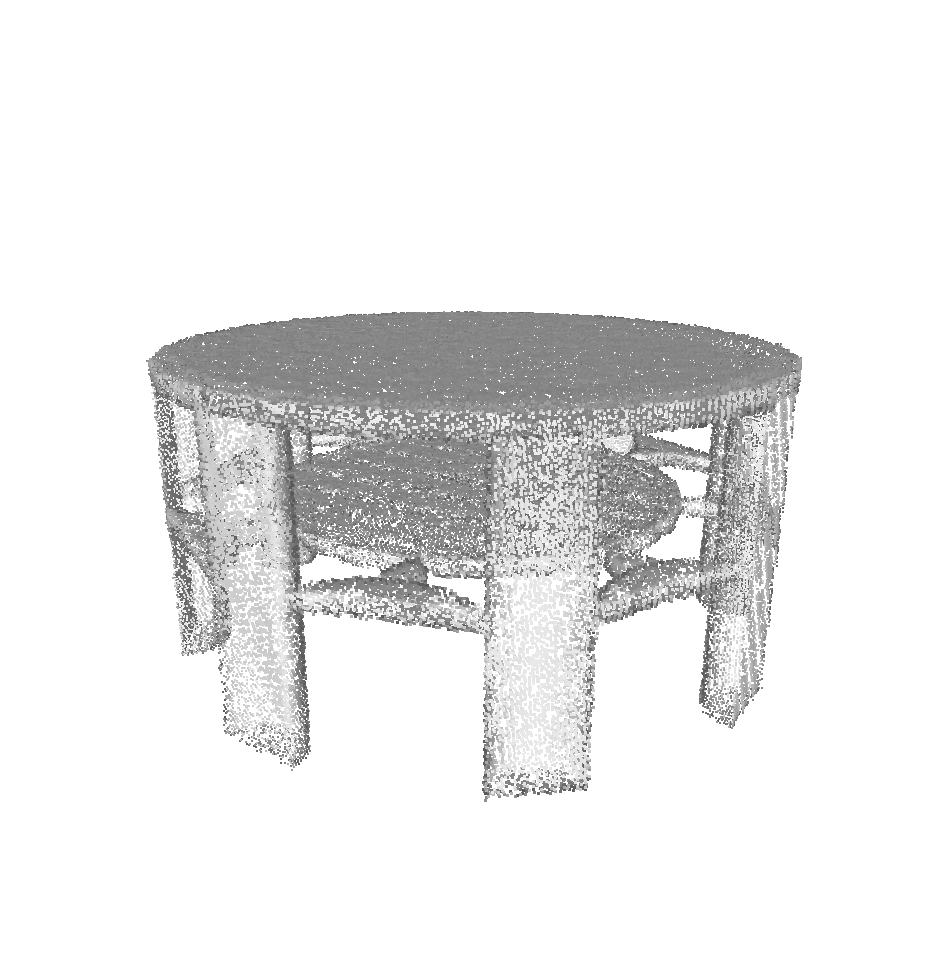}}%
\subfloat{\includegraphics[width=0.8in]{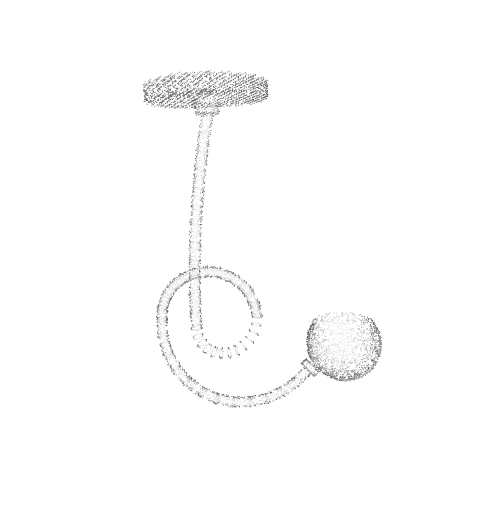}}%
\caption{Training examples from ShapeNet.}
\label{fig:shapenet}
\end{figure}

\textbf{Datasets.}
We randomly select $12,714$ 3D mesh models from the core dataset of ShapeNet~\cite{chang2015shapenet} for training, including 55 categories of common objects.  We sample the mesh model into point clouds by randomly generating points on the surfaces of the mesh.  To ensure the uniform  distribution of the points, we set the point density as $2\times10^{-5}$ when sampling each mesh surface.  Fig.~\ref{fig:shapenet} shows some examples of these point clouds used for training from ShapeNet.  These point clouds are then voxelized on a $ 256\times265\times256 $ occupancy space. We randomly collect $64\times64\times64$ cubes from each voxelized point cloud, resulting in $ 2\times10^{5} $ cubes in total used in this work.

\textbf{Strategy.} 
Loss function used for training is defined in Eq.~\eqref{loss}.  We set rate-distortion trade-off $\lambda$ from 0.75 to 16 to derive various models with different compression performance. 
During training, we first train the model at high bit rate by setting $\lambda$ to 16 and then use it to initialize model for lower bit rates. Applying the pre-trained model from higher bit rates for transfer learning not only ensures faster convergence but also guarantees reliable and stable outcomes. The learning rate is set to $10^{-5}$, and the batch size is set to 8.
Training iteration executes more than $2\times10^{5}$ batches for model derivation.   Here, we use the Adam~\cite{kingma2014adam} to optimize the proposed network. We set its parameters $\beta_{1}$ and $\beta_{2}$ to 0.9 and 0.999, respectively.

\subsection{Performance Evaluation}
We apply trained models to do tests, aiming to validate the efficiency of our proposed method in subsequent discussions.

\begin{figure}[t]
\centering
\subfloat[\emph{Loot}]{\includegraphics[width=0.8in]{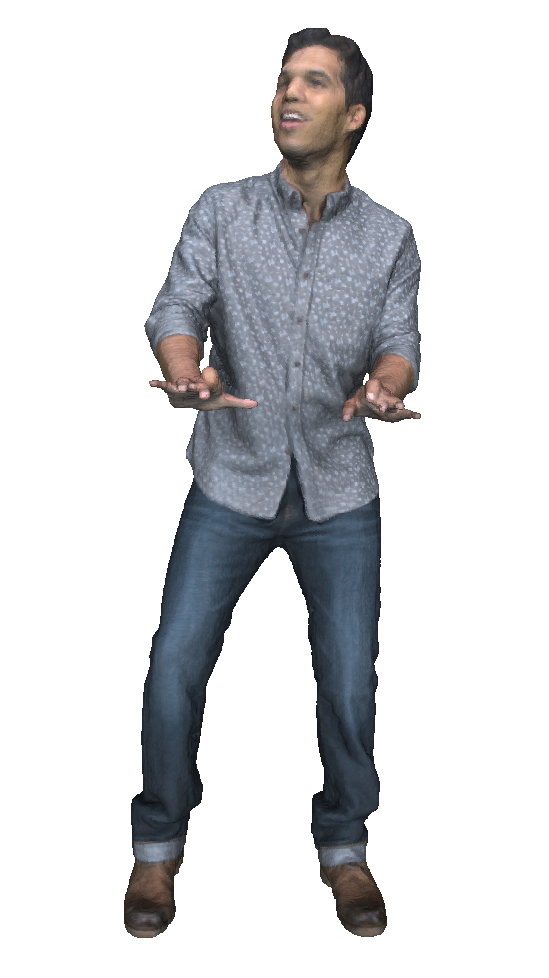}}
\subfloat[\emph{Redandblack}]{\includegraphics[width=0.8in]{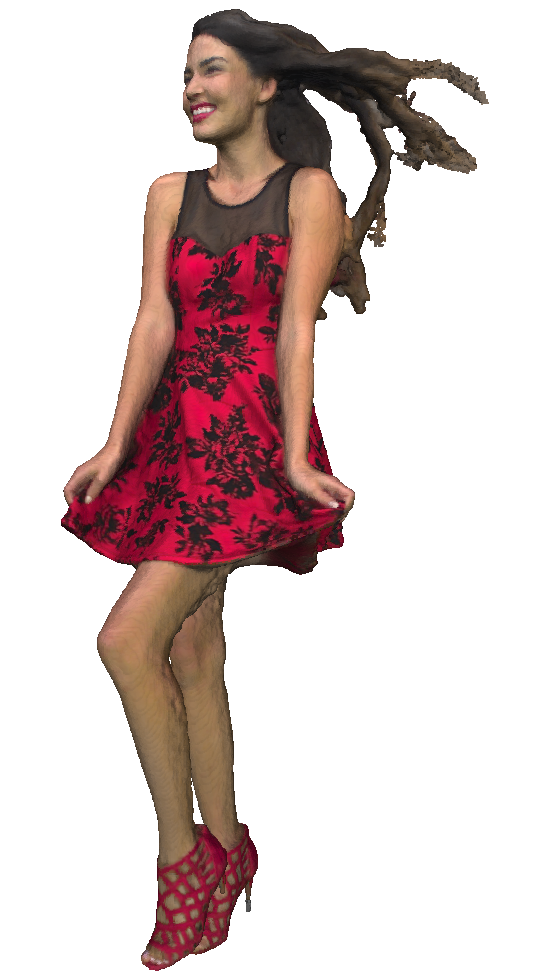}}%
\subfloat[\emph{Solider}]{\includegraphics[width=0.8in]{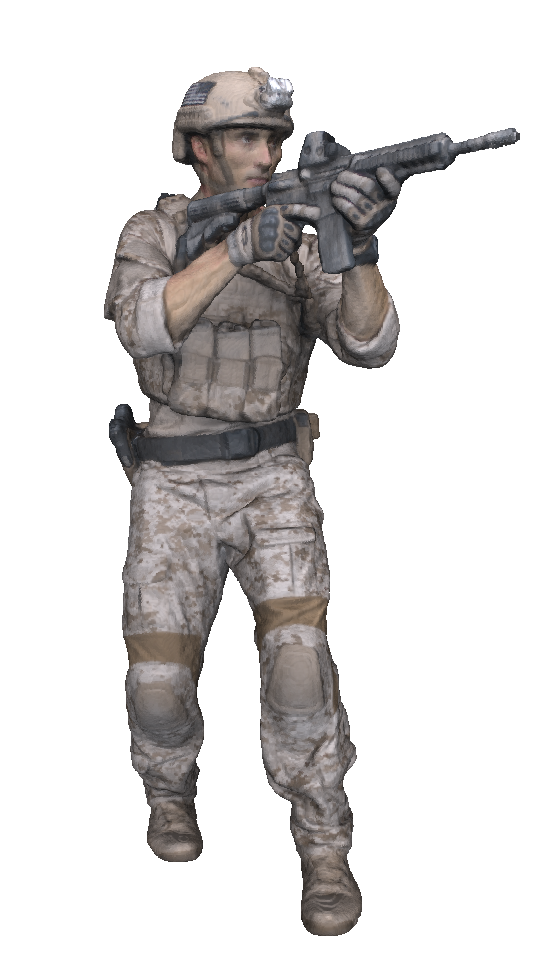}}%
\subfloat[\emph{Longdress}]{\includegraphics[width=0.8in]{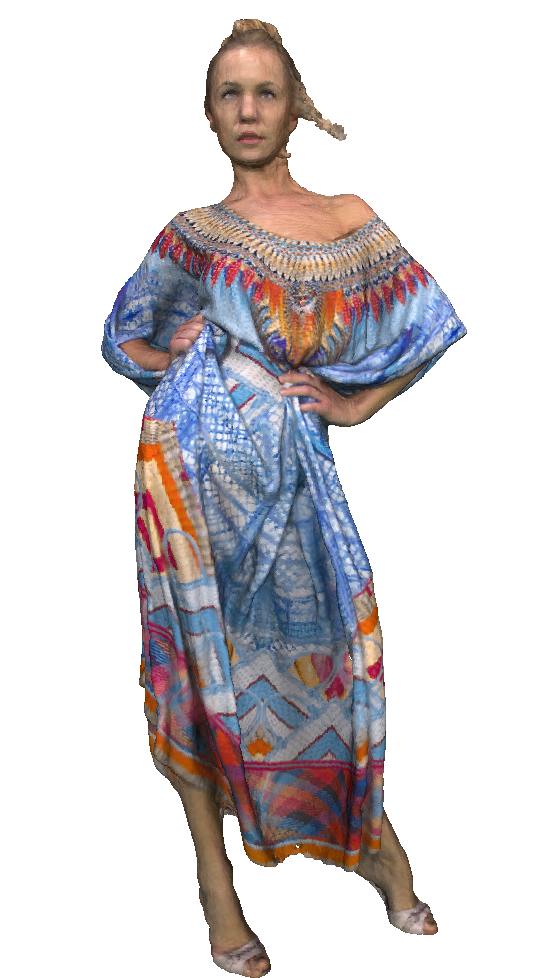}}\\%
\subfloat[\emph{Andrew}]{\includegraphics[width=0.8in]{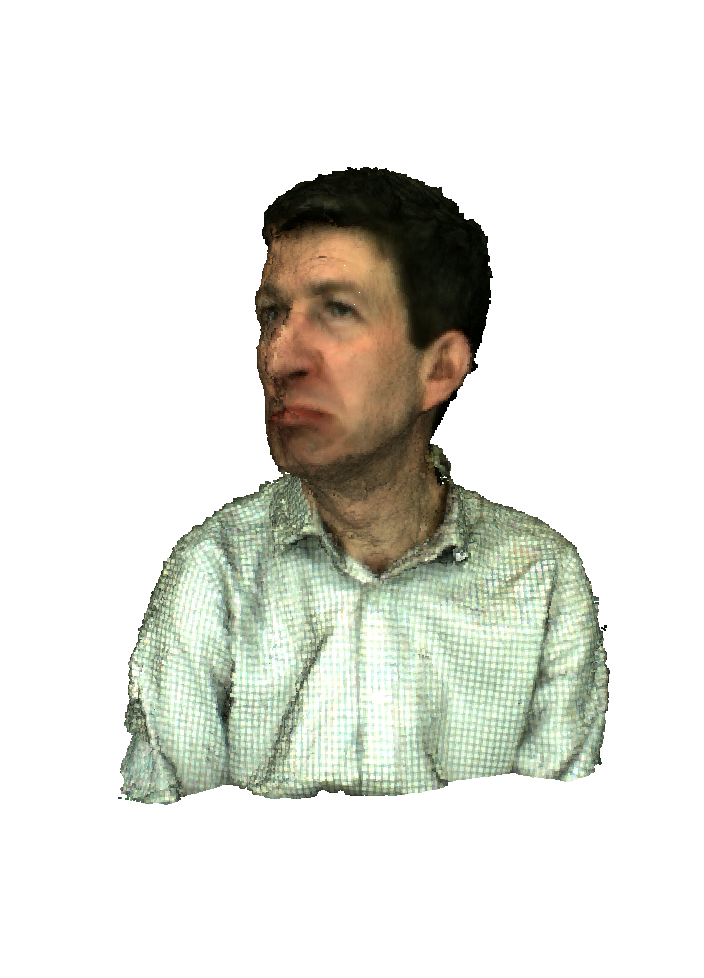}}
\subfloat[\emph{David}]{\includegraphics[width=0.8in]{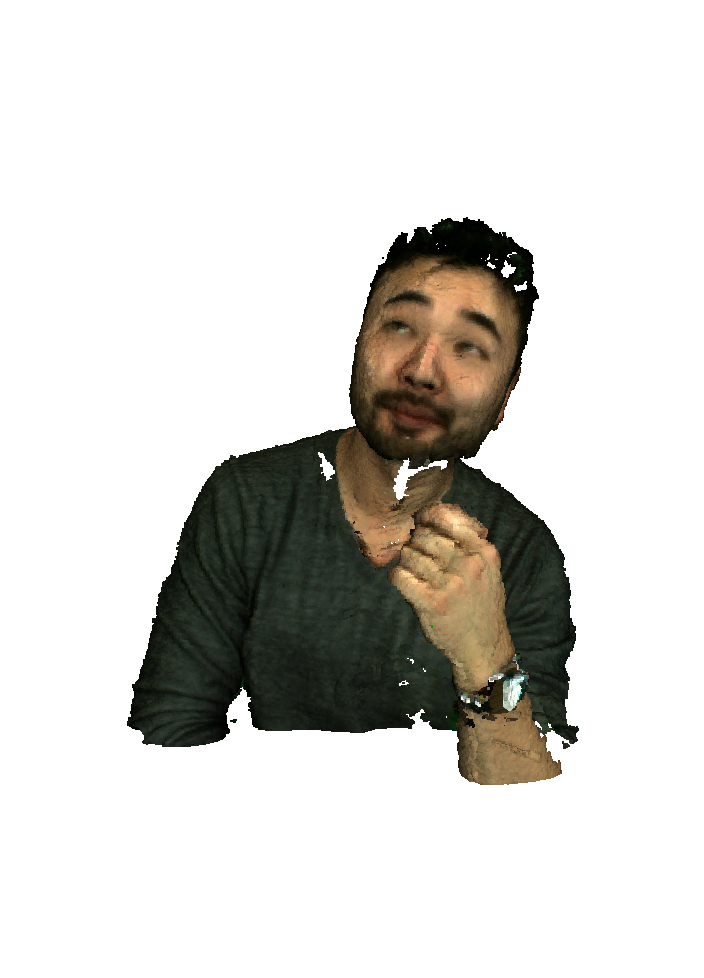}}%
\subfloat[\emph{Phil}]{\includegraphics[width=0.8in]{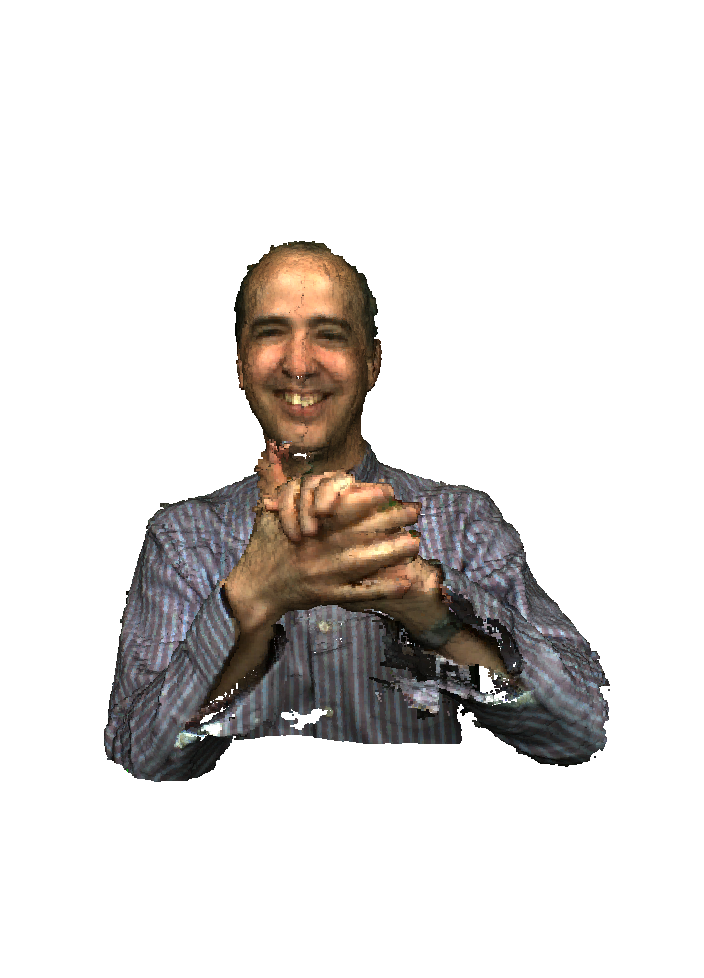}}%
\subfloat[\emph{Sarah}]{\includegraphics[width=0.8in]{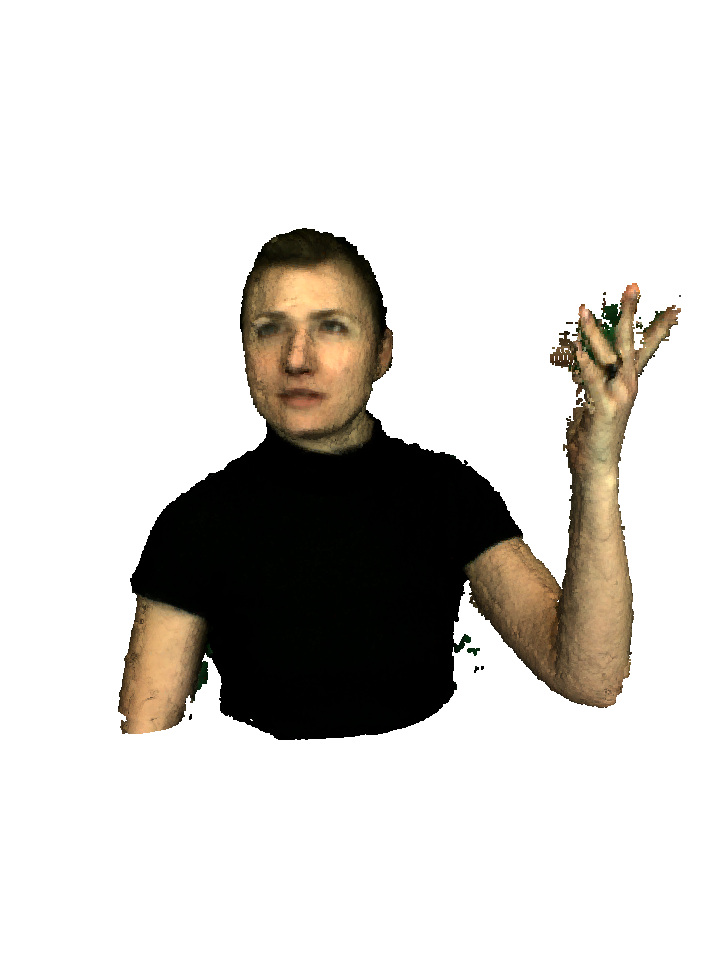}}\\%
\subfloat[\emph{Egyptian Mask}]{\includegraphics[width=1.1in]{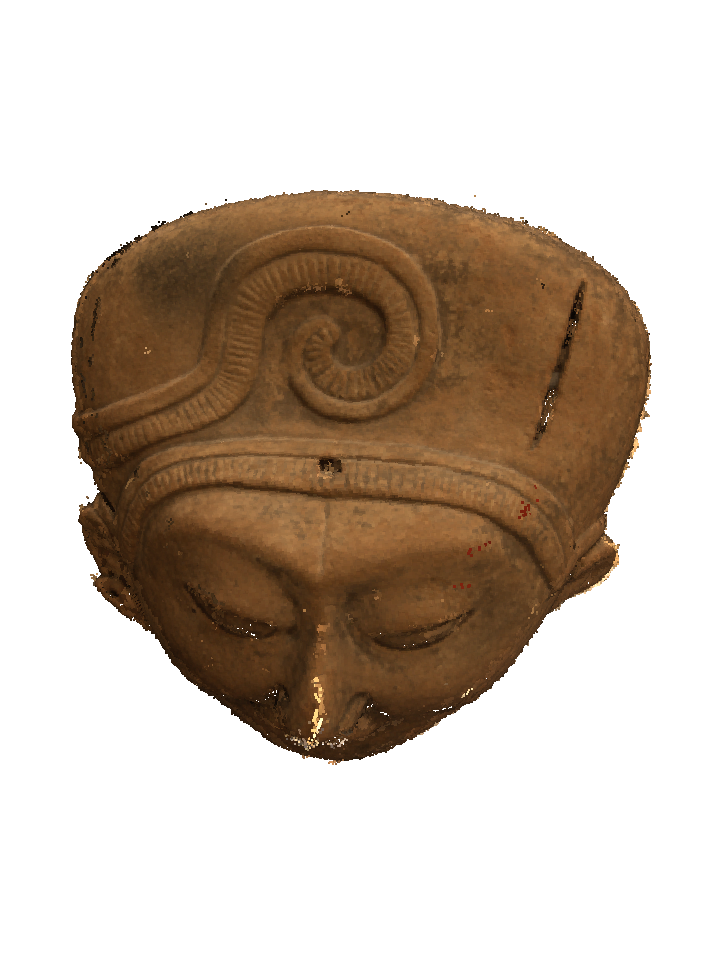}}
\subfloat[\emph{Statue Klimt}]{\includegraphics[width=1.1in]{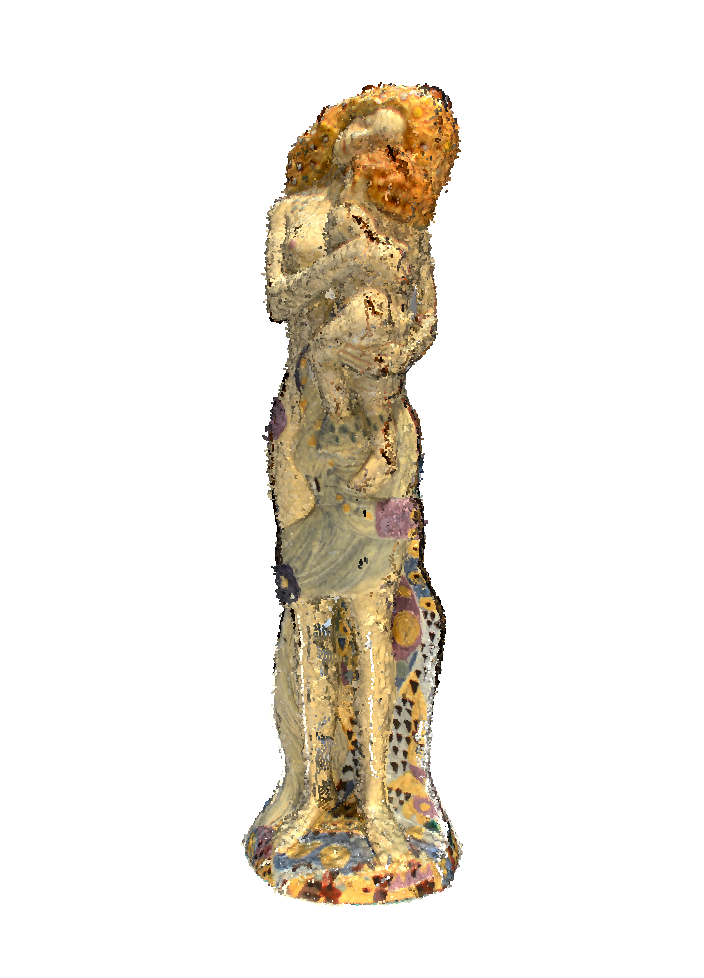}}%
\subfloat[\emph{Shiva}]{\includegraphics[width=1.1in]{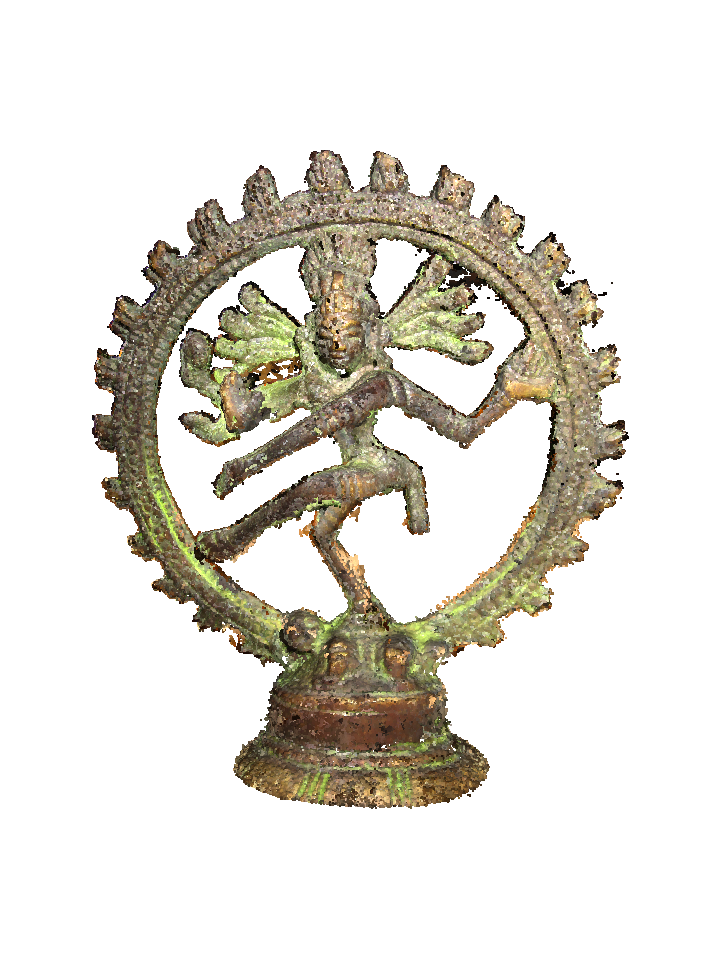}}%
\caption{{\bf  Testing Datasets.} Class A with full bodies shown in (a)-(d), Class C with inanimate objects for culture heritage in (i)-(k) used for MPEG PCC Common Test Condition (CTC)~\cite{Sebastian2018common}; Class B with half bodies in (e)-(h) used by JPEG Pleno~\cite{microsoft2019microsoft}. Note that Class C is static point cloud with one frame, and Class A and B are dynamic point clouds with multiple frames.}
\label{fig:testdata}
\end{figure}

\textbf{Testing Datasets.}
We choose three different test sets that are adopted by MPEG PCC~\cite{Sebastian2018common} and JPEG Pleno~\cite{microsoft2019microsoft} groups, to evaluate the performance of the proposed method, as shown in  Fig.~\ref{fig:testdata} and in Table~\ref{table:testdataset}.  
These testing datasets present different structures and properties.
Specifically, Class A (full bodies) exhibits smooth surface and complete shape, while Class B (upper bodies) presents noisy and incomplete surface (even having visible holes and missing parts). Another three inanimate objects in Class C have higher geometry precision but more sparse voxel distribution. Frames in ~\ref{table:testdataset} used for evaluation are also suggested by the MPEG PCC group.

\begin{table}[t]
\renewcommand{\arraystretch}{1.3}
\caption{Details for Testing Datasets}
\label{table:testdataset}
\centering
\begin{tabular}{|c|c|c|c|c|}
\hline
 \multicolumn{2}{|c|}{Point Cloud}& Points\# & Precision  & Frame\# \\
\hline
\multirow{4}{*}{A}&Loot        & 805285 &	10 & 1200 \\
\cline{2-5}
&Redandblack                  & 757691 &	10 & 1550 \\
\cline{2-5}
&Soldier                      & 1089091 & 10 & 690 \\
\cline{2-5}
&Longdress                    & 857966 &	10 & 1300 \\
\hline
\multirow{4}{*}{B}&Andrew & 279664 & 9 & 1\\ 
\cline{2-5}
&David & 330791 & 9 & 1\\
\cline{2-5}
&Phil & 370798 & 9 & 1\\
\cline{2-5}
&Sarah & 302437 & 9 & 1\\
\hline
\multirow{3}{*}{C}&Egyptian Mask & 272684 & 12 & - \\
\cline{2-5}
& Statue Klimt & 499660 & 12 & - \\
\cline{2-5}
&Shiva & 1009132 & 12 & - \\
\hline
\end{tabular}
\end{table}

\begin{figure}[t]
\centering
\subfloat{\includegraphics[width=1.72in]{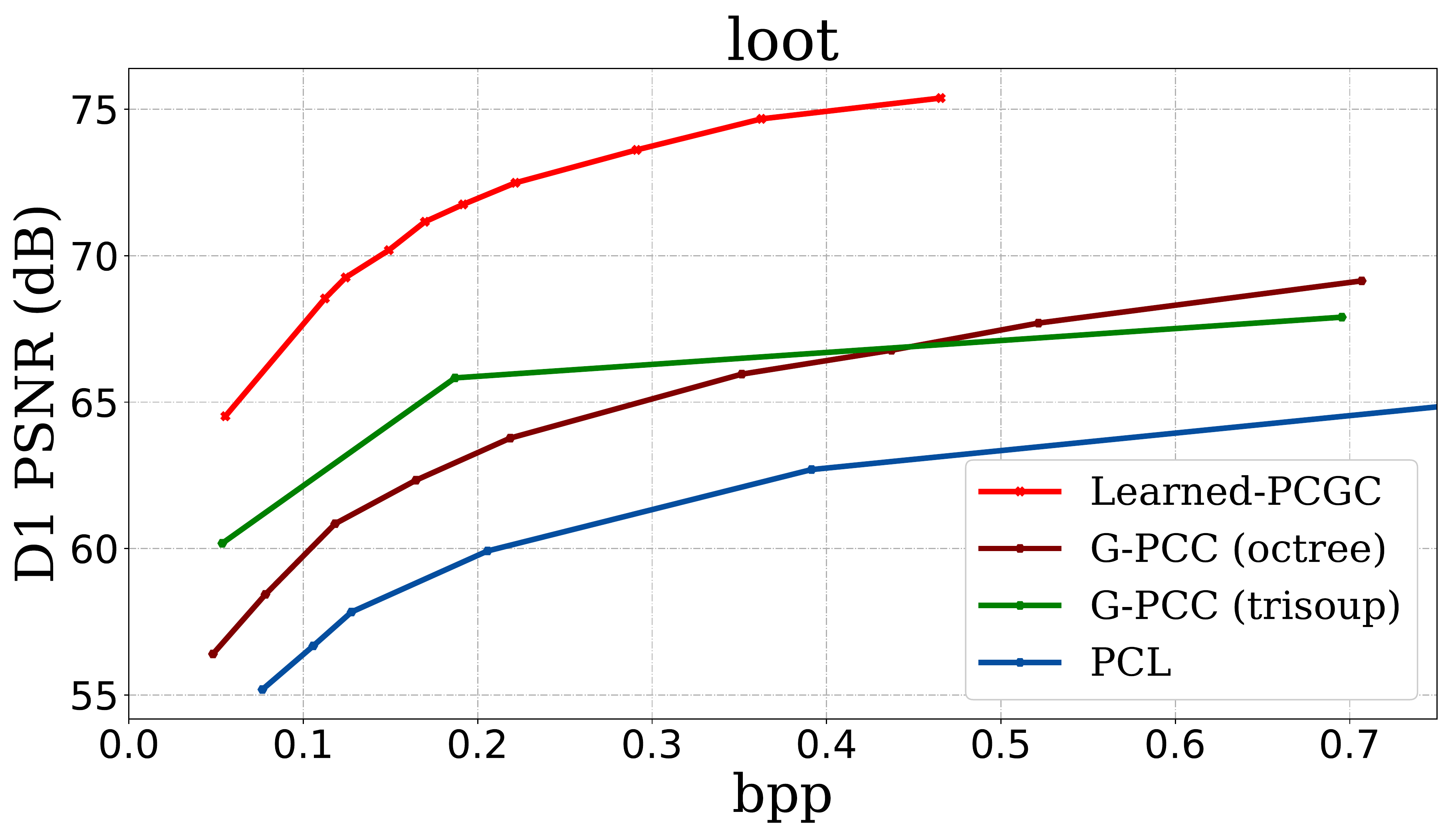}}
\subfloat{\includegraphics[width=1.72in]{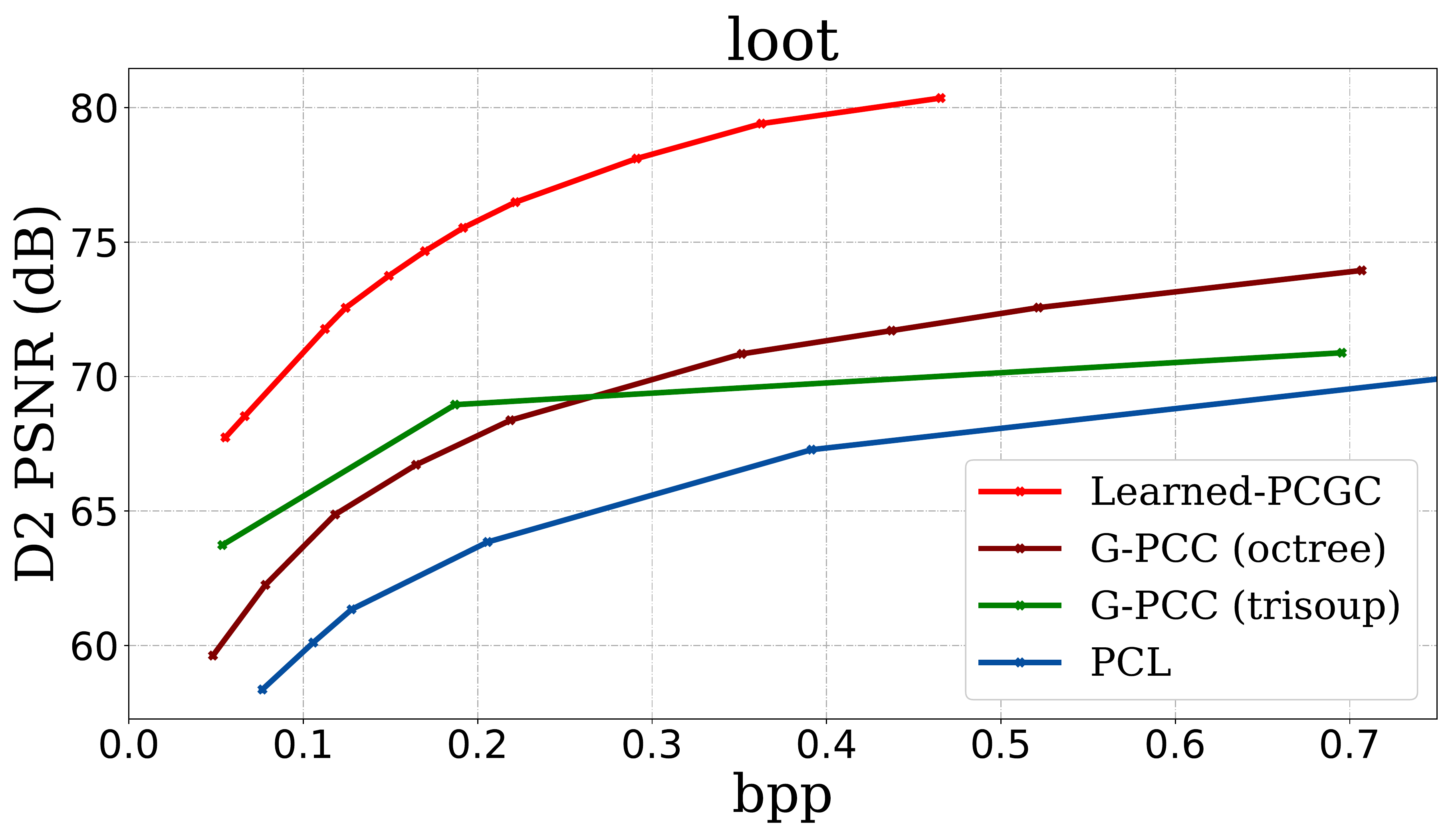}}\\%
\subfloat{\includegraphics[width=1.72in]{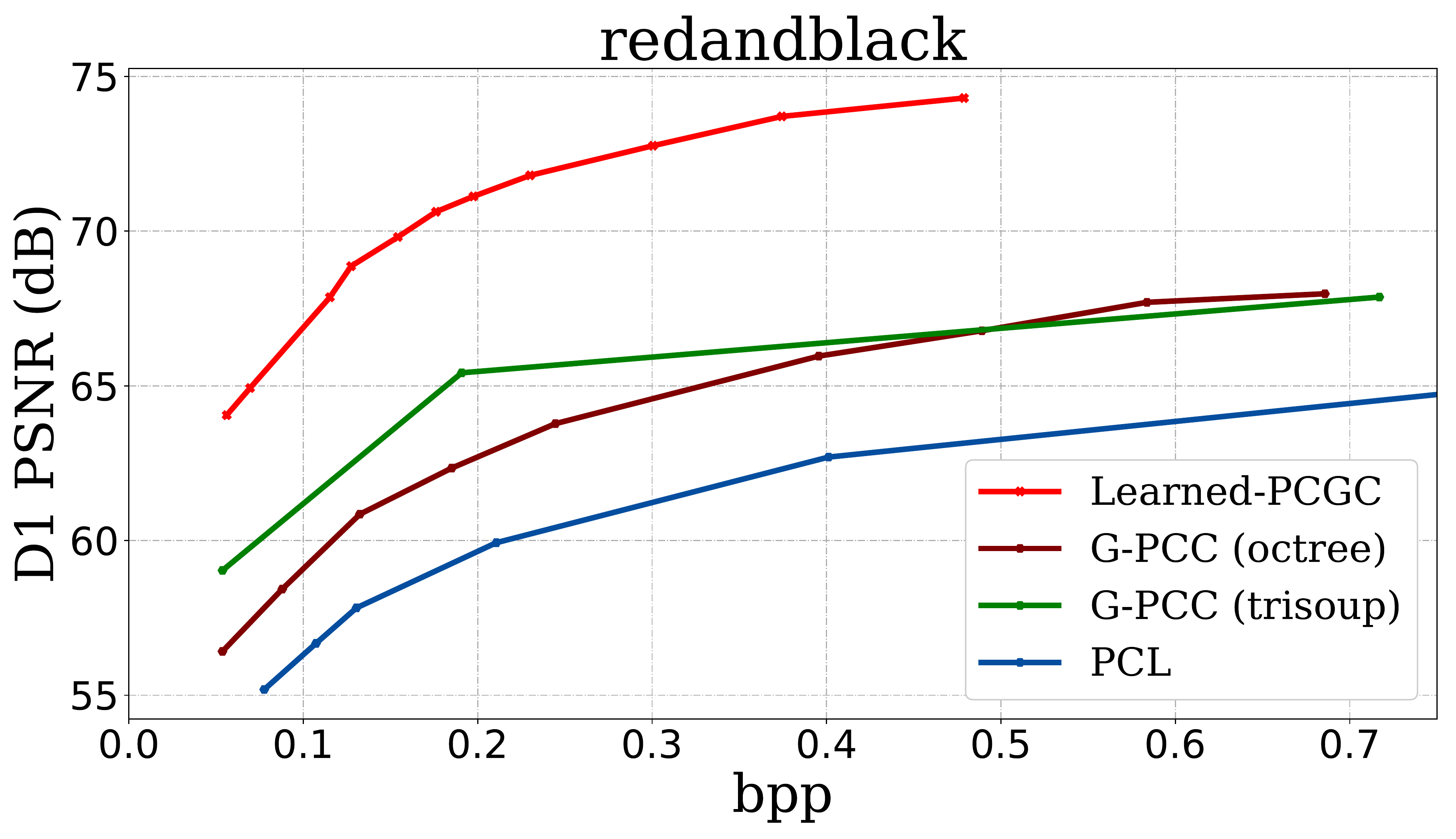}}%
\subfloat{\includegraphics[width=1.72in]{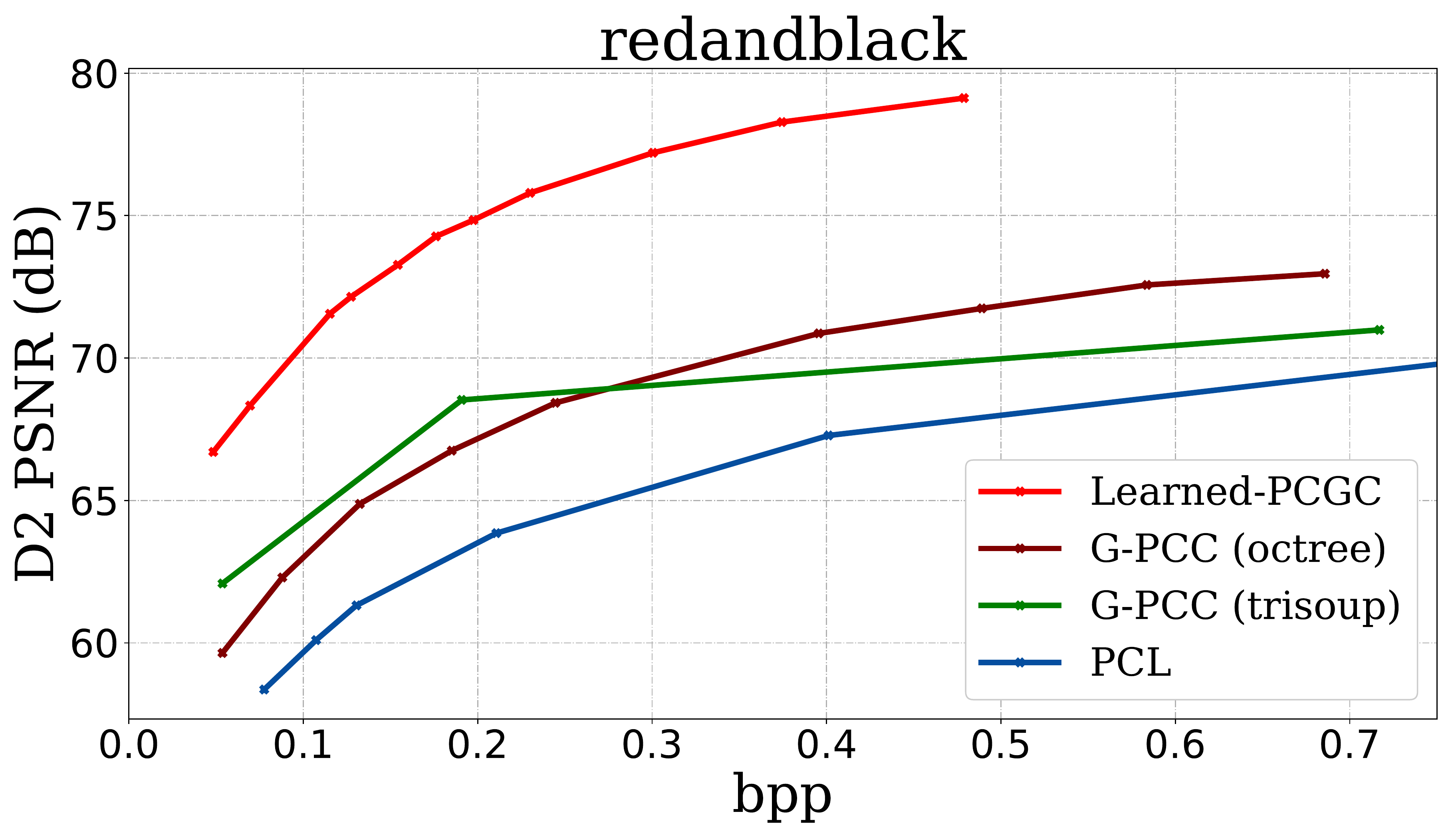}}\\%
\subfloat{\includegraphics[width=1.72in]{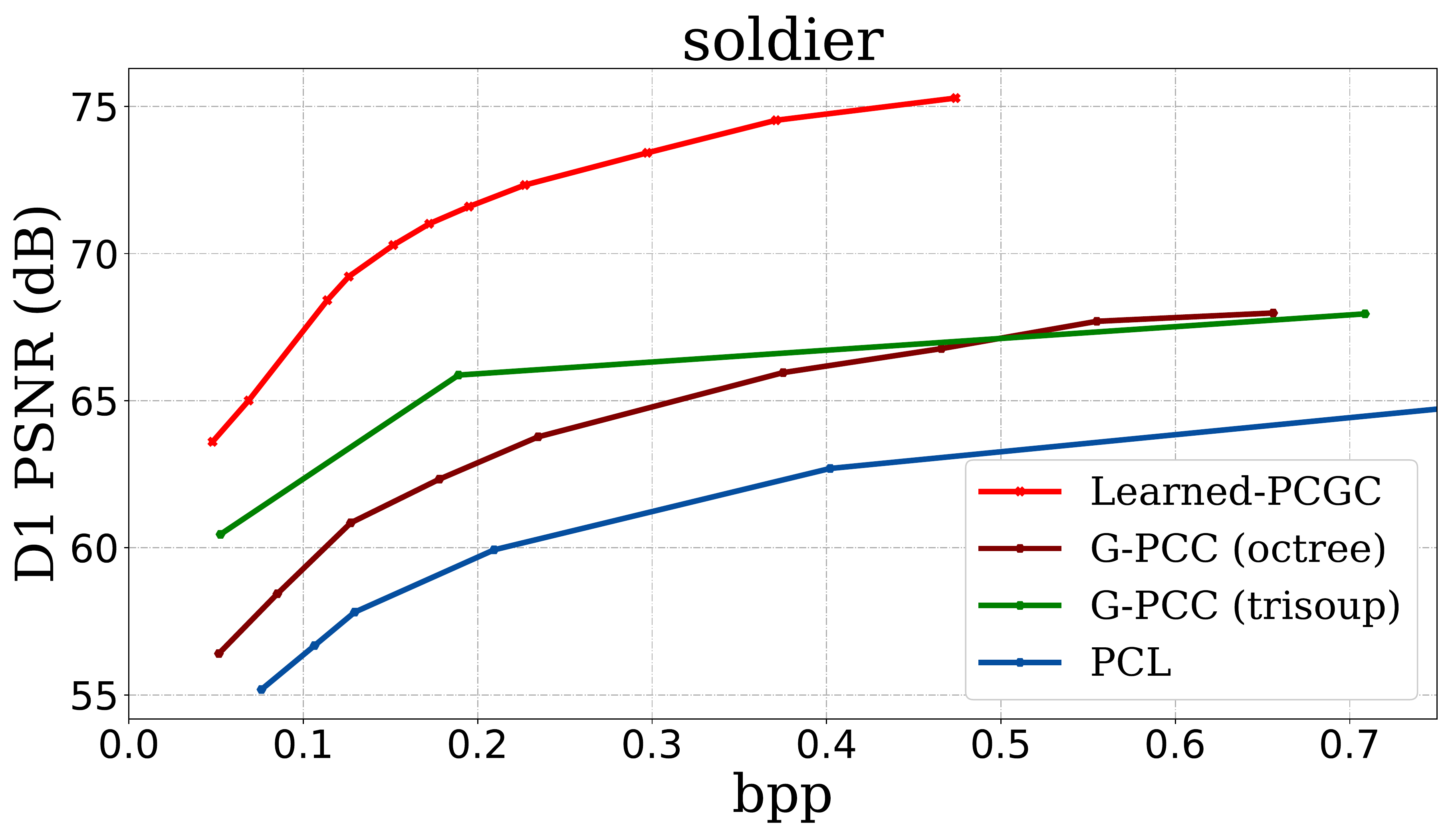}}%
\subfloat{\includegraphics[width=1.72in]{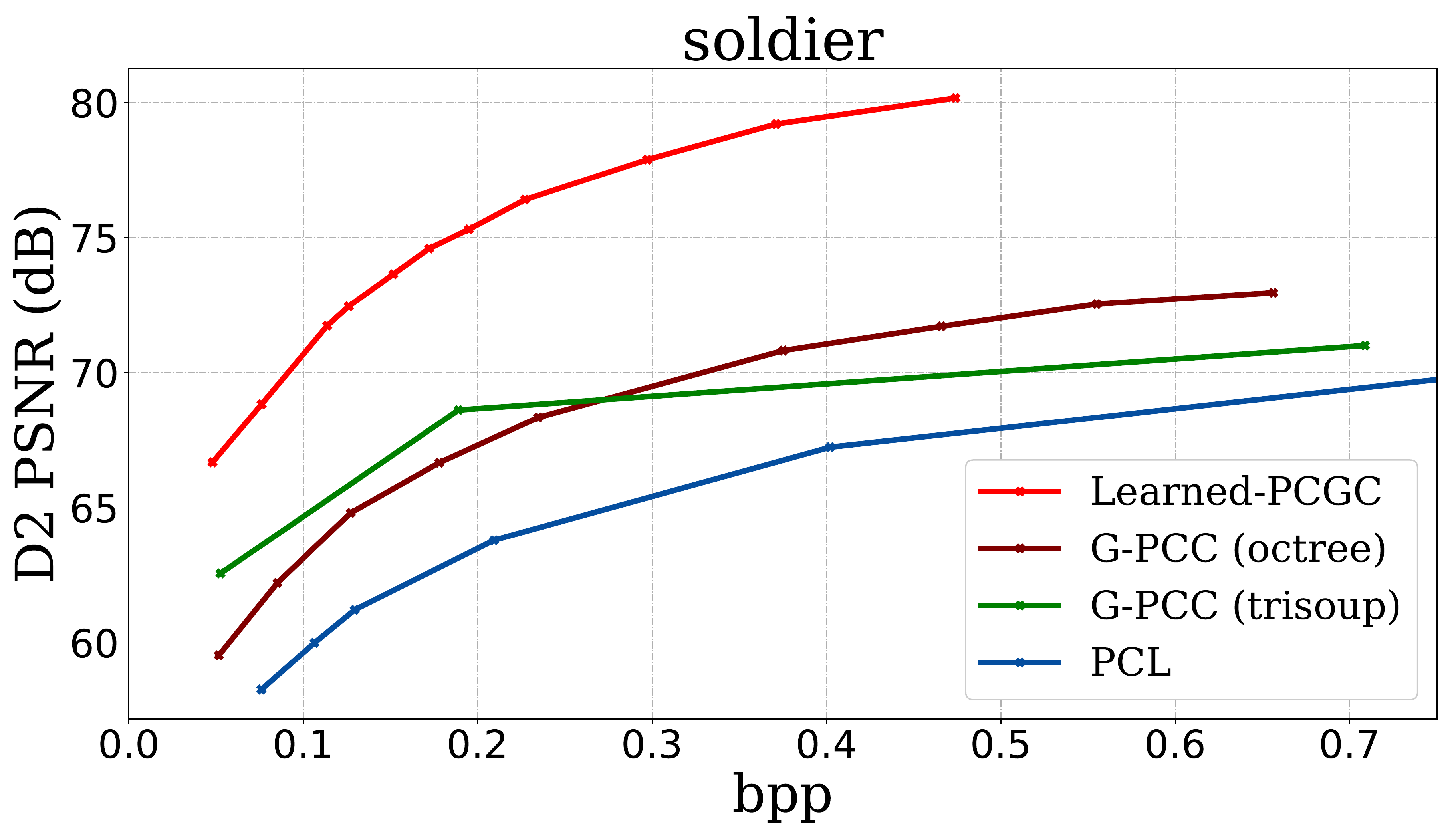}}\\%
\subfloat{\includegraphics[width=1.72in]{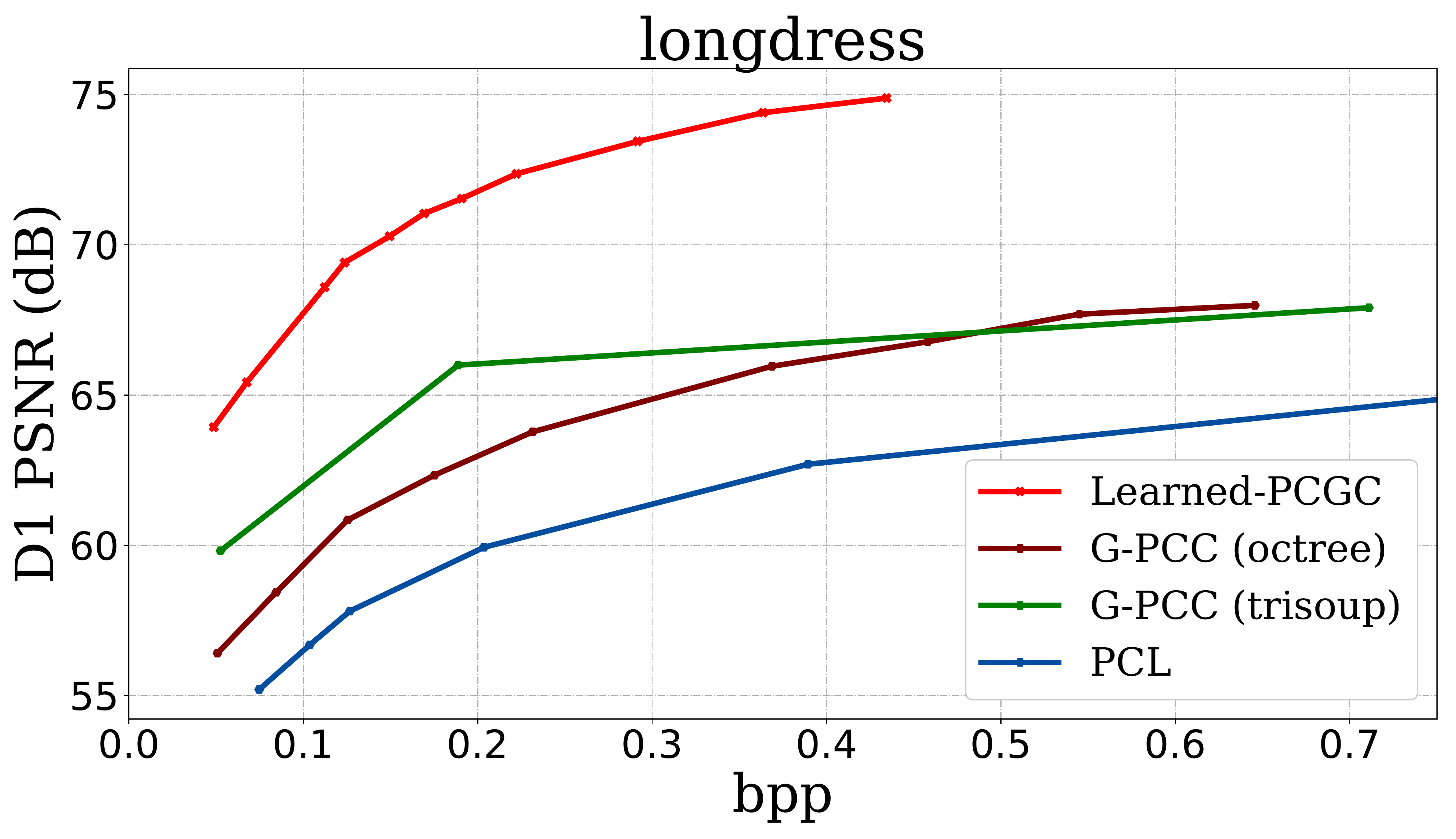}}%
\subfloat{\includegraphics[width=1.72in]{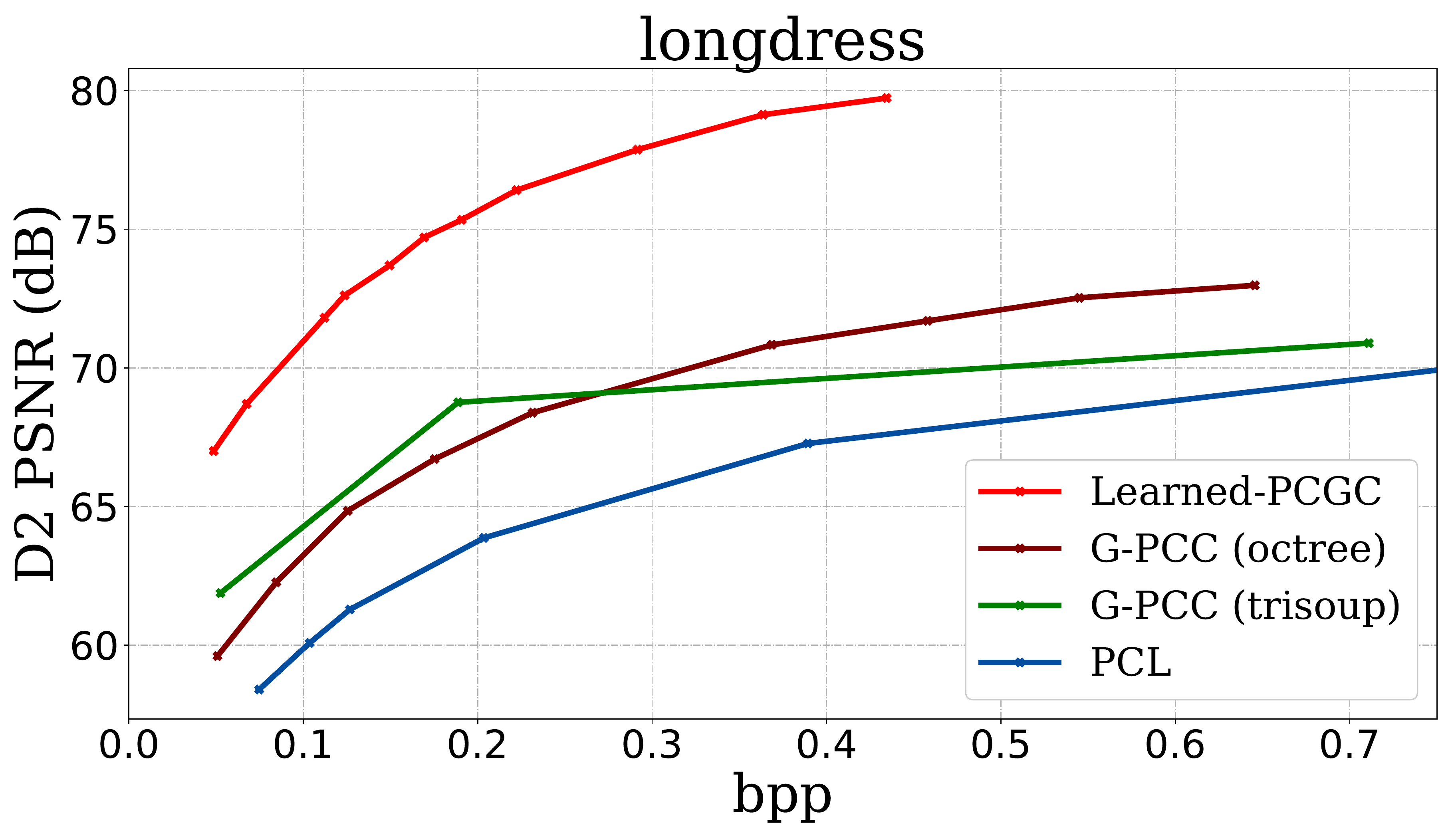}}%
\caption{R-D curves of Class A point clouds for PCL, G-PCC (octree), G-PCC (trisoup) and our Learned-PCGC: (left) D1 based PSNR, (right) D2 based PSNR.}
\label{rdcurve8i}
\end{figure}

{\bf Objective Comparison.} 
We mainly compare our method with other PCGC algorithms, including 1) octree-based codec in Point Cloud Library (PCL)~\cite{PCL2011}; 2) MPEG PCC test model (TMs): TM13 for category 1 (static point cloud data), a.k.a., G-PCC; and 3) MPEG PCC TM2 for category 2 (dynamic content), a.k.a., V-PCC.
Geometry model can be different in G-PCC method using respective \emph{octree}  or \emph{trisoup} representation. The former one is using the octree coding mechanism similar to the implementation in PCL, and the latter is based on triangle soup representation of the geometry. They are noted as G-PCC (octree) and G-PCC (trisoup), respectively.

For a fair comparison, we have tried to enforce the similar bit rate ranges for PCL, G-PCC (octree), G-PCC (trisoup) and our method. Such bit rate range is applied as suggested by the MPEG PCC Common Test Condition (CTC)~\cite{Sebastian2018common}.
\begin{itemize}
\item For PCL, we use the \emph{OctreePointCloudCompression} approach in PCL-v1.8.0~\cite{PCL2011} for geometry compression only. We set octree resolution parameters from 1 to 64 to obtain serial rate points. 
\item For G-PCC, the latest TM13-v6.0~\cite{tmc13} is used with parameter settings following the CTC~\cite{Sebastian2018common}. 
For G-PCC (octree), we set \emph{positionQuantizationScale} from 0.75 to 0.015, leaving other parameters as default. 
For G-PCC (trisoup), we set \emph{tirsoup\_node\_size\_log2}  to 2, 3, 4, and \emph{positionQuantizationScale}  to 1 for Class A and B, and 0.125 or 0.25 for Class C\footnote{Downscaling is applied for Class C point cloud because they are typically sparse but with higher precision. }. 
\item Our Learned-PCC is trained in an end-to-end fashion for individual bit rates by adapting $\lambda$ and scaling factor $s$.
\end{itemize}

\begin{table*}[t]
\renewcommand{\arraystretch}{1.3}
\caption{BD-Rate Gains against PCL, G-PCC (cotree), G-PCC (trisoup) in D1 and D2 based BR-Rate Measurement.}
\label{BDBRG-PCC}
\centering
\begin{tabular}{|c|c|c|c|c||c|c|c|}
\hline
\multicolumn{2}{|c|}{\multirow{2}{*}{Point Cloud} }& \multicolumn{3}{c||}{D1 (p2point)} &\multicolumn{3}{c|}{D2 (p2plane)} \\   \cline{3-5} \cline{6-8}
                        \multicolumn{2}{|c|}{} & PCL & G-PCC (octree) & G-PCC (trisoup) & PCL & G-PCC (octree) & G-PCC (trisoup) \\ 
\hline
\multirow{5}{*}{A}& Loot                         & -91.50 & 	-80.30  & -68.58 &  -87.50 &	-73.49	& -68.91     \\ \cline{2-8}
&Redandblack                  &  -90.48	  &     -79.47  & -68.10 & -86.70 &	-73.33 &	-68.22    \\ \cline{2-8}
&Soldier                      &  -90.93 &	-79.67  &	-62.14  &  -87.07 &	-73.08 &	-67.39    \\ \cline{2-8}
&Longdress                    &  -91.22 &	-80.46 &	-62.97 &	-87.34 &	-74.09 &	-68.35   \\ \cline{2-8}
&\textbf{Average}            & \textbf{-91.03} & \textbf{-79.98} & \textbf{-65.44} & \textbf{-87.15} &	\textbf{-73.49}  & \textbf{-68.21} \\ 
\hline
\multirow{5}{*}{B}& Andrew                       &  -88.64 &	-77.57 &	-74.63  &-81.61&	-66.79&	-65.23 \\ \cline{2-8}
&David                        & -87.56&	-75.25&	-72.23  & -82.52 &	-68.13 &	-66.95\\ \cline{2-8}
& Phil                        &   -88.31&	-77.72&	-75.42 & -82.02	&-68.74&	-66.33 \\ \cline{2-8}
& Sarah                       &  -88.62 &	-76.91 &	-79.42     &  -83.36&	-69.51&	-72.61   \\ \cline{2-8}
& \textbf{Average}          & \textbf{-88.28} &	\textbf{-76.86} &	\textbf{-75.42} &  \textbf{-83.30} &	\textbf{-68.29} &	\textbf{-67.78} \\ 
\hline
\multirow{4}{*}{C}& Egyptian Mask                &-84.31 &	-73.53 &	-50.14  & -85.12 &	-74.02 &	-40.80 \\ \cline{2-8}
&Statue Klimt                  &--83.45 &	-75.89 &	-60.33  &  -74.66 &	-62.33 &	-47.56\\ \cline{2-8}
&Shiva                        & -77.30 &	-68.92 &	-64.91  & --67.89 &	-56.42 &	-51.85\\ \cline{2-8}
&\textbf{Average}             &   \textbf{-81.68} &	\textbf{-72.78}&	\textbf{-58.46} & \textbf{-75.89} &	\textbf{-64.25} &	\textbf{-46.73} \\ 
\hline
\multicolumn{2}{|c|}{\textbf{Overall average}}     &  \textbf{-87.48} &	\textbf{-76.88} &  \textbf{-67.17} & \textbf{-82.34} & \textbf{-69.08} & \textbf{-62.20}  \\ 
\hline
\end{tabular}
\end{table*}

Objective comparison is evaluated using the BD-Rate, shown in Table~\ref{BDBRG-PCC}. There are two distortion metrics widely used for point cloud geometry compression. One is the mean-squared-error (MSE) with point-to-point (D1-p2point) distance, and the other is the MSE with point-to-plane (D2-p2plane) distance measurement~\cite{tian2017geometric, mekuria2017performance}. 
Bit rate is represented using bits per input point (bpp), or bits per occupied voxel (bpov).

\begin{figure}[t]
\centering
\subfloat{\includegraphics[width=1.72in]{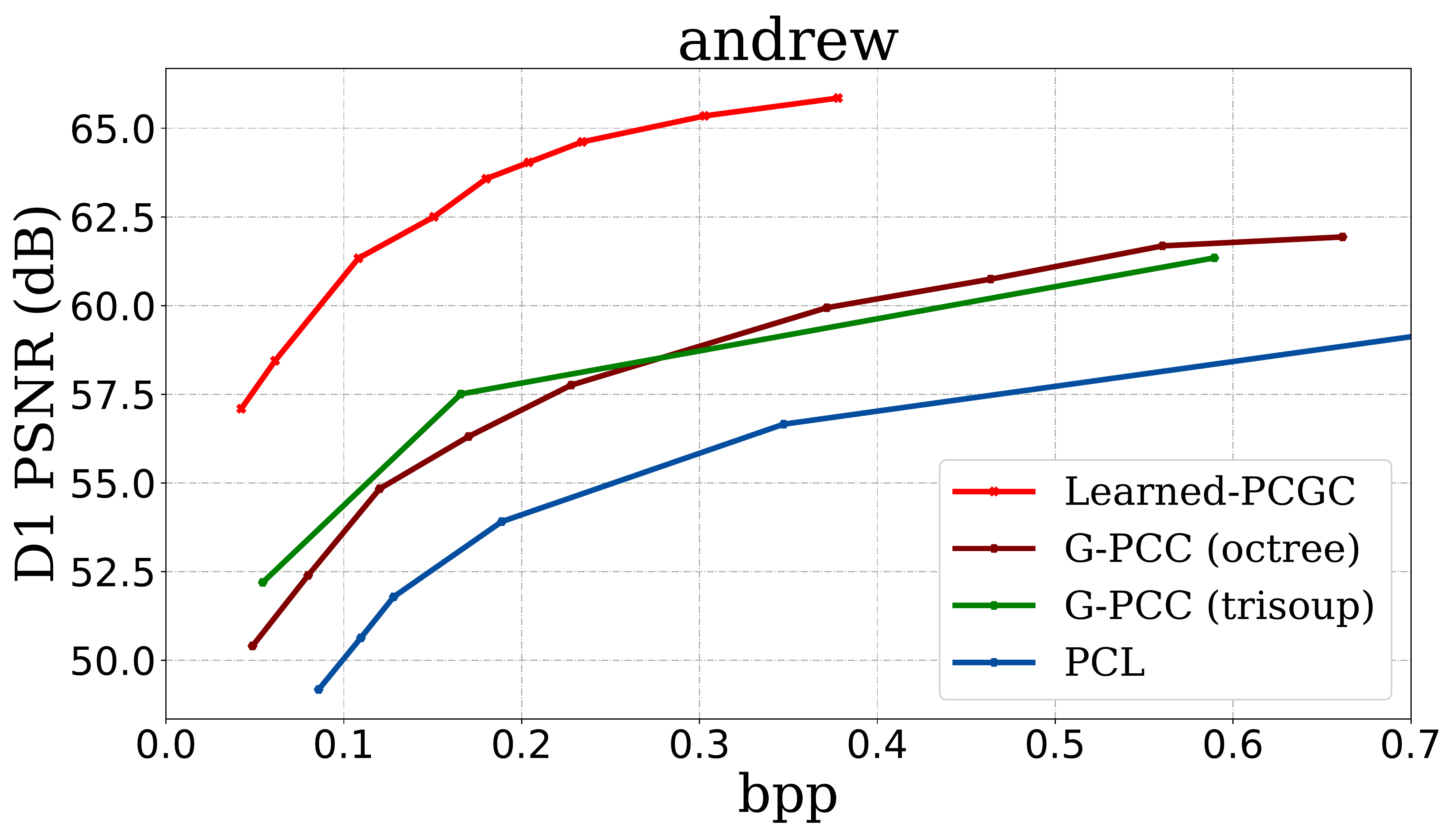}}%
\subfloat{\includegraphics[width=1.72in]{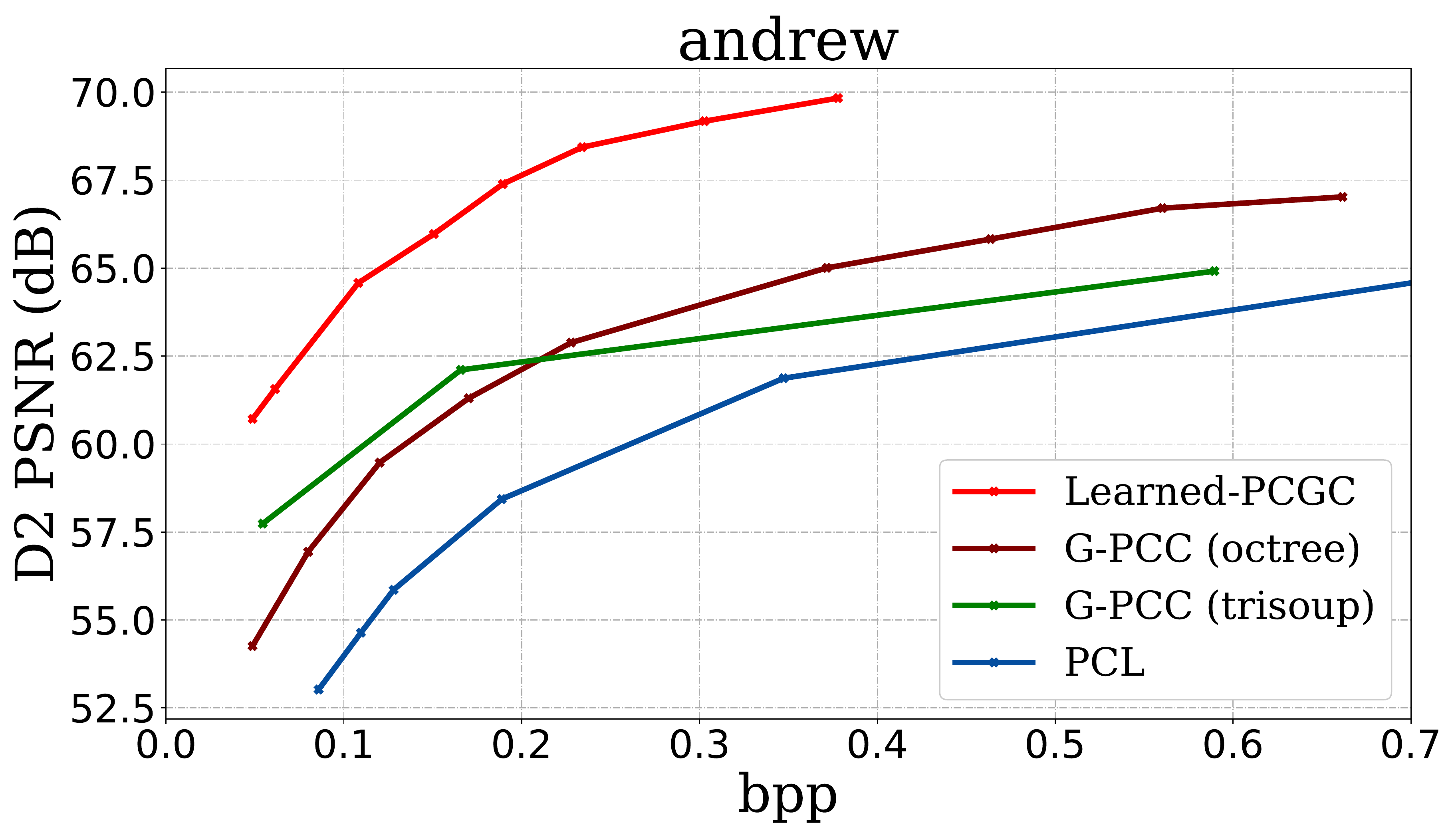}}\\%
\subfloat{\includegraphics[width=1.72in]{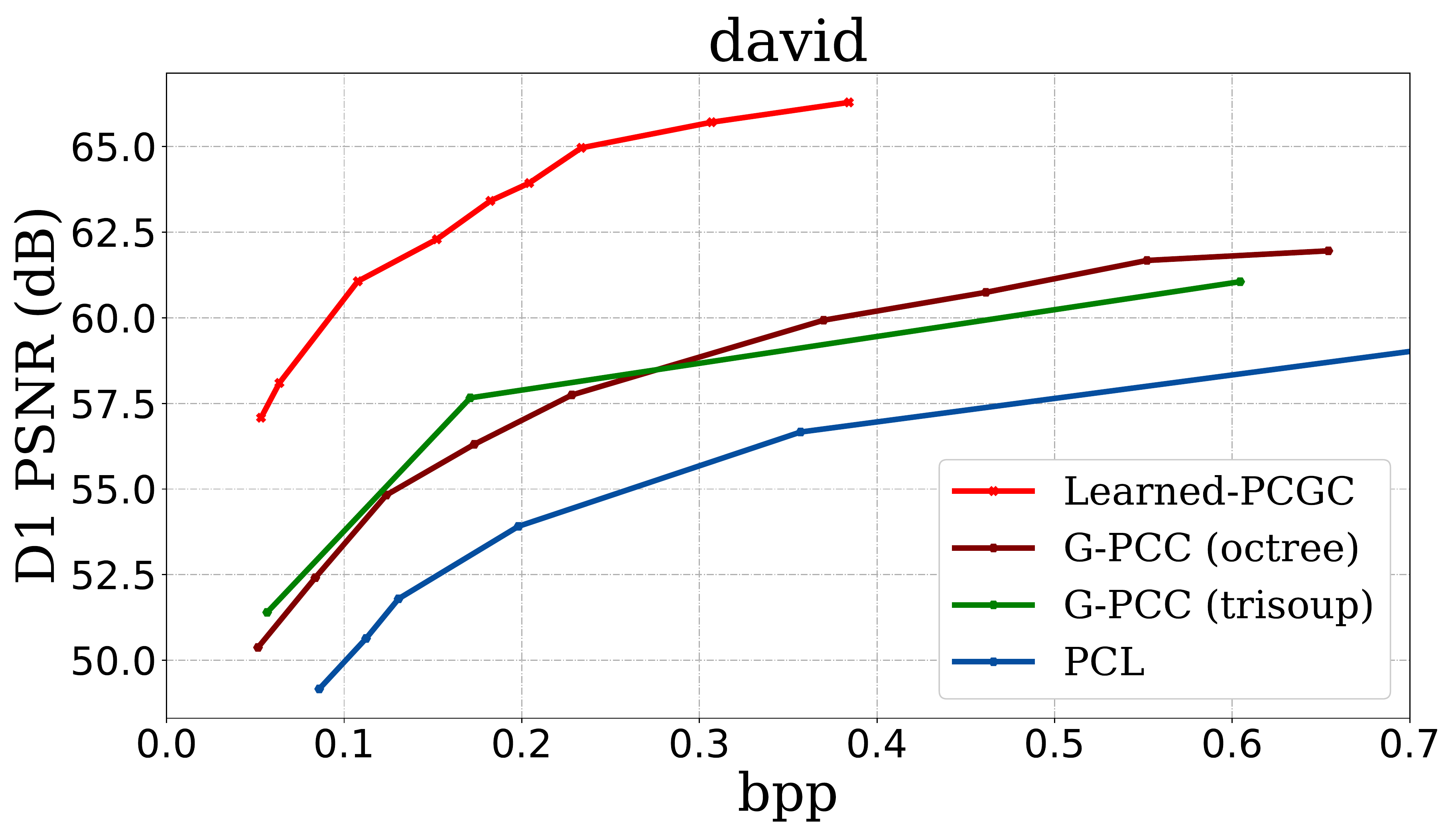}}%
\subfloat{\includegraphics[width=1.72in]{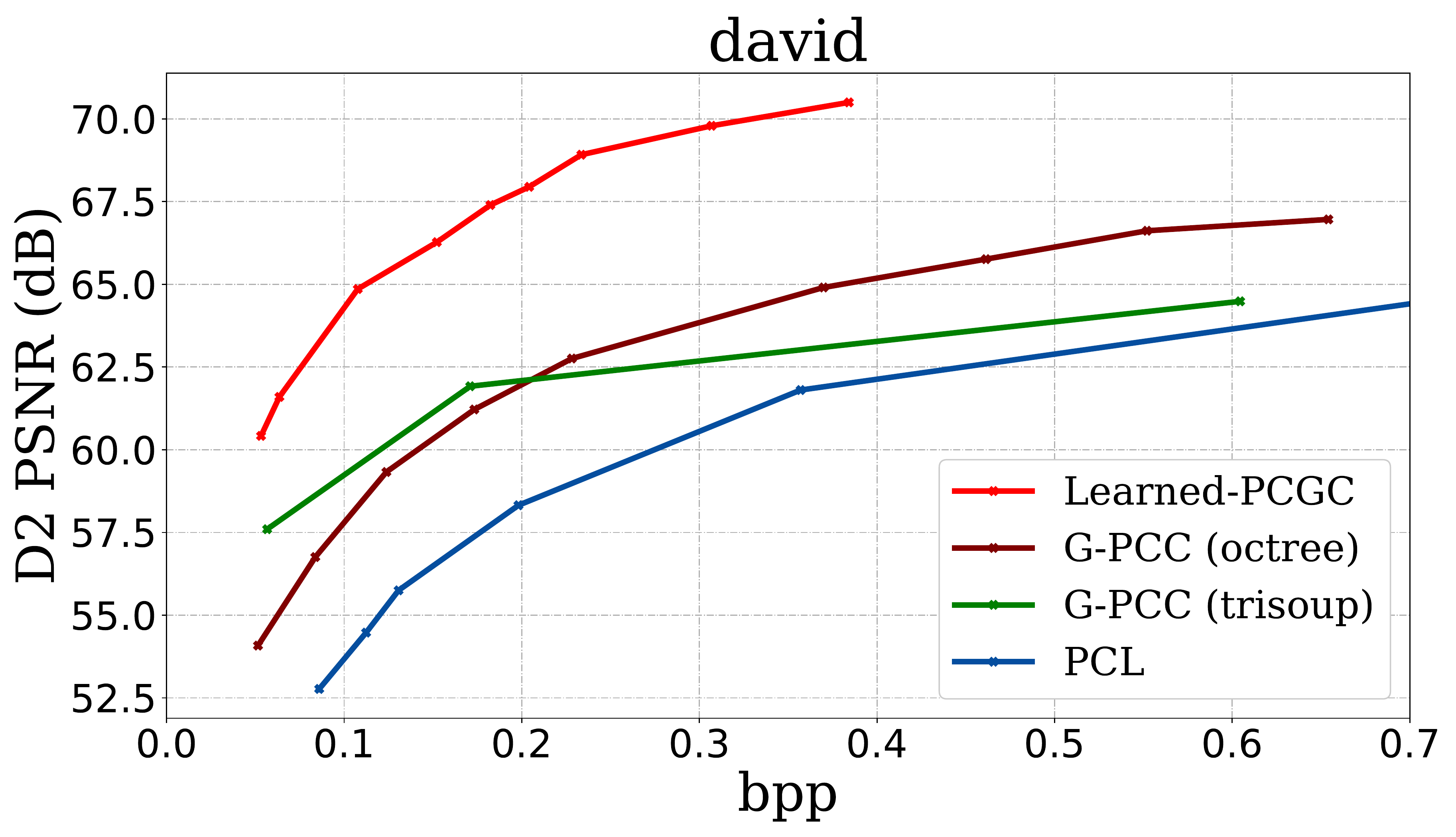}}\\%
\subfloat{\includegraphics[width=1.72in]{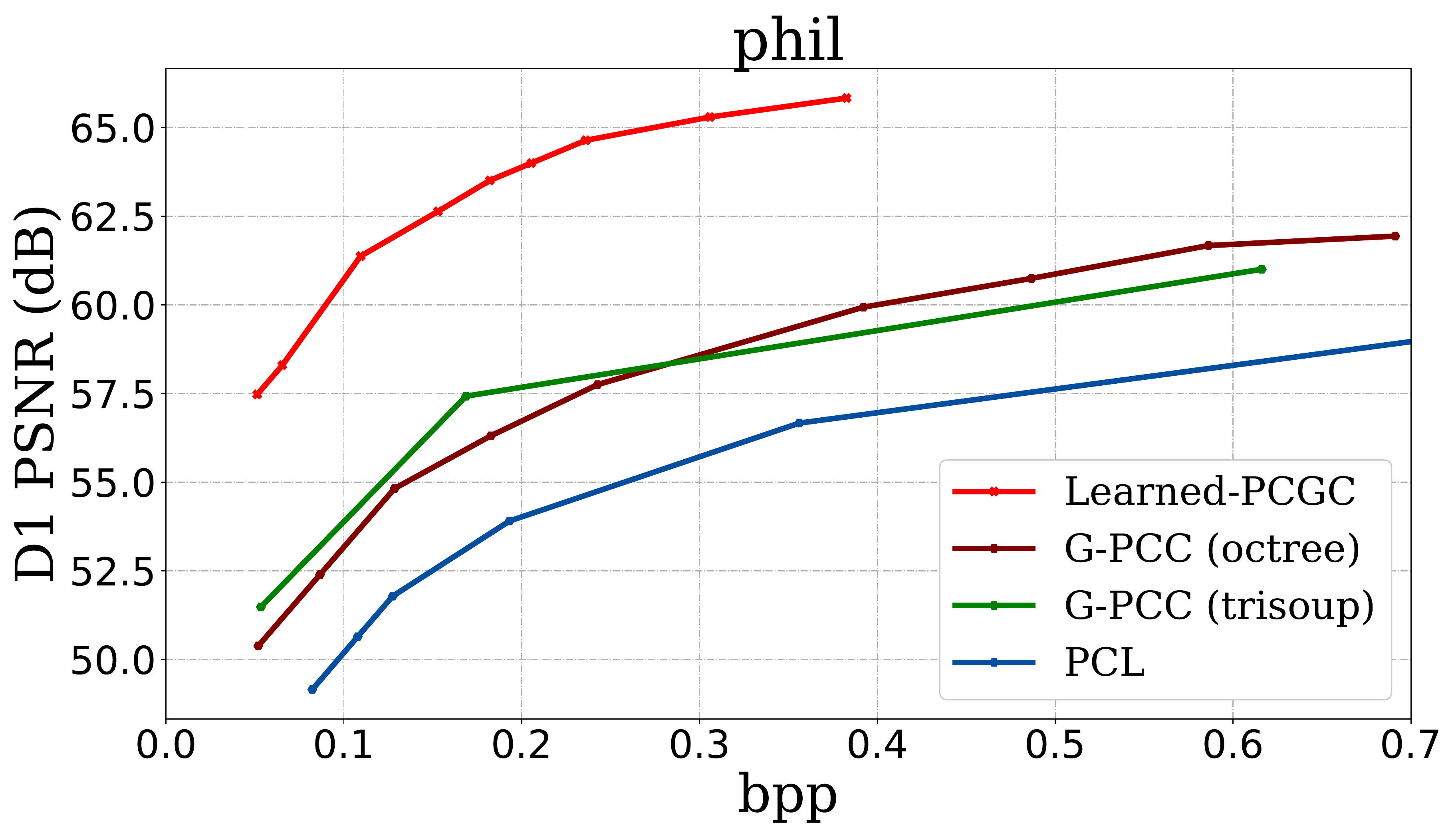}}%
\subfloat{\includegraphics[width=1.72in]{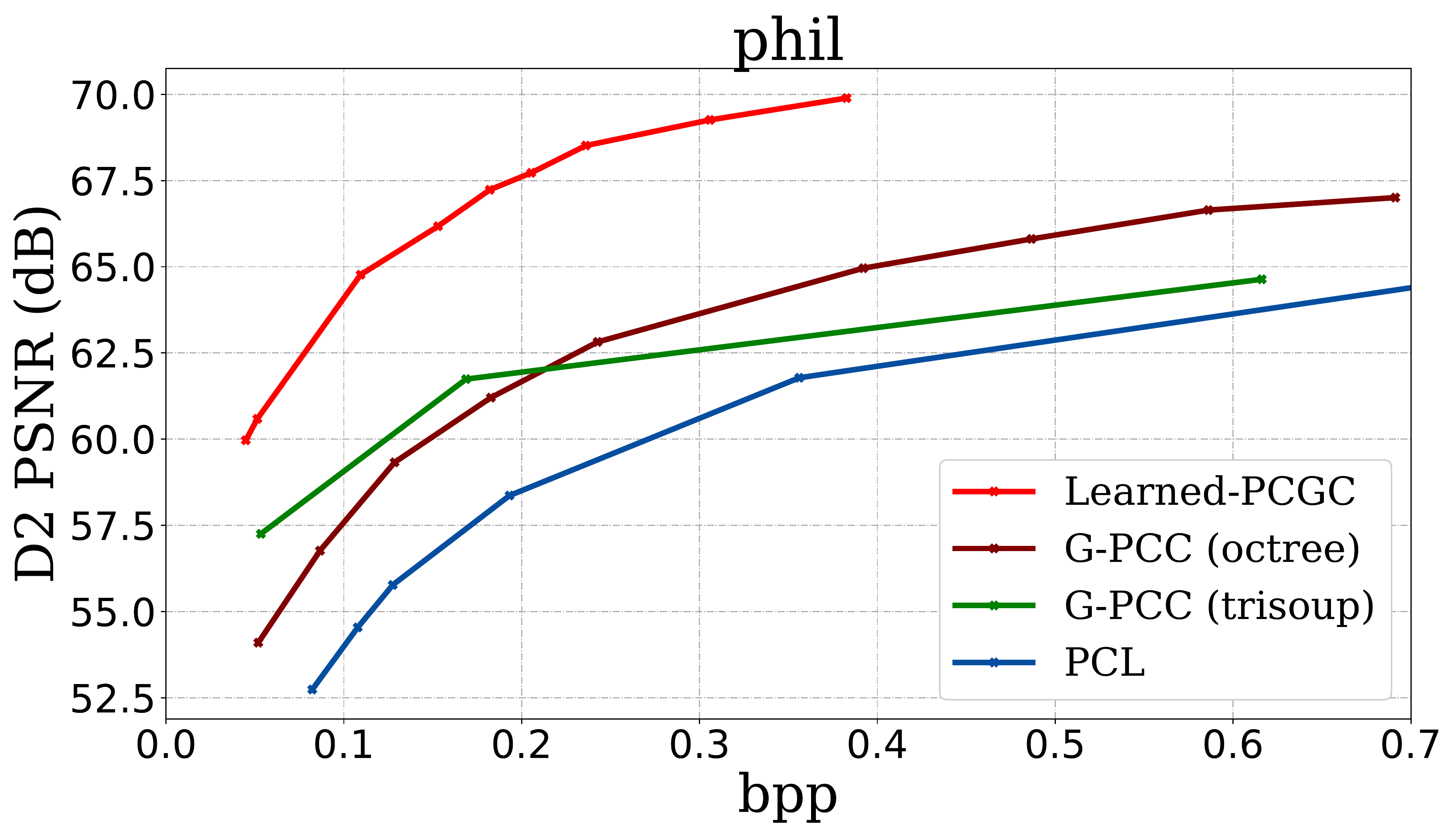}}\\%
\subfloat{\includegraphics[width=1.72in]{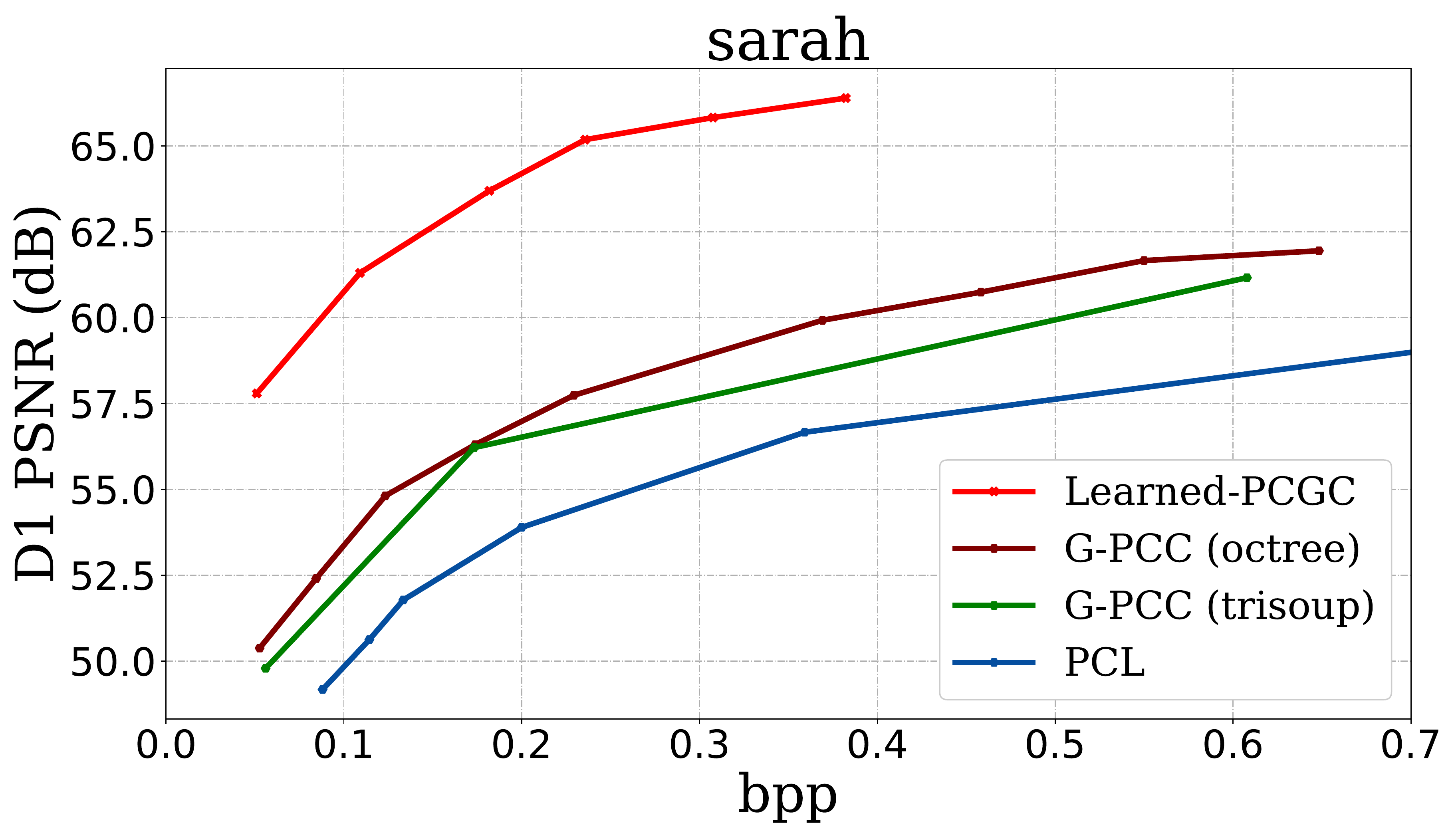}}%
\subfloat{\includegraphics[width=1.72in]{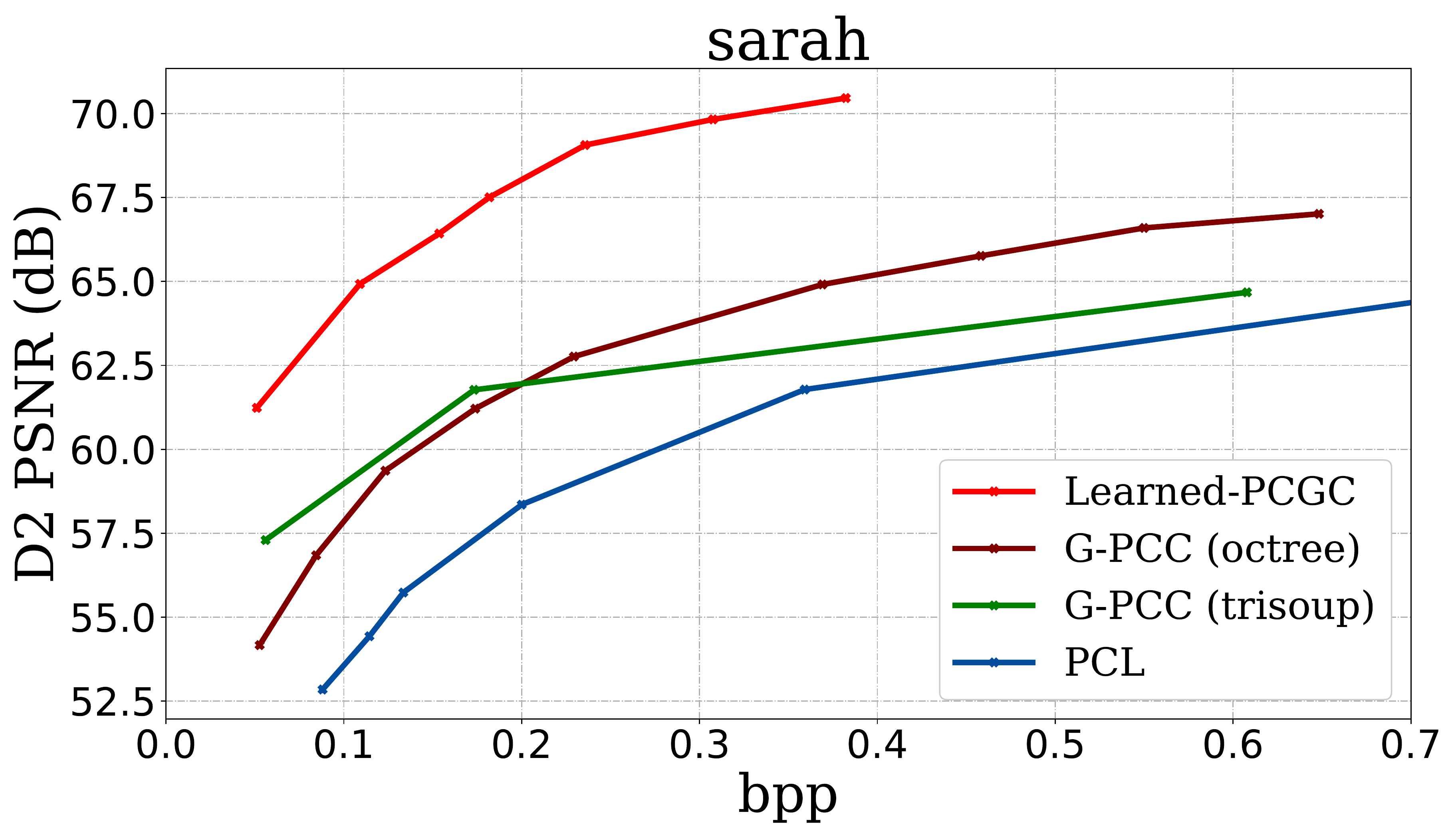}}%
\caption{R-D curves of Class B point clouds for PCL, G-PCC (octree), G-PCC (trisoup) and our Learned-PCGC: (left) D1 based PSNR, (right) D2 based PSNR.}
\label{rdcurvemicrosoft}
\end{figure}

\begin{figure}[t]
\centering
\subfloat{\includegraphics[width=1.72in]{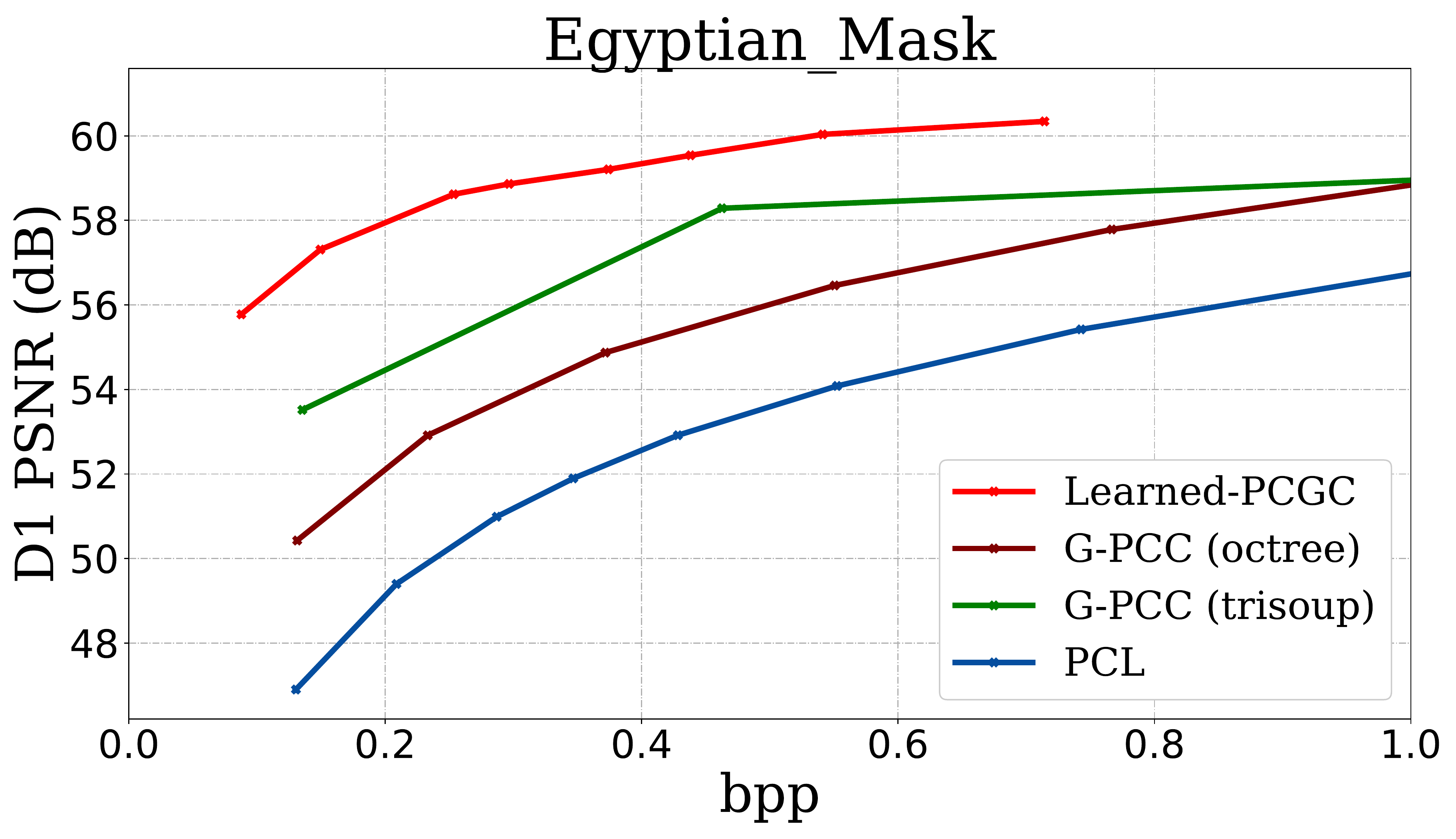}}%
\subfloat{\includegraphics[width=1.72in]{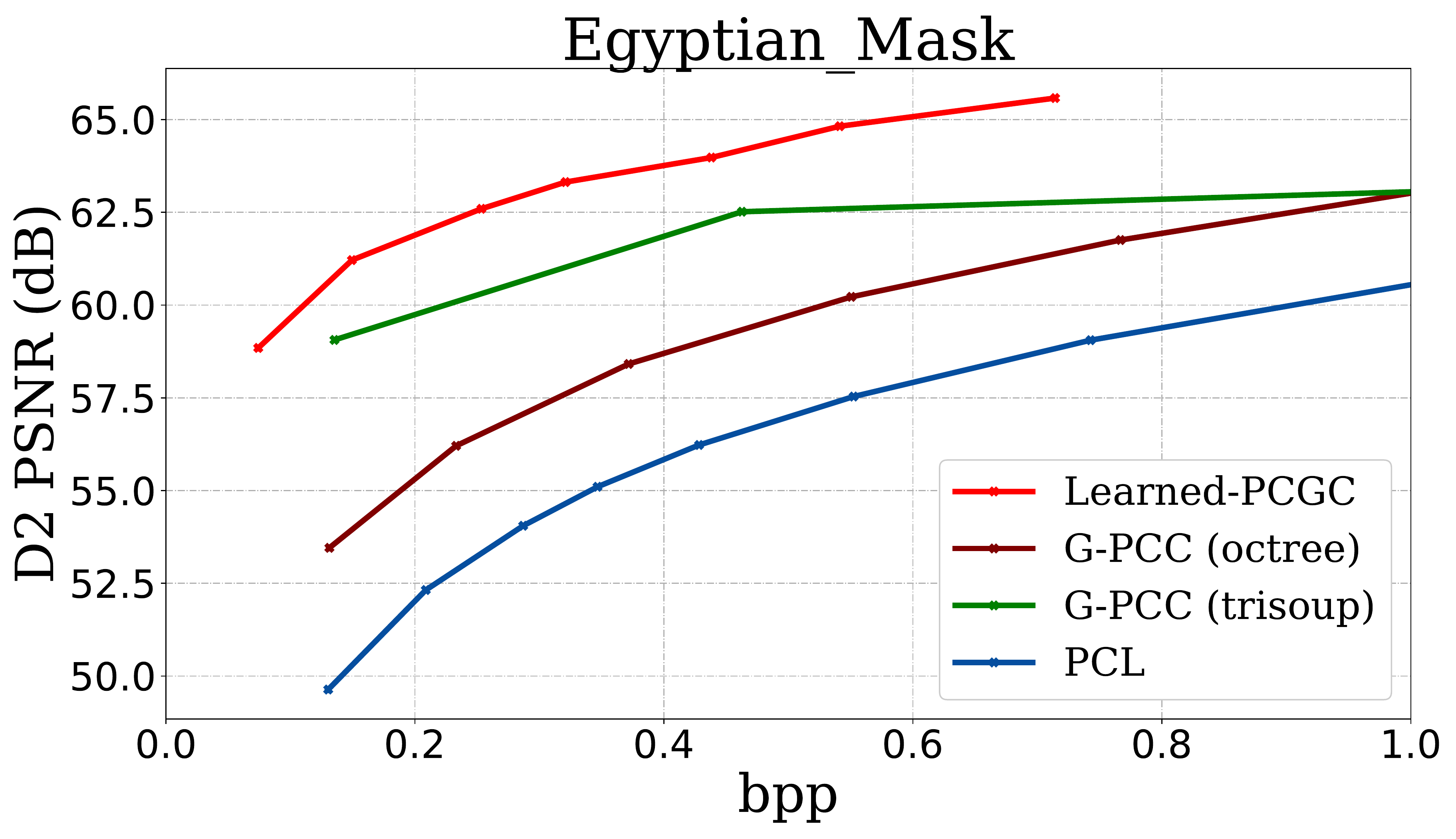}}%

\subfloat{\includegraphics[width=1.72in]{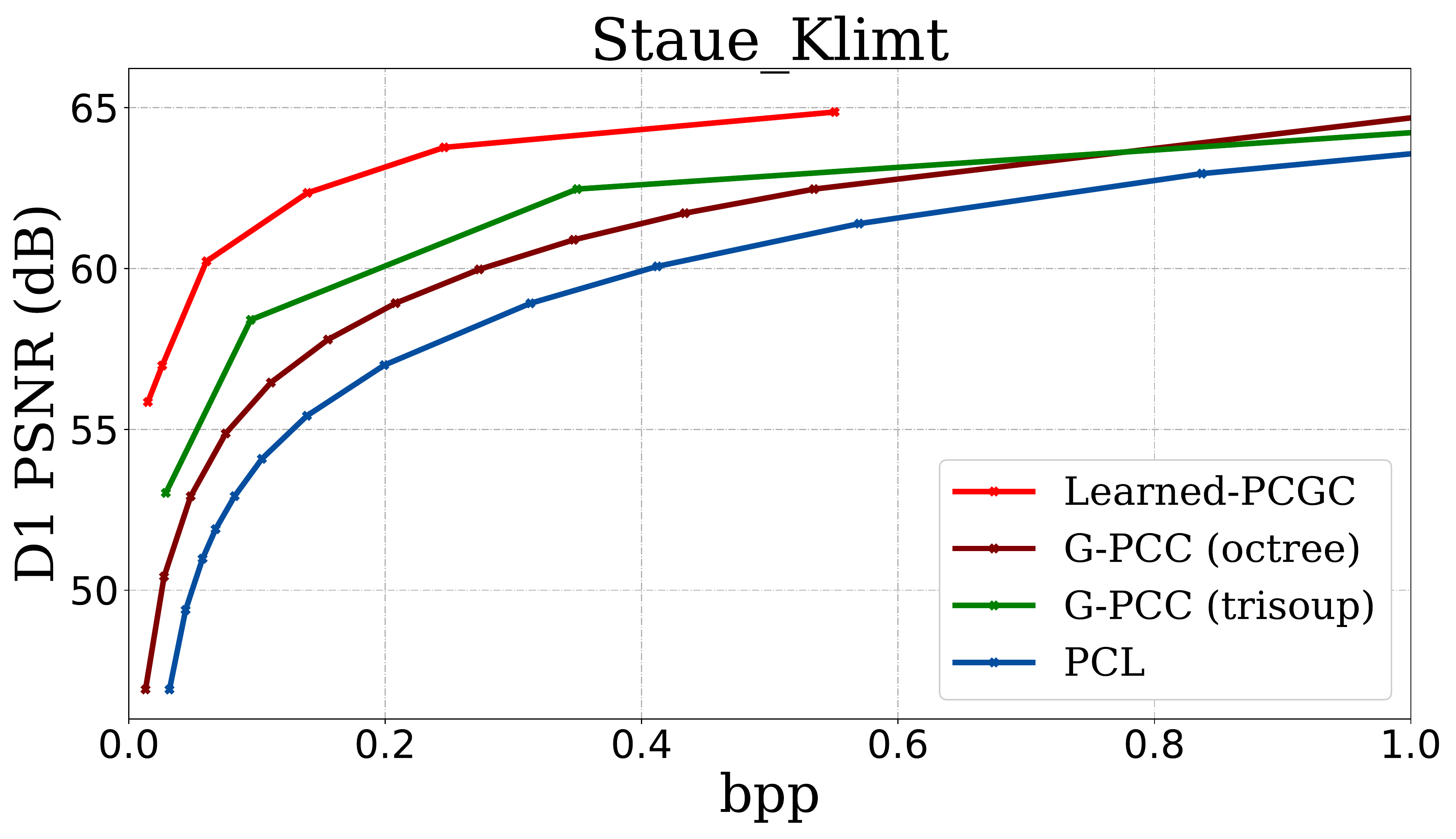}}%
\subfloat{\includegraphics[width=1.72in]{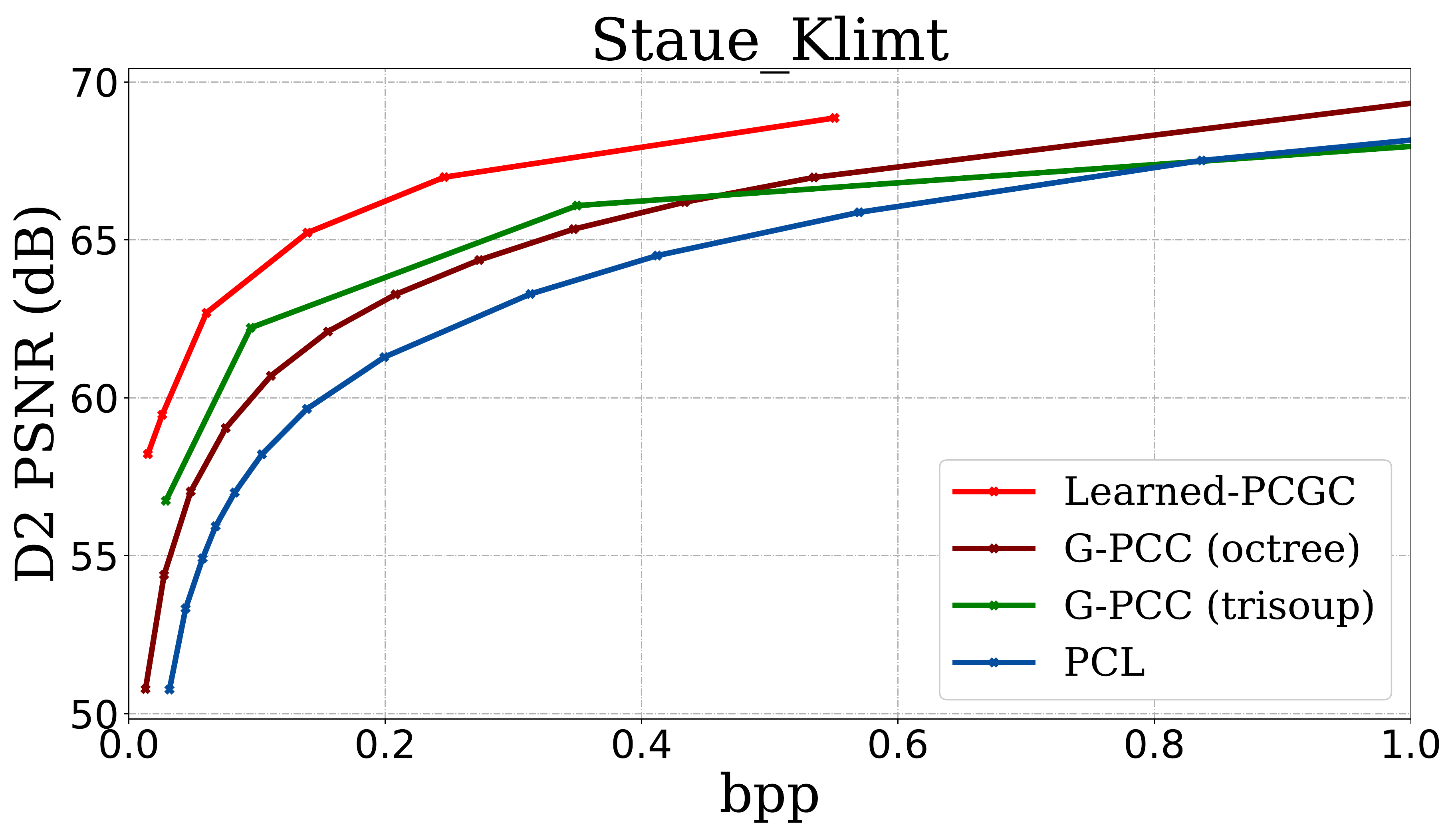}}%

\subfloat{\includegraphics[width=1.72in]{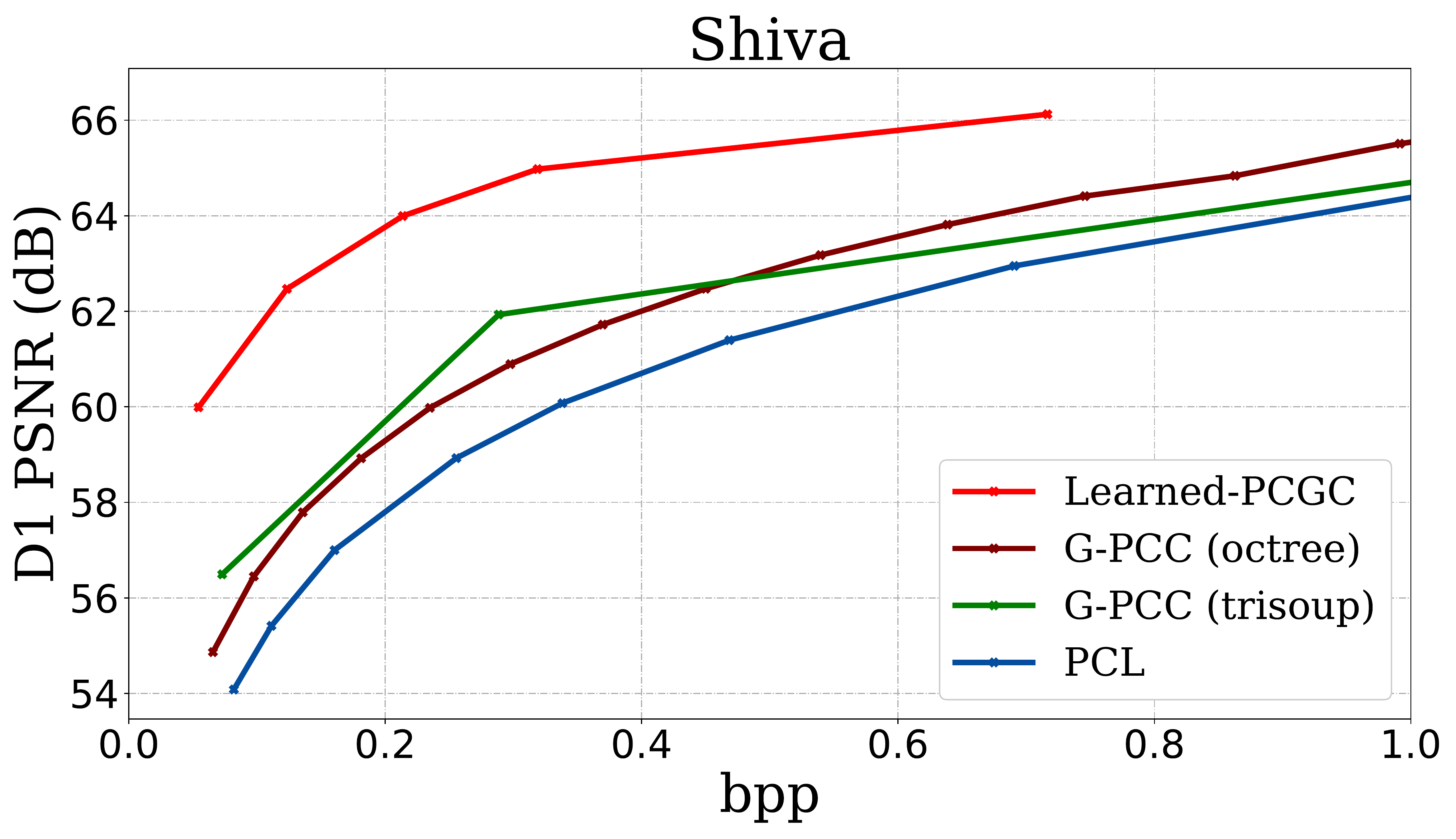}}%
\subfloat{\includegraphics[width=1.72in]{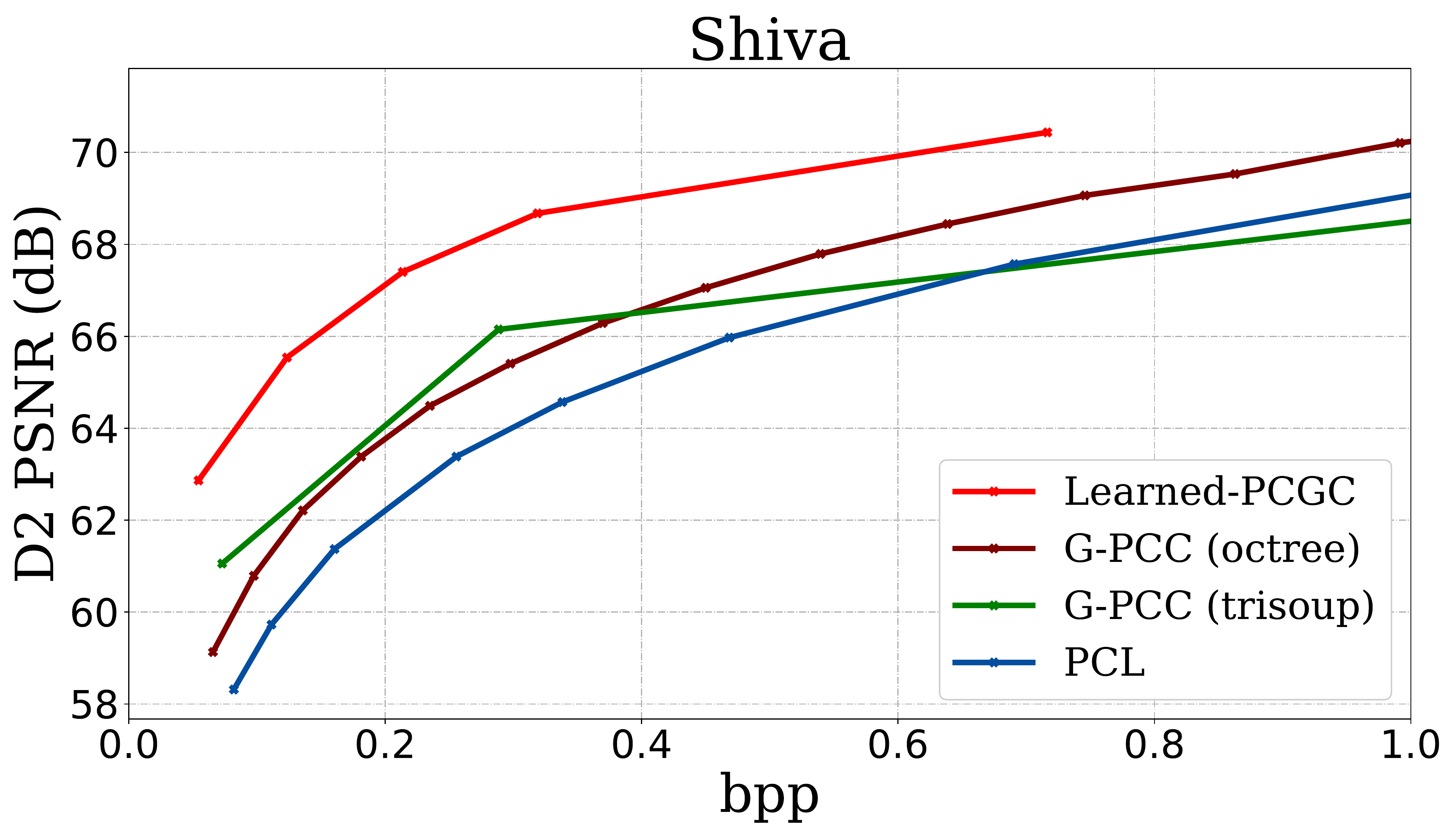}}%
\caption{R-D curves of Class C point clouds for PCL, G-PCC (octree), G-PCC (trisoup) and our Learned-PCGC: (left) D1 based PSNR, (right) D2 based PSNR.}
\label{rdcurveinanimate}
\end{figure}

Our method offers averaged -88\% and -82\% gains against PCL, -77\%  and -69\% gains against G-PCC (octree), -67\% and -62\% gains against G-PCC (trisoup), measured via respective D1 and D2 based BD-Rate. Illustrative Rate-distortion curves are presented in Figs.~\ref{rdcurve8i},~\ref{rdcurvemicrosoft}, and ~\ref{rdcurveinanimate}. 

As reported previously, our Learned-PCGC exceeds current G-PCC and PCL based geometry compression by a significant margin. For all testing Classes, e.g., dense or sparse voxel distributions, complete or incomplete surface, etc, the compression efficiency of our method consistently remains. On the other hand,  our training dataset is the watertight point clouds generated from the ShapeNet~\cite{chang2015shapenet}, not directly covering the sparse voxel distribution as in Class C. However, our model still works  by applying a simple scaling. All these observations justify the generalization of our method for various application scenarios.

In addition to the comparisons with those 3D model based geometry compression (e.g., PCL, G-PCC in Table~\ref{BDBRG-PCC}), we have also extended the discussion to the projection-based approach, e.g., V-PCC.

We use the latest TM2-v6.0~\cite{tmc2} to demonstrate the efficiency of standard compliant V-PCC solution.  For a fair comparison, we set the mode to ``All-intra (AI)'' and only compress the single frame of the dynamic point cloud, following the same test condition aforementioned~\cite{Sebastian2018common}. We set a variety of quantization parameters (QP = 32, 28, 24, 20, and 16) to derive sufficient bit rates as well for coded geometry. Bit rates for geometry components (e.g., metadata, occupancy map, depth map~\cite{tmc2}) are separated from the attributes for performance validation.

\begin{figure}[t]
\centering
\subfloat{\includegraphics[width=1.72in]{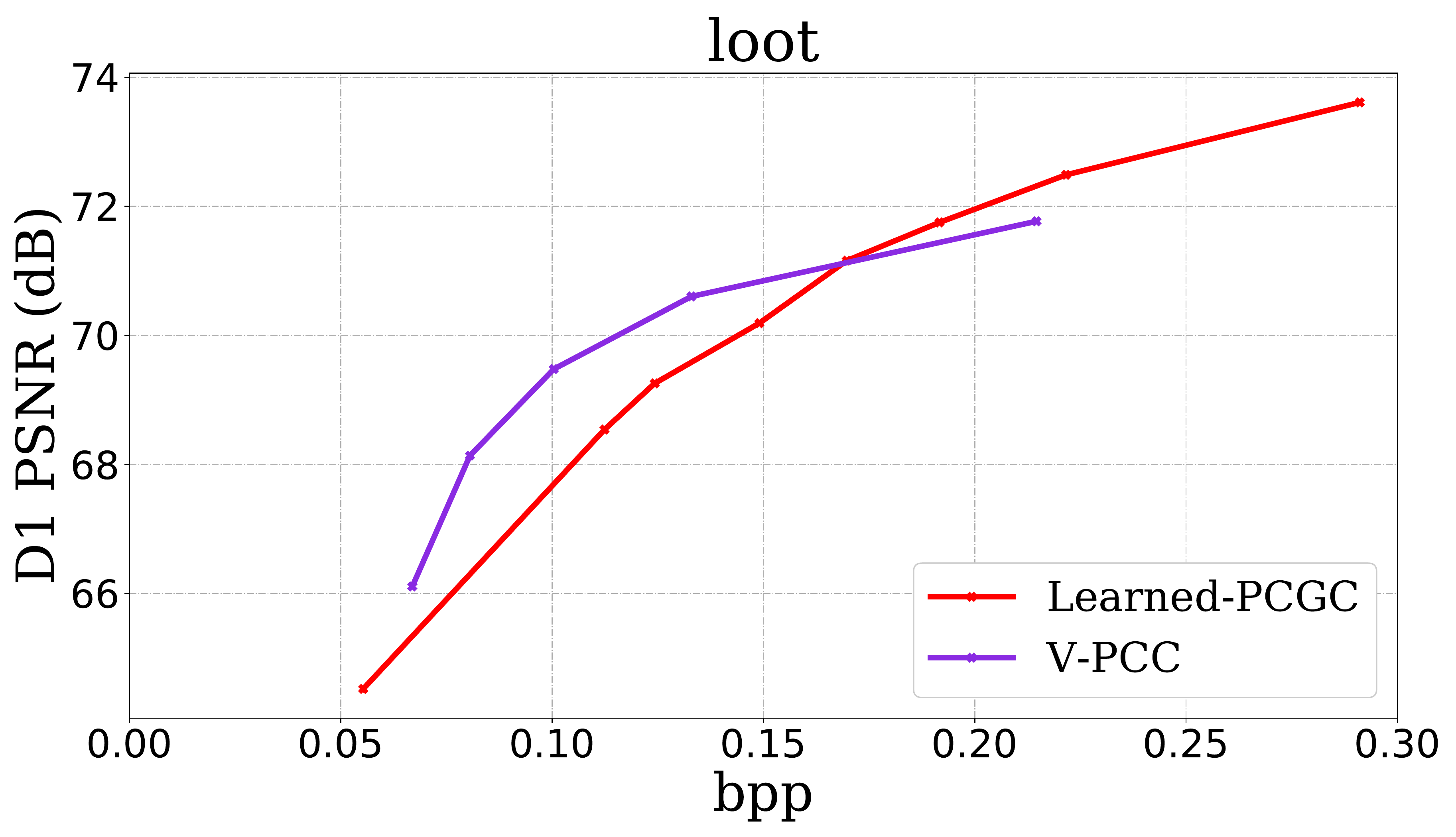}}
\subfloat{\includegraphics[width=1.72in]{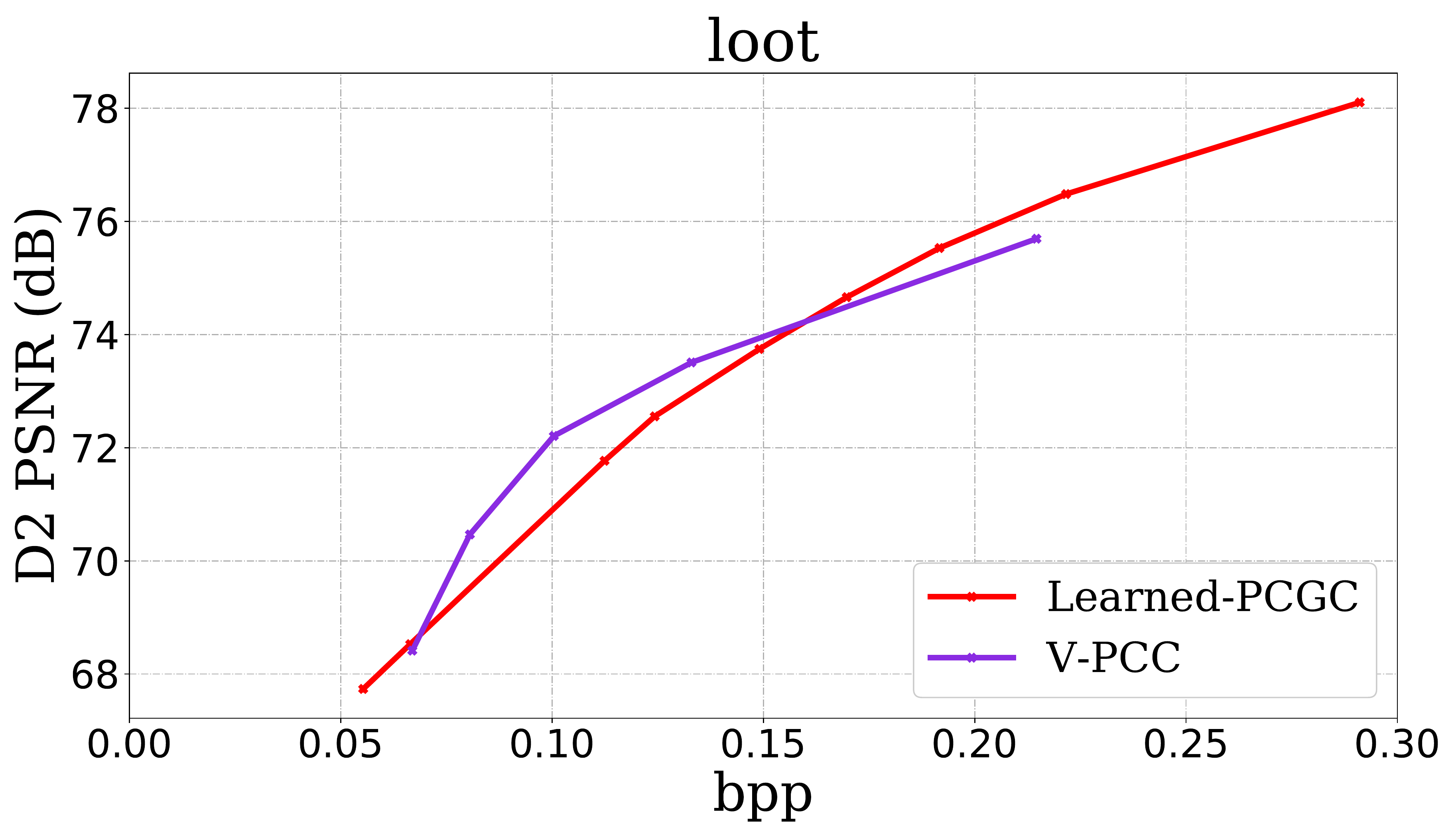}}%
 
\subfloat{\includegraphics[width=1.72in]{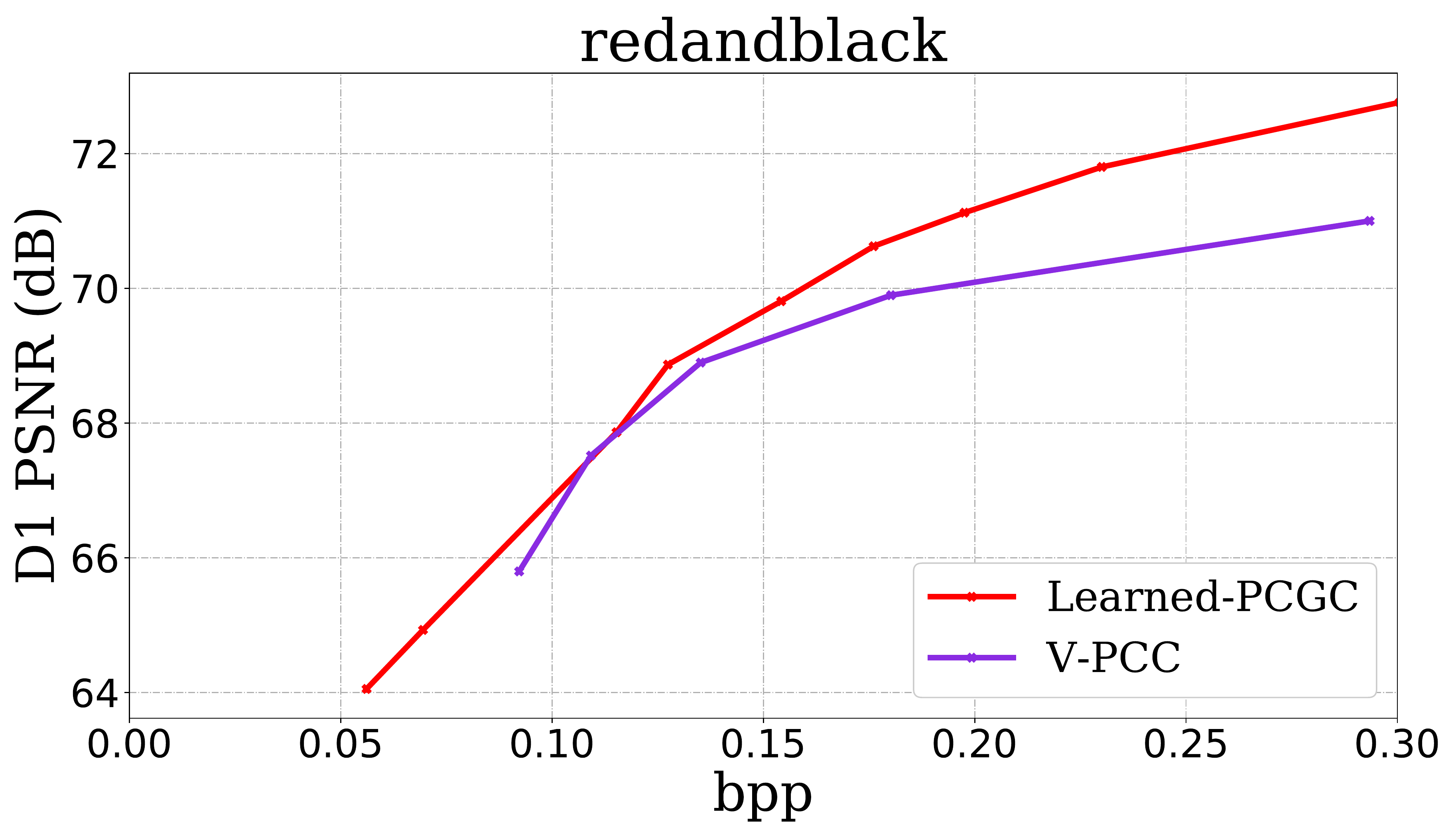}}%
\subfloat{\includegraphics[width=1.72in]{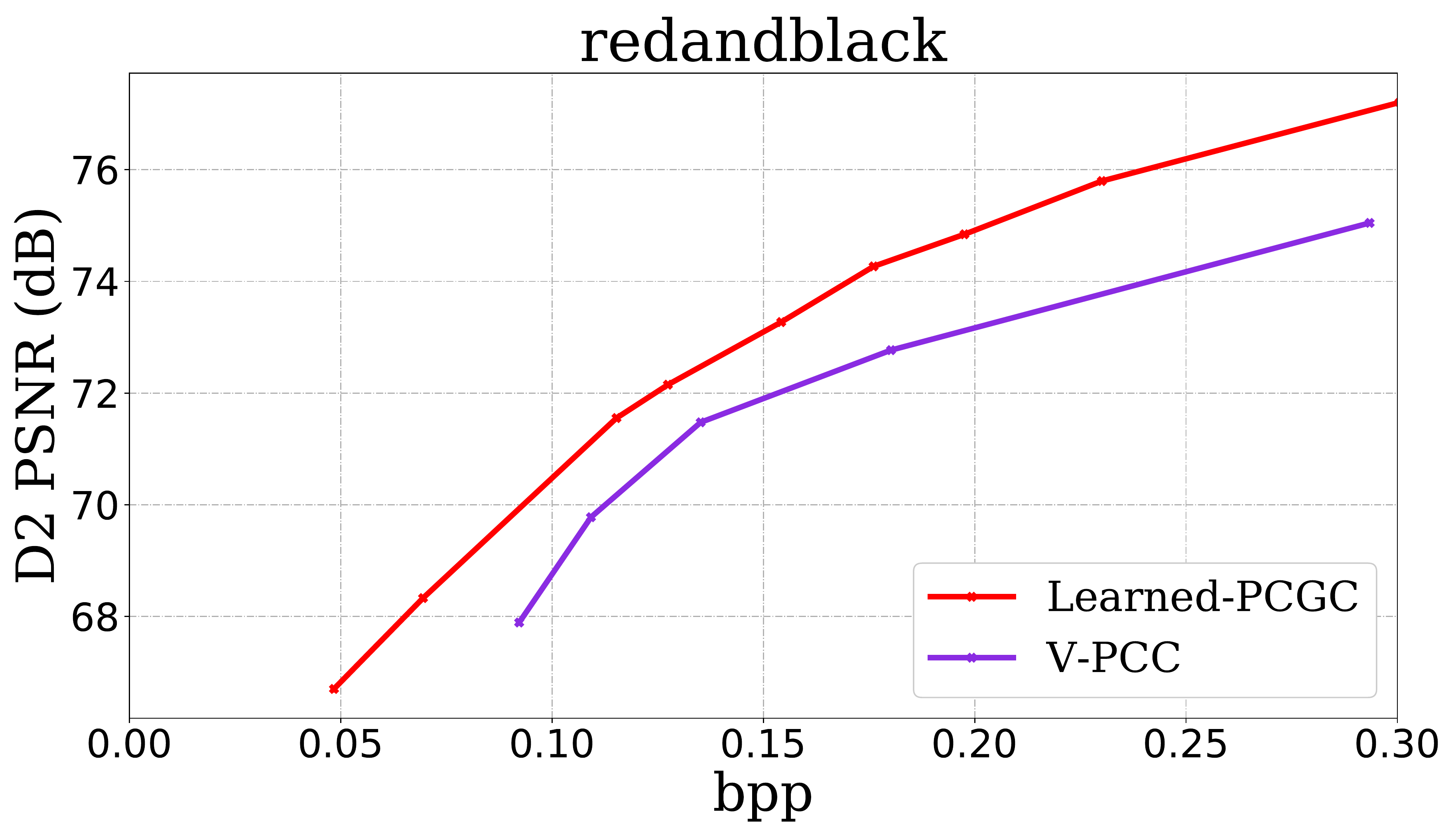}}%
 
\subfloat{\includegraphics[width=1.72in]{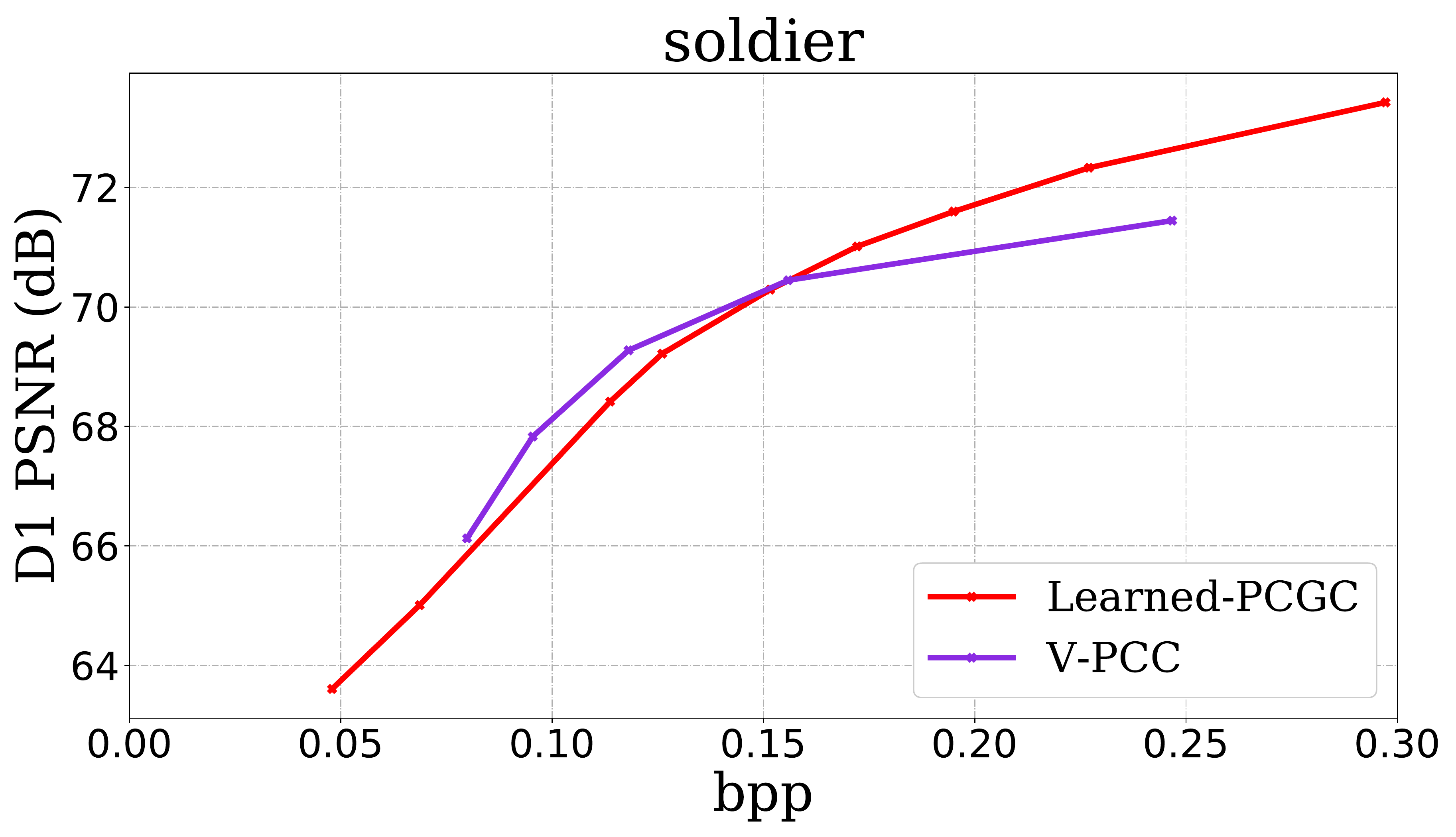}}%
\subfloat{\includegraphics[width=1.72in]{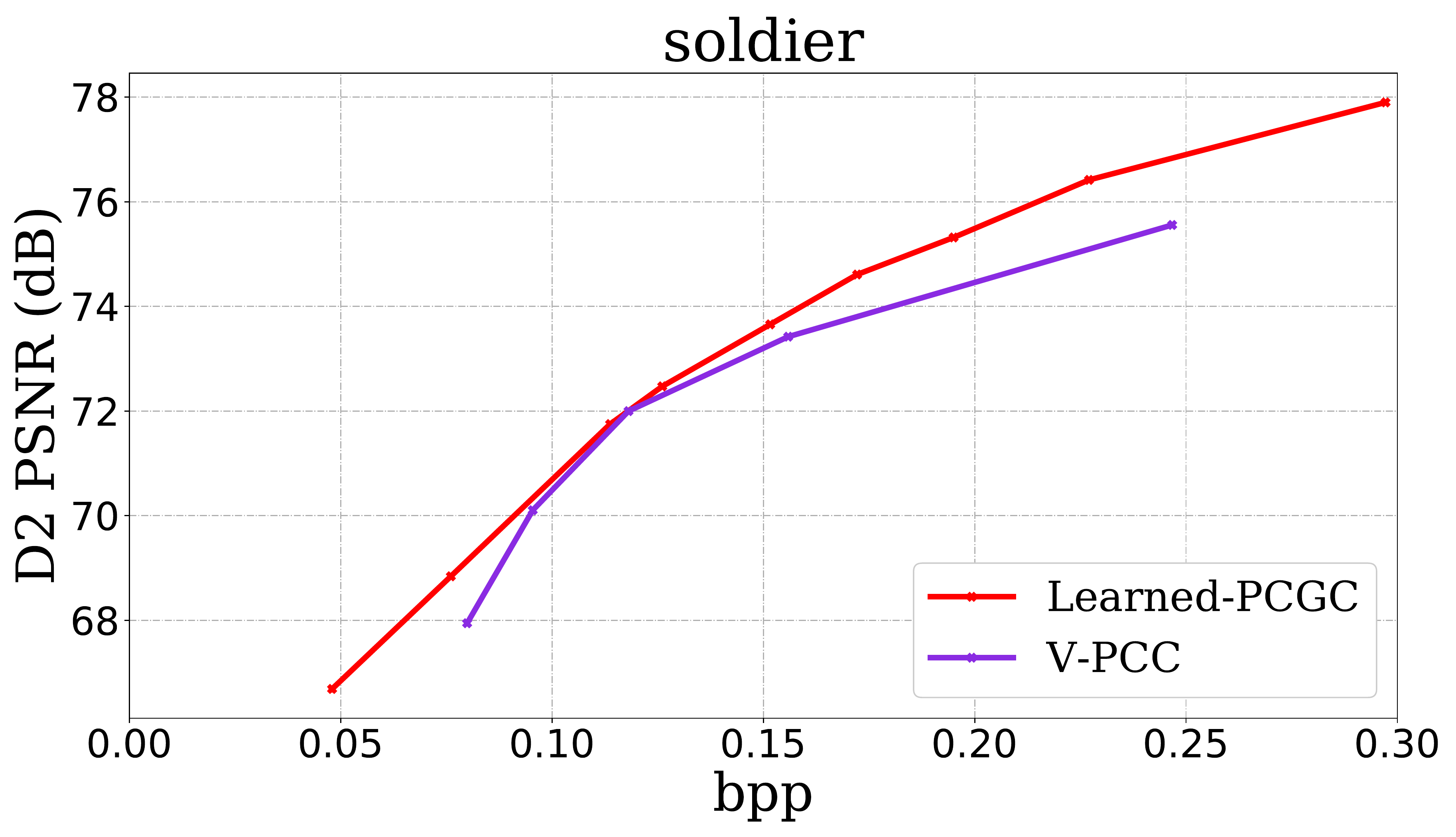}}%
 
\subfloat{\includegraphics[width=1.72in]{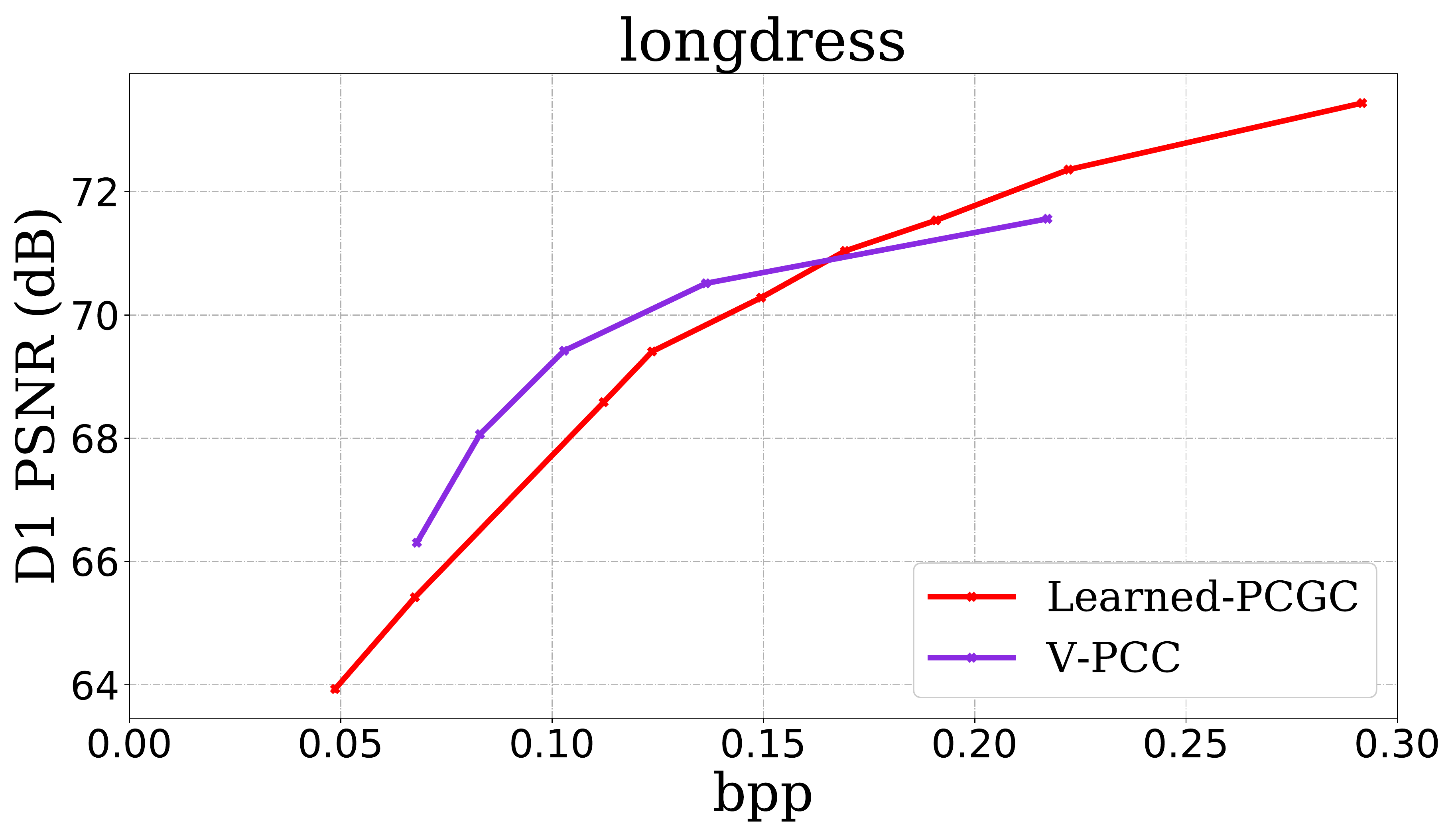}}%
\subfloat{\includegraphics[width=1.72in]{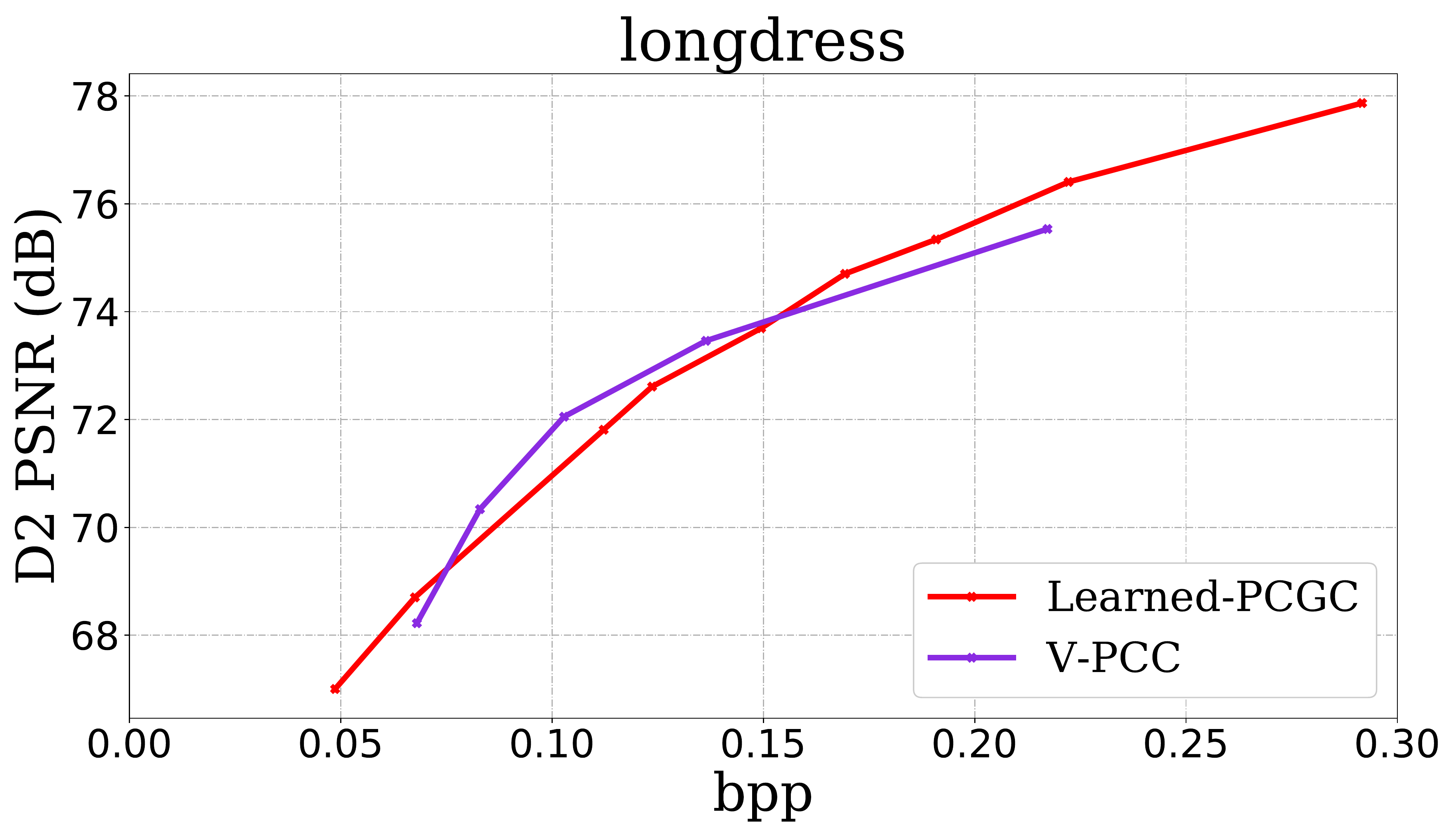}}%

\caption{Rate-Distortion Performance Comparision between V-PCC and our Learned-PCGC using Class A point clouds: (left) D1 PSNR, (right) D2 PSNR.}
\label{rdcurvevpcc}
\end{figure}

Again, our Learned-PCGC achieves comparable performance with V-PCC based geometry compression, as shown in Fig.~\ref{rdcurvevpcc}.  BD-Rate improvements are further put in Table~\ref{bdbrvpcc}. Results have shown that averaged +8.16\% D1 BD-Rate loss but -4.31\% D2 BD-Rate gains are captured. 
V-PCC performs better on \emph{Loot} and \emph{Longdress}, while our Learned-PCGC works better on \emph{Redandblack} and \emph{Soldier}, as reported in Table~\ref{bdbrvpcc}. 
Our analysis suggests that the more occluded region, the better compression efficiency of our Learned-PCGC. This is because our method inherently captures the voxel distribution in 3D space, {regardless of occlusion or shape incompleteness that cannot be well exploited by the project-based method.}

\begin{table}[t]
\centering
\renewcommand{\arraystretch}{1.3}
\caption{BD-Rate Efficiency of Learned-PCGC against V-PCC}
\label{bdbrvpcc}
\begin{tabular}{|c|c|c|}
\hline
\multirow{2}{*}{Point Cloud} & \multicolumn{2}{c|}{V-PCC}  \\ \cline{2-3} 
& D1 (p2point) & D2 (p2plane) \\
\hline
Loot                       &	21.00 	& 8.59 \\
\hline
Redandblack                  &	-8.99 &	-21.87 \\
\hline
Soldier                      &	3.84 &	-7.47 \\
\hline
longdress                    &	16.81  &	3.51  \\
\hline
\textbf{Average}           &	8.16  &	\textbf{-4.31} \\
\hline
\end{tabular}
\end{table}

\begin{figure}[hbtp]
\centering
\includegraphics[width=3.6in]{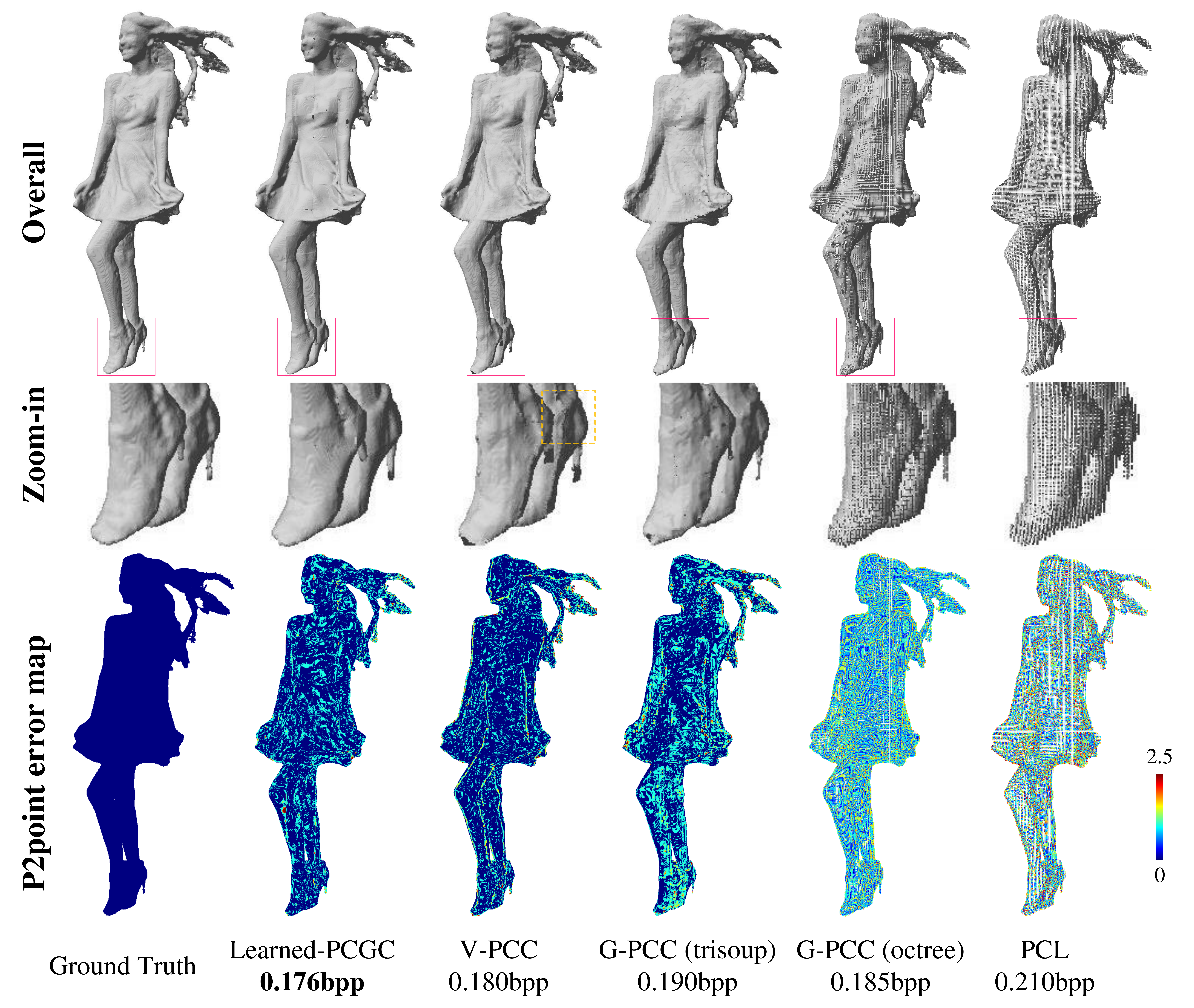}
\caption{Visual comparison of ``redandblack'' for ground truth, our Learned-PCGC, V-PCC, G-PCC (trisoup), G-PCC (octree),  and PCL. Compressed bits are set closely for all methods.}
\label{fig:vis_redandblack}
\end{figure}

\begin{figure}[hbtp]
\centering
\subfloat{\includegraphics[width=3.6in]{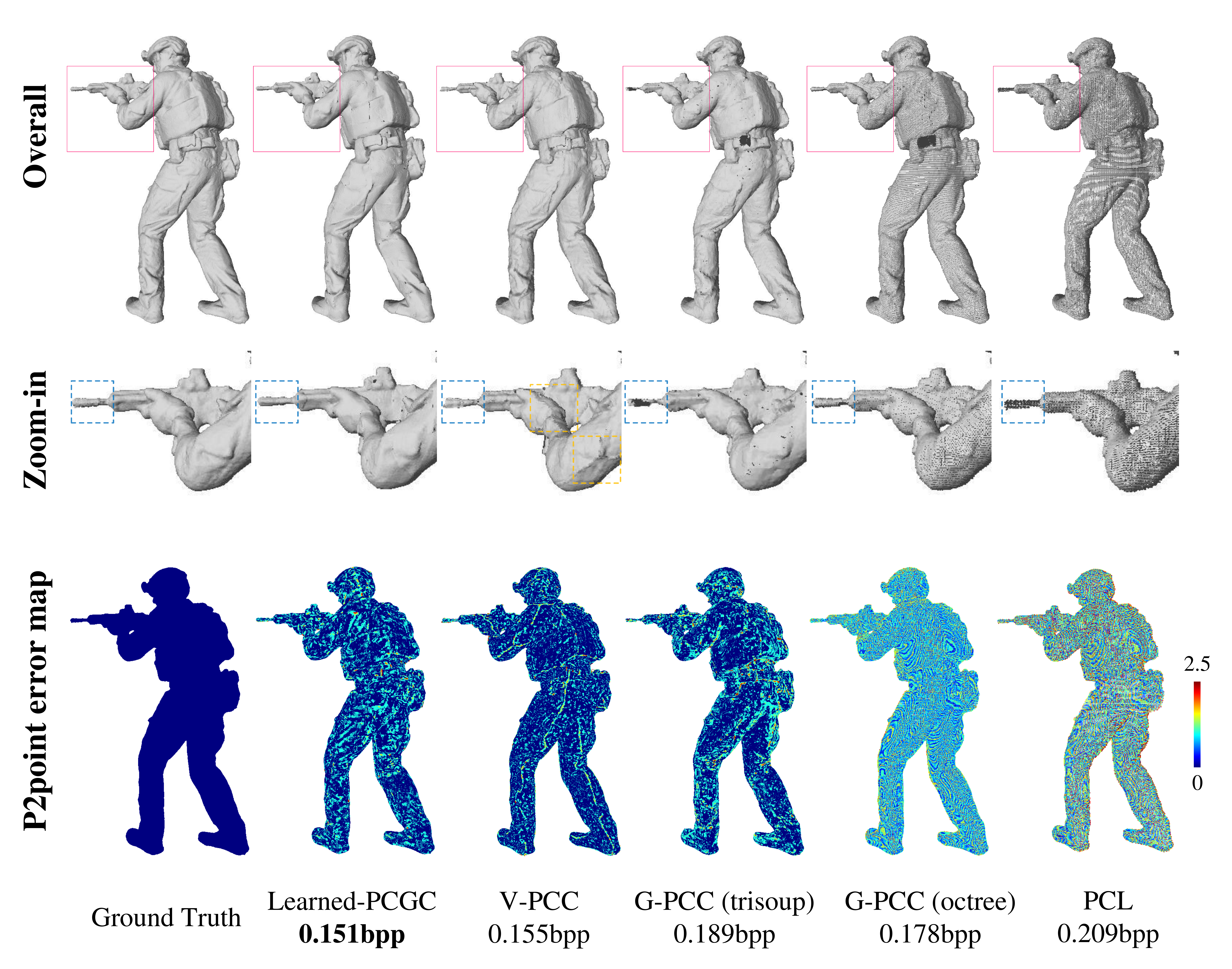}}%
\caption{Visual comparison of ``soldier'' for ground truth, our Learned-PCGC, V-PCC, G-PCC (trisoup), G-PCC (octree), and PCL. Compressed bits are set closely for all methods.}
\label{fig:vis_soldier}
\end{figure}

\begin{figure}[hbtp]
\centering
\subfloat{\includegraphics[width=3.6in]{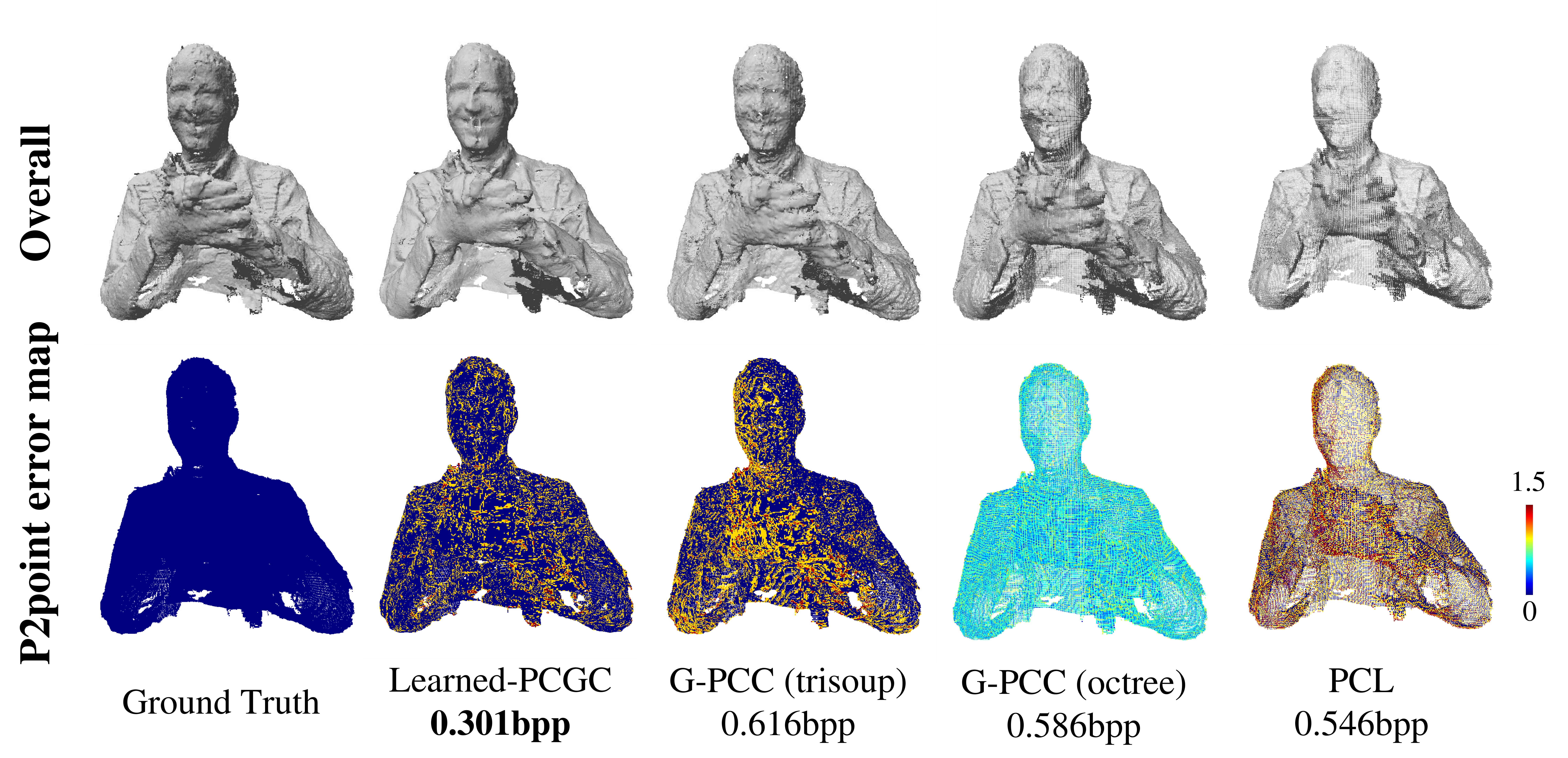}}%
\caption{Visual comparison of ``phil'' for ground truth, our Learned-PCGC,  G-PCC (trisoup), G-PCC (octree), and PCL. Note that bits consumed by our method is about half of those existing methods.}
\label{fig:vis_phil}
\end{figure}

{\bf Subjective Evaluation}
We show the decoded point clouds from different methods and the ground truth in Figs.~\ref{fig:vis_redandblack},~\ref{fig:vis_soldier}, and ~\ref{fig:vis_phil}, we recommend zooming in to see the detail.  To visualize the point clouds, we first compute the normal for each point using $20$ neighbor points, then we set parallel lighting in the front and render the points as Lambert unit.  By this means, we could observe the detailed geometry which is more intuitive than vertex-color rendered image. {We also plot the error map based on the point-to-point (P2point) D1 distance between decoded point clouds and ground truth to visualize the error distribution.} We can see that our method preserves the detailed geometry and generates visually high-quality point clouds.  Though V-PCC performs well in quantitative objective comparison, its reconstructed point clouds contain obvious seams as shown in the yellow dotted box, shown in zoomed-in area of Fig.~\ref{fig:vis_redandblack} and~\ref{fig:vis_soldier}.  This is because its method encodes point clouds by projecting them to different views, so it is difficult to avoid seams when fusing projected point clouds in the decoding phase.   {We also find that G-PCC (tirsoup) codec may lose geometry details (e.g., visible holes shown in Figs.~\ref{fig:vis_redandblack} and~\ref{fig:vis_soldier}). The reconstructed point clouds of G-PCC (octree) codec and PCL are much sparser as they could only retain much fewer points at comparative bit rate budget.}

An interesting observation is that our reconstructed point cloud fills some broken parts in the ground truth PC. The broken part is produced due to incomplete or failed scans.  We highlight the repaired part using the blue dotted box in Fig.~\ref{fig:vis_soldier}.  We think this is because we use ShapeNet~\cite{chang2015shapenet} to generate the point cloud samples for training, where most of them are fine mesh models designed by CAD software.  The high-quality training data make the distortion of our reconstruction is inclined to complete and smooth shapes with lower noise.  In contrast, the distortion of other methods is inclined to random noise.
%

\section{Ablation Studies}
We further extend our studies by examining various aspects of our Learned-PCGC, including the partition size, hyperpriors, adaptive thresholding, to demonstrate the robust and reliable performance of our method.

{\bf Partition Size.}
Analogous to the size of the coding tree unit in HEVC, we could set different partition sizes $W$ to explore its impact on the coding efficiency and implementation complexity for practice. 

\begin{figure}[t]
\centering
\subfloat{\includegraphics[width=1.72in]{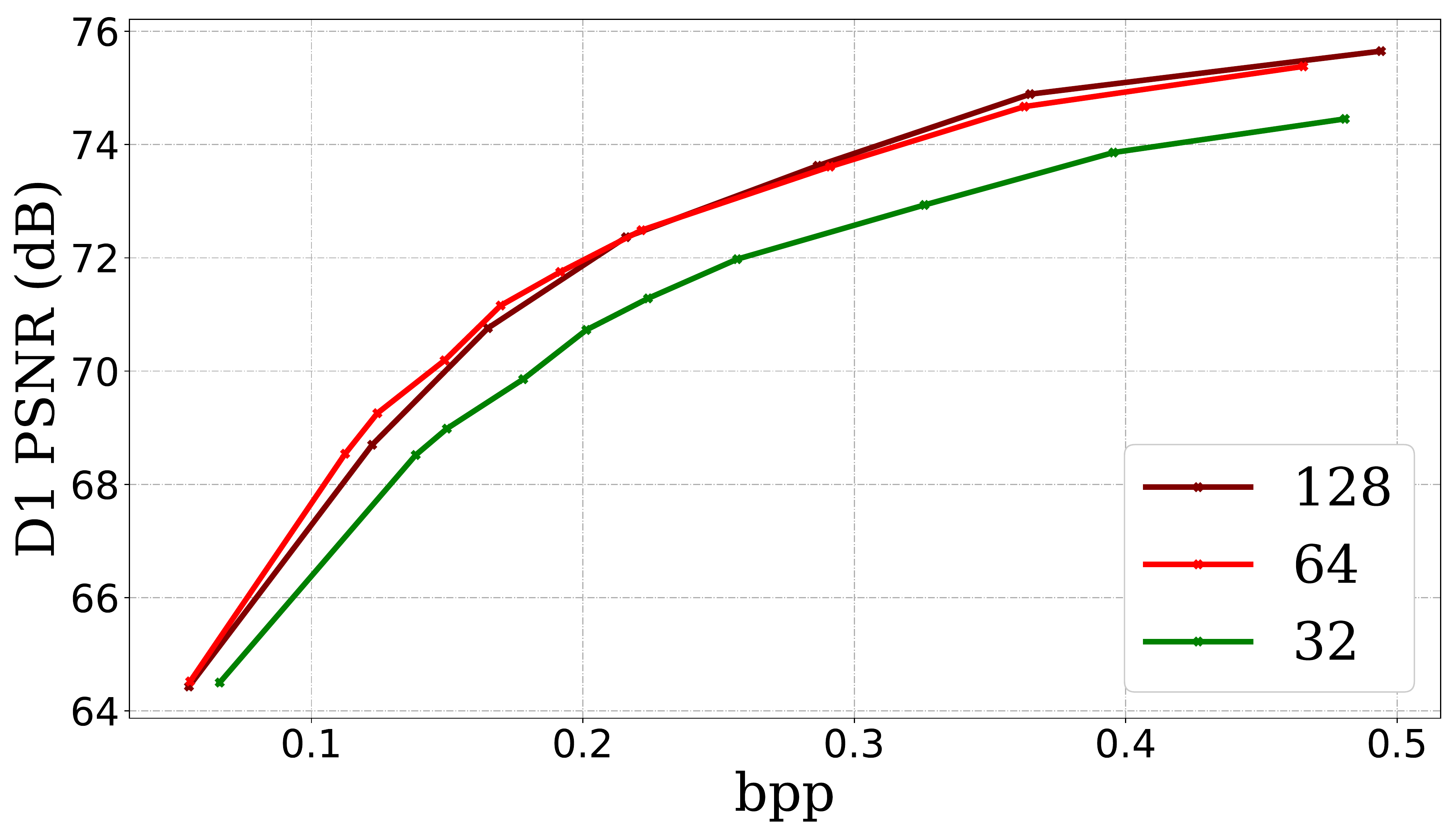}}%
\subfloat{\includegraphics[width=1.72in]{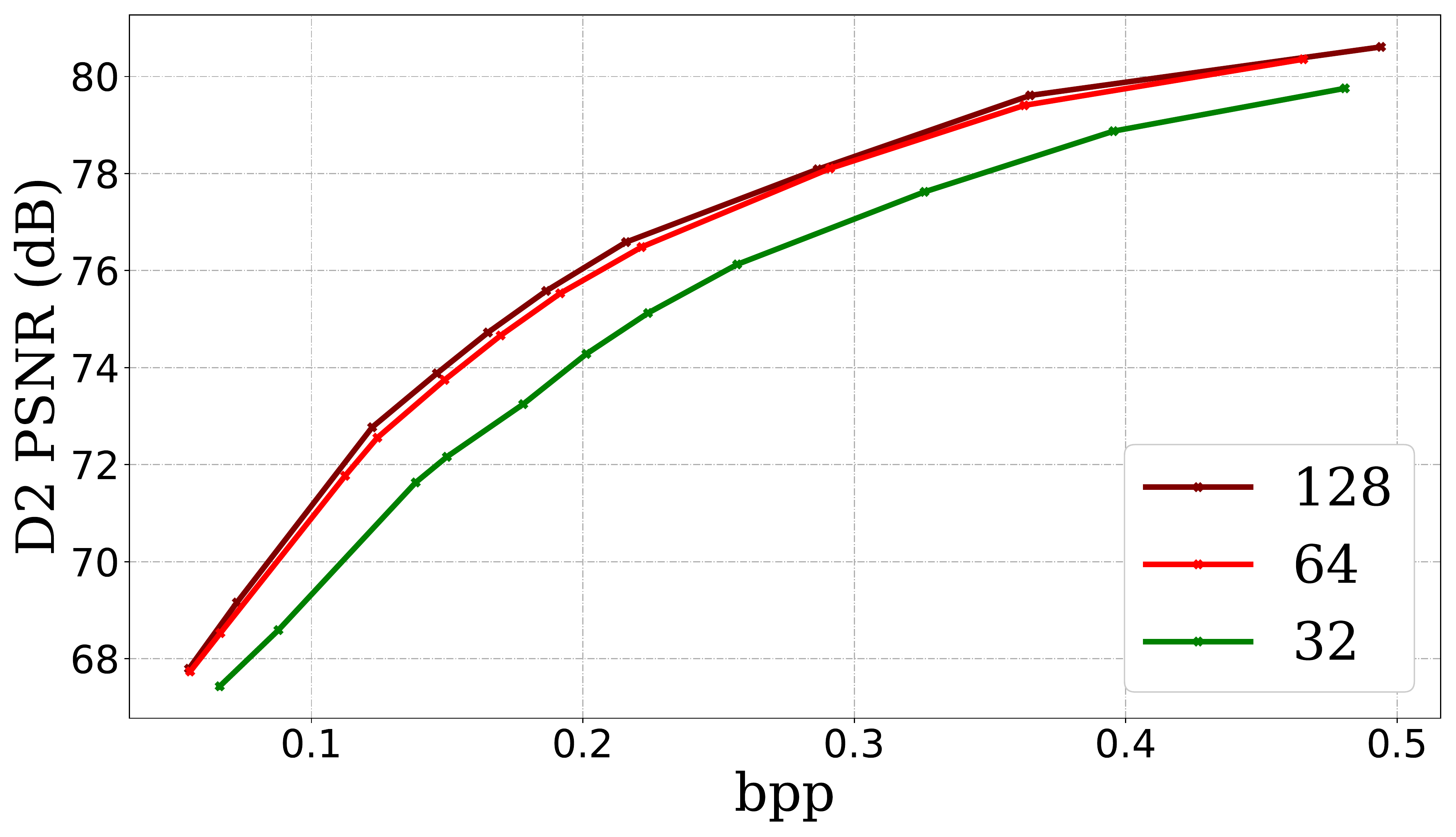}}%
\caption{Rate-distortion efficiency at different cube sizes ($W$). ``Loot" is presented as an example.}
\label{fig:cubesize}
\end{figure}

In the subsequent discussion, we have exemplified our studies using  \emph{Loot} at three different $Ws$, i.e., $W=$32, 64 and 128. Other testing materials share the similar outcomes. As illustrated in Fig.~\ref{fig:cubesize}, BD-Rate gains about 20\% from $W = 32$ to $W = 64$, but almost keeps the same from $W = 64$ to $W = 128$.

In addition to the BD-Rate, we have also provided other factors, e.g., the total number of cubes (cube\#), metadata overhead (meta\_bits), time (second) and memory consumption (mem) when executing the simulation, in Table~\ref{table:cubesize}.  Time and memory consumption given here for processing each cube, are tested on a computer with an Intel i7-8700 CPU and a GTX1070 GPU (with 8G memory).  All of these factors have substantial impacts on the algorithm complexity for implementation.  For example, the smaller is $W$, the better is parallel processing with less memory consumption and computational time. However, it comes with {more blocky artifacts} and BD-Rate sacrifice. Thus, a good choice of $W$ needs to balance the BD-Rate performance and implementation complexity. In this work, we choose $W = 64$.

\begin{table}[t]
\renewcommand{\arraystretch}{1.3}
\caption{Implementation Factors for Various $W$.}
\label{table:cubesize}
\centering
\begin{tabular}{|c|c|c|c|c|}
\hline
$W$ & cube\#  & meta\_bits (bpp) & time & mem\\
\hline
128  & 51	 & 0.0015 & 0.78s & 2208 MB \\
\hline
\textbf{64} & 212 & 0.0046	 & 0.13s & 414 MB \\
\hline
32 & 790 &  0.0142 & 0.06s & 252 MB \\
\hline
\end{tabular}
\end{table}

{\bf Hyperpriors.}  
Hyperpriors $z$ have been used for accurate conditional entropy modeling for image compression in~\cite{balle2018variational,liu2019non}. Here we further examine its efficiency  in our Learned-PCGC.

Compared with the scenario only using a  \emph{factorized entropy model} for latent representations $\hat{y}$,  hyperpriors could improve the \emph{context modeling} and lead to better rate-distortion performance with the conditional probability exploration, yielding about 14.65\% BD-Rate gains from our experiments. 

{\bf Adaptive Thresholding.} Thresholding mechanism is applied to classify decoded voxel into a binary decision (e.g., 1 or 0) for its occupancy state. We aim to find a threshold that leads to the minimum distortion (e.g., D1 or D2) for reconstruction. 

\begin{figure}[t]
\centering
\subfloat{\includegraphics[width=1.72in]{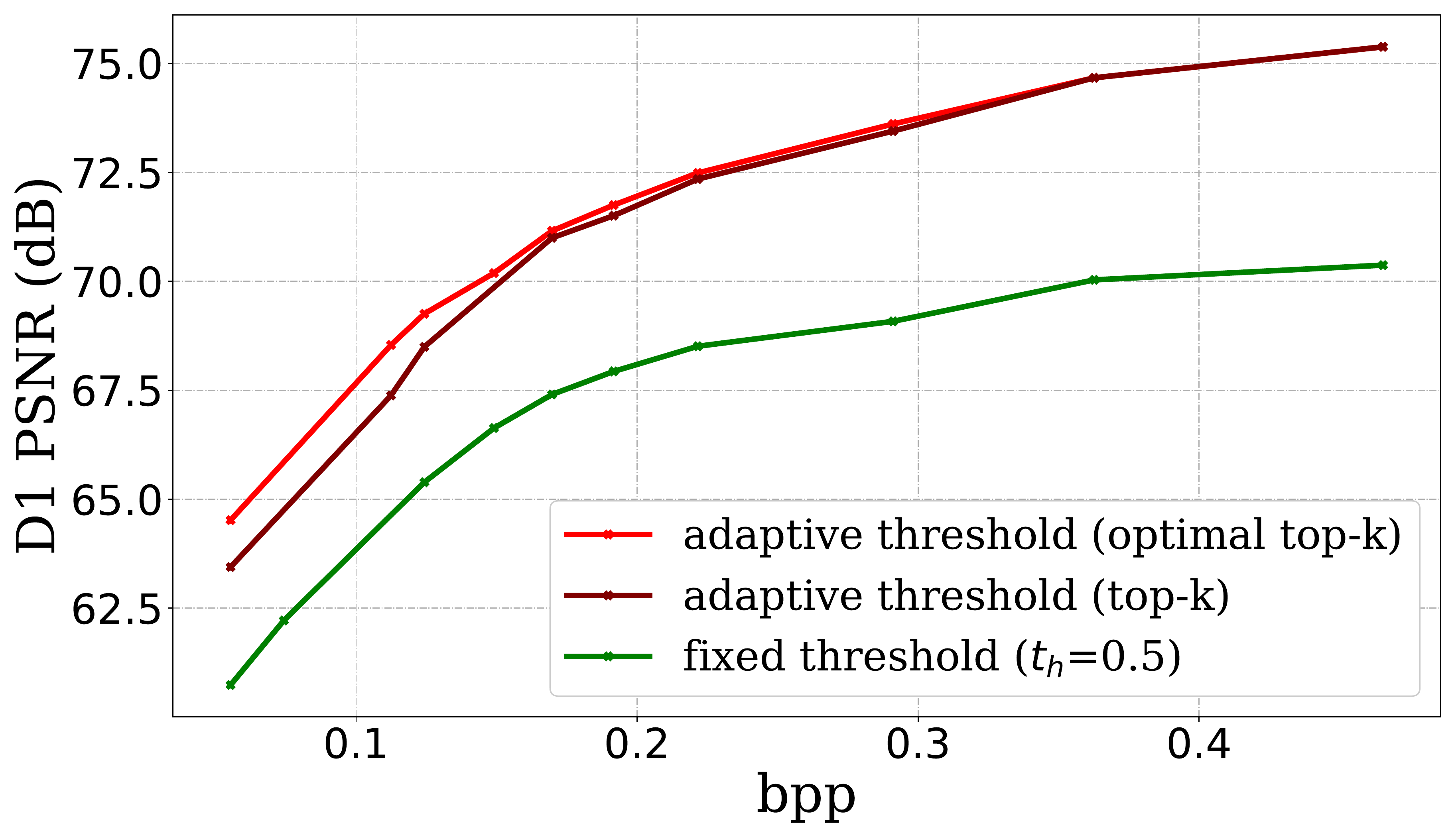}}%
\subfloat{\includegraphics[width=1.72in]{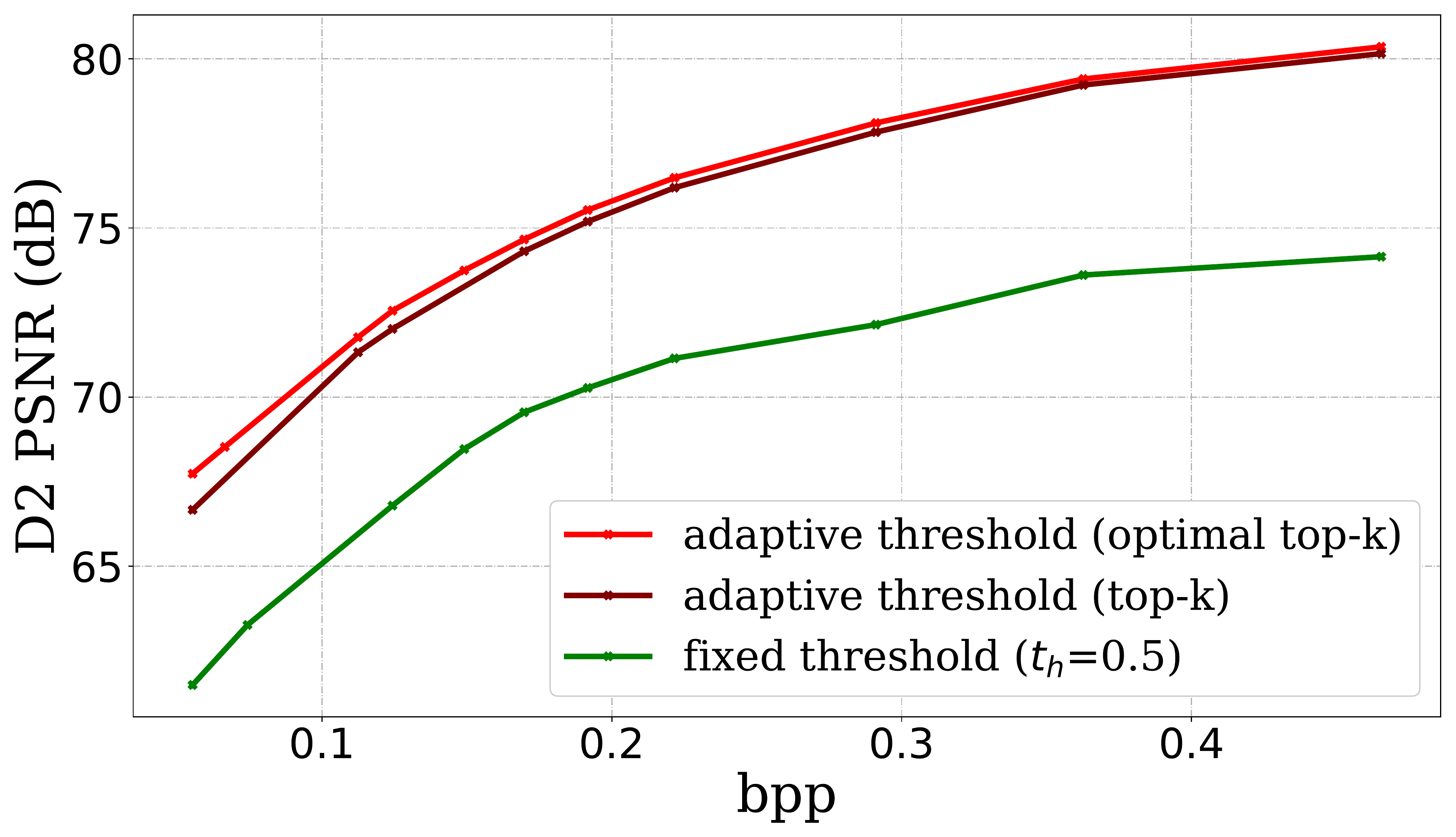}}%
\caption{BD-Rate illustration for ``Loot'' with adaptive thresholding based classification.}
\label{pointsnumbersselection}
\end{figure}

A straightforward way is to set a na\"ive value, such as the {$t_h =$ 0.5} as the global threshold to do classification for all cubes. However, performance suffers. Instead, we propose  order the decoded voxels $\tilde{x}$ and select top-$k$ ones, e.g., $k$ = {\tt num\_occupied\_voxel},  as the adaptive threshold, for each cube, leading to a noticeable BD-Rate gains in Fig.~\ref{pointsnumbersselection}.

\begin{figure}[t]
\centering
\includegraphics[width=3in]{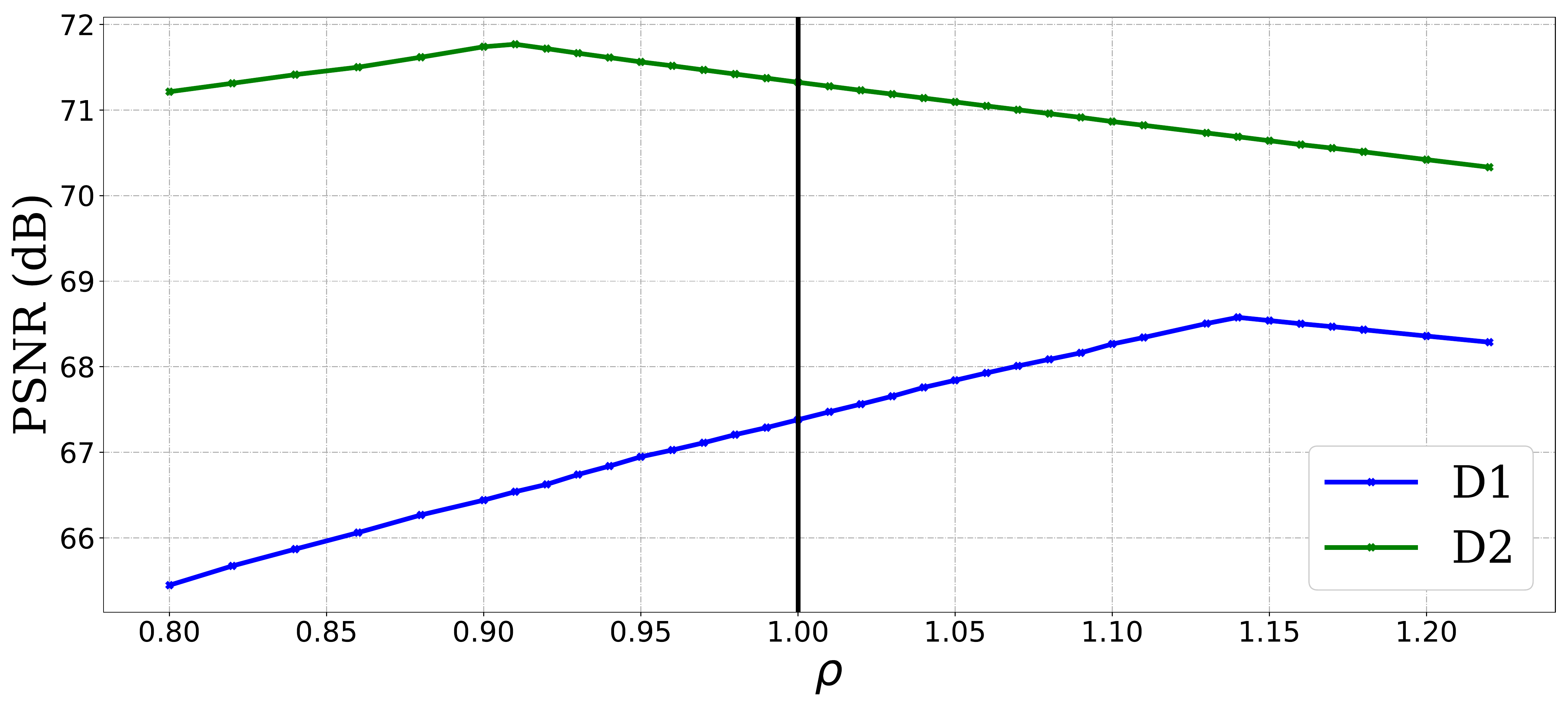}
\caption{Optimizing top-$k$ selection for voxel classification to minimise the D1 or D2 distortion.}
\label{fig:ratio}
\end{figure}

Since we are optimizing the top-$k$ selection to minimize D1 or D2 for classification, we further deeply study whether adapting $k$ can bring more gains by fine-tuning. Similarly as illustrated in Fig.~\ref{pointsnumbersselection}, adjusting $k$ for a fine-tuned $k_f$, i.e.,
\begin{align}
    k_f = \rho\cdot k, \mbox{~~~} 0.5< \rho < 2, \label{eq:optimal_k}
\end{align}  would yield BD-Rate improvement. For example, on average,  BD-Rate is gained about 8.9\% with optimal $\rho$ in  \eqref{eq:optimal_k}, or equivalent $k$, when minimizing D1 distortion; while about 6.7\% when minimizing D2 distortion. Optimal $\rho$ differs for D1 and D2 measures respectively, due to their fundamental variations in distance calculation, as visualized in Fig.~\ref{fig:ratio} for ``Loot'' at 0.11 bpp. More voxels are selected for optimal D1 distortion measurement, e.g., $\rho$ =1.14, while less voxels, e.g., $\rho$ = 0.91 are used for better reconstruction for D2 distortion.  It indicates that D2 measurement is more suitable for sparser point cloud.

{\bf Convolution Kernels.}
Our Learned-PCGC, including both main and hyper encoder-decoder pairs, requires 658,092 parameters in total for all embedded convolutions. In the current implementation, each parameter is buffered using 4-byte floating format. It is about $\approx$ 2.52MB storage, which is fairly small on-chip buffer requirement compared with other popular algorithms, {such as AlexNet~\cite{AlexNet} with 60 Million parameters, or GoogleNet~\cite{GoogleNet} using 4 Million parameters.} 

Our experiments have also revealed that current stacked VRN  with small convolutional kernels and deep layers offer much better performance compared with an alternative approach using the shallow network with larger convolutional kernel sizes.
This is mainly because that larger convolutions can not efficiently capture the spatial information due to sparse spatial distribution of voxels in a 3D space. But, deeper layers (with down-scaling) offers an effective way to exploit correlation in a variety of scales.

\section{Conclusion And Future Work}
\label{sec:conclusion}

A learning-based point cloud geometry compression method, so-called Learned-PCGC,  is presented in this work, which consists of stacked 3D convolutions for the exaction of latent features and hyperpriors, a VAE structure for accurate entropy modeling of latent features, and a weighted BCE loss in training and an adaptive thresholding scheme in inference for correct voxel classification. 

We have demonstrated the state-of-the-art efficiency of proposed Learned-PCGC, for point cloud geometry compression, objectively, and subjectively, in comparison to those existing standardized methods, for example, over 62\% and 67\% BD-Rate gains over G-PCC (trisoup),  and over 69\% and 76\% BD-Rate gains over G-PCC (octree), when the distortion is measured using D2 or D1 criteria respectively. On the other hand, our method also provides comparable compression efficiency against the projection-based MPEG V-PCC.  Subjective evaluations have also evident the superior performance of our proposed method with noticeable perceptual improvements.
 Additional ablation studies deeply dive into a variety of aspects of our proposed method by carefully analyzing the performance and efficiency.

As for future studies, there are several interesting avenues to explore. For example, recent PointConV~\cite{wu2018pointconv} might be borrowed to improve the efficiency of convolutions for the point cloud.
On the other hand, traditional distortion measurements, such as D1 and D2, still suffer from a low correlation with subjective assessment. A better objective metric is highly desired.

\section{Acknowledgement}
We are deeply grateful for the constructive comments from anonymous reviewers to improve this paper.



\end{document}